%% file: Main_survey.tex
\documentclass[]{interact}

\usepackage{multirow}
\usepackage{enumerate}

\input{H_mathdefs_interactive.tex}

\input{format_interactive.tex}
\usepackage{amsmath,amssymb,amsfonts}
\usepackage{graphicx}
\usepackage{textcomp}
\usepackage{xcolor}
\usepackage{hyperref}
\usepackage{caption}
\usepackage{subfigure}

\usepackage{epstopdf}

\theoremstyle{plain}
\newtheorem{theorem}{Theorem}[section]
\newtheorem{lemma}[theorem]{Lemma}

\theoremstyle{definition}
\newtheorem{definition}[theorem]{Definition}

\theoremstyle{remark}
\newtheorem{remark}{Remark}

\newtheorem{assumption}[theorem]{Assumption}
\begin{document}

\articletype{Review Article}

\title{ A review of path following control strategies for autonomous robotic vehicles: theory, simulations, and experiments   }

\author{
\name{Nguyen Hung\thanks{CONTACT Nguyen T. Hung. Email: nguyen.hung@tecnico.ulisboa.pt}, Francisco Rego, Joao Quintas, Joao Cruz, Marcelo Jacinto, David Souto, Andre Potes, Luis Sebastiao, Antonio Pascoal   }
\affil{Institute for Systems and Robotics (ISR), Instituto Superior Técnico (IST), University of Lisbon, Lisbon, Portugal}
}


\maketitle

\begin{abstract}
This article presents an in-depth review of the topic of path following for autonomous robotic vehicles, with a  specific focus on vehicle motion in two dimensional space (2D). From a control system standpoint, path following can be formulated as the problem of stabilizing a path following error system that describes the dynamics of position and possibly orientation errors of a vehicle with respect to a path, with the errors defined in an appropriate reference frame. In spite of the large variety of path following methods described in the literature we show that, in principle, most of them can be categorized in two groups: stabilization of the path following error system expressed either in the vehicle's body frame or in a frame attached to a \emph{``reference point"} moving along the path, such as a Frenet–Serret (F-S) frame or a Parallel Transport (P-T) frame. With this observation, we provide a unified formulation that is simple but general enough to cover many methods available in the literature. We then discuss the advantages and disadvantages of each method, comparing them from the design and implementation standpoint. We further show experimental results of the path following methods obtained from field trials testing with under-actuated and fully-actuated autonomous marine vehicles. In addition, we introduce open-source Matlab and Gazebo/ROS simulation toolboxes that are helpful in testing path following methods prior to their integration in the combined guidance, navigation, and control systems of autonomous vehicles.  
\end{abstract}


\input{INTRODUCTION.tex}
\input{PROBLEM_FORMULATION.tex}

\input{GUIDANCE_IN_2D.tex}
\input{SIMULATIONS.tex}
\input{EXPERIMENTS.tex}

\input{DISCUSSION.tex}
\input{CONCLUSION.tex}

\section*{Disclosure statement}
 The authors confirm that there are no known conflicts of interest associated with this publication.
 
 \section*{Acknowledgment}
 This work was partially funded by Fundação para a Ciência e a Tecnologia (FCT) and FEDER
 funds through the projects UIDB/50009/2020 and LISBOA-01-0145-FEDER-031411, the H2020-FETPROACT-
 2020-2 RAMONES project -Radioactivity Monitoring in Ocean Ecosystems (Grant agreement ID: 101017808) and
 the H2020-MSCA-RISE-2018 ECOBOTICS.SEA project -Bio-inspired Technologies for a Sustainable Marine Ecosystem
 (Grant agreement ID: 824043).

\bibliographystyle{alpha}
\bibliography{ALLreference.bib}{}

\input{APPENDIX.tex}
\end{document}

%% file: H_mathdefs_interactive.tex
\newcommand{\abs}[1]{|#1|}														
\newcommand{\Enorm}[1]{{ \left\| {#1} \right\|}}						        
\newcommand{\bs}[1]{\boldsymbol{#1}}										    

\renewcommand{\x}{{\bf x}}

\newcommand{\N}{\mathbb{N}}

\newcommand{\B}{\mathcal{B}}							
\newcommand{\I}{\mathcal{I}}							
\newcommand{\Iframe}{\left\{ \mathcal{I} \right\}} 	
\newcommand{\Bframe}{\left\{ \mathcal{B} \right\}} 	
\newcommand{\xI}{x_{\I}}							    
\newcommand{\yI}{y_{\I}}							    

\newcommand{\p}{{\bf p}}
\newcommand{\pd}{{\bf p}_{\rm d}}

\newcommand{\vcy}{v_{\rm cy}}

\newcommand{\vd}{v_{\rm d}}
\newcommand{\xbar}{{\bar{\bf x}}}

\newcommand{\e}{{\bf e}}
\newcommand{\f}{ {\bf f} }
\renewcommand{\u}{{\bf u}}
\newcommand{\ubar}{\bar{\bf u}}

\newcommand{\U}{\mathbb{U}}

\newcommand{\vc}{{\bf v}_{\rm c}}
\newcommand{\Tp}{T_{\rm p}}

\newcommand{\xd}{x_{\rm d}}
\newcommand{\yd}{y_{\rm d}}

\newcommand{\minus}{\text{-}}

\newcommand{\R}{\mathbb{R}}

\newcommand{\rmT}{{\rm T}}
\newcommand{\subscript}[1]{\scriptscriptstyle{#1}}
\newcommand{\eB}{\e_{\subscript{\B}}}
\newcommand{\eP}{\e_{\subscript{\mathcal{P}}}}
\renewcommand{\P}{\mathcal{P}}
\newcommand{\RPI}{R^{\subscript{\mathcal{I}}}_{\subscript{\mathcal{P}}}}
\newcommand{\RIP}{R^{\subscript{\mathcal{P}}}_{\subscript{\mathcal{I}}}}
\newcommand{\RBI}{R^{\subscript{\mathcal{I}}}_{\subscript{\mathcal{B}}}}
\newcommand{\RIB}{R^{\subscript{\mathcal{B}}}_{\subscript{\mathcal{I}}}}
\newcommand{\rmB}{\subscript{\mathcal{B}}}
\newcommand{\rmP}{\subscript{\mathcal{P}}}

%% file: format_interactive.tex
\usepackage[utf8x]{inputenc}
\usepackage{relsize,enumerate}
\usepackage{amsmath,amsfonts} 
\usepackage{siunitx}
\usepackage{amstext}
\usepackage{amssymb}

\numberwithin{figure}{section}
\usepackage{graphicx}
\usepackage{caption}
\usepackage{float}
\usepackage{epstopdf}
\usepackage{cite}
\usepackage{upgreek}

\usepackage{algorithmicx}
\usepackage{algorithm}
\usepackage{algpseudocode}
\newtheorem{condition}{Condition}

%% file: INTRODUCTION.tex
\section{Introduction}
Path-following (PF) is one of the most fundamental tasks to be executed by autonomous vehicles. It consists of driving a vehicle to and maintaining it on a pre-defined path while tracking a path-dependent speed profile. Unlike trajectory tracking, the path is not parameterized by time but rather by any other useful parameter that in some cases may be the path length. Thus, there is more flexibility in making the vehicle first converge to the path smoothly then move along it while tracking a given speed assignment. Path following is useful in many applications where the main objective is to accurately traverse the path, while maintaining a certain speed is a secondary task. Stated in simple terms, it is not required for the vehicle to be at specific positions at specific instants of time, a strong requirement in trajectory tracking. All that is required is for the vehicle to go through specific points in space while trying to meet speed assignments, but absolute time is not of overriding importance. From a technical standpoint, when compared with trajectory tracking, path following has the potential to exhibit smoother convergence properties and reduced actuator activity \cite{aguiar2007trajectory}.
For these reasons, in a vast number of applications path-following is performed by a variety of heterogeneous vehicles, of which marine vehicles \cite{FranciscoLBC2019,fossen2011handbook,Dula2020}, ground vehicles such as Mars rovers \cite{helmick2004path,Fox1997,Kanjanawanishkul2009}, fixed wing aerial vehicles \cite{Nelson2007,Park2007,sujit2014unmanned}, autonomous cars \cite{GUO2019,snider2009automatic,de1998feedback,Rokonuzzaman}, and quad-rotors \cite{rubi2019survey} are representative examples. \\
The task of deriving control strategies to solve the PF problem is technically challenging, specially in the presence of non-holonomic constraints, and often involves the use of a nonlinear control techniques such as backstepping \cite{Encarnacao2000,lapierre2006nonsingular}, feedback linearization \cite{akhtar2012path}, sliding mode control \cite{Dagci2023}, vector field \cite{Nelson2007,Wilhelm2019}, linear model predictive control (MPC) \cite{Raffo2010,Kanjanawanishkul2009}, and nonlinear MPC (NMPC) \cite{HungJRNC2019,alessandretti2013trajectory,GUO2019,SHIBATA2018}. Other path following algorithms exploit learning-based methods such as Learning-based MPC (LB-MPC) \cite{Ostafew2016,Rokonuzzaman2020}, reinforcement learning-based control \cite{MARTINSEN2018,Kamran2019,Wang2021}, among others. Due to the proliferation of robotic vehicles and their applications, the past decades have witnessed the development of a multitude of path-following methods. Therefore, in order to provide a critical assessment of the collective work done, this paper contains a comprehensive survey, aimed at comparing different PF methods from the design and implementation standpoint and discussing the advantages and disadvantages of each method. In particular, we show that an important classification of path-following algorithms is given by the choice of reference frame in which the path-following error is defined. In general, the latter is defined either in the vehicle's body frame or in a frame attached to a \emph{``reference point"} moving along the path such as the Frenet–Serret (F-S) frame or the Parallel Transport (P-T) frame. \\
In the literature, one can find other survey papers of path following methods such as \cite{sujit2014unmanned} for fixed-wing aerial vehicles, \cite{rubi2019survey} for quadrotors, or autonomous car-like robot \cite{Rokonuzzaman}. However, the aforementioned survey papers do not describe in detail the theory that supports the methods described and, while they contain simulation results, they do not present a comparison of the performance of the methods in field tests with real vehicles.
Moreover, the above surveys focus on specific types of autonomous vehicles, and do not consider some of the unique characteristics of under-actuated marine vehicles such as non-holonomic constraints in \cite{rubi2019survey} or the requirement that a vehicle track a desired speed profile along the path \cite{sujit2014unmanned,Rokonuzzaman}. 
In the current review paper, instead of focusing on a particular type of vehicle, we emphasize the common principle underlying path following methods, that can be applied and extended to a large class of vehicles, the simplified motion of which can be described by the same class of kinematic models. Furthermore, the present paper also provides a rigorous theoretical proof of the methods reviewed that is absent in the previous surveys. We introduce simulation toolboxes written in Matlab and ROS/Gazebo that are helpful in testing the path following methods and integrating them in guidance, navigation, and control systems. To conclude, we report experimental results with a Medusa class autonomous marine vehicles \cite{abreu2016medusa} that have been widely used in EU projects such as WiMust \cite{PedroWimust} and MORPH \cite{PedroMorph}. In short, the main contributions of this paper include  
\begin{enumerate}[(i)]
	\item An in-depth review of standard path-following methods in two dimensional space (2D) explaining in detail the theoretical principles of the different methods.
	\item A discussion of the advantages and disadvantages of each method, comparing them from the design and implementation standpoint.
	\item A Matlab simulation toolbox and ROS/Gazebo simulation packages of path-following methods.
	\item A description of experimental results of field trials at sea with the Medusa under-actuated and fully-actuated robotic vehicles, followed by an assessment of the performance obtained in real-life situations.
\end{enumerate} 
The paper is organized as follows. The path following problem is formulated in Section \ref{sec: problem}. Section \ref{sec: pf method in 2D} describes path following methods for under-actuated vehicles. Section \ref{section: extension with unknown disturbance} extends the path following methods to the case where external unknown disturbances occur. Section \ref{sec: fully actuated} reviews a path following method developed for  fully-actuated vehicles with arbitrary heading. The implementation in Matlab and Gazebo/ROS
simulation toolboxes for testing path following methods are introduced in Section \ref{sec:Implementation}. Experimental results with autonomous marine vehicles are presented in Section \ref{sec: experiement}. Section \ref{sec: discussion} provides a discussion on advantages and disadvantages of the path following methods and practical issues when the vehicles dynamics are taken into account. Finally, Section \ref{sec: conclusion} contains the main conclusions.  

%% file: PROBLEM_FORMULATION.tex
\section{Problem formulation and common principles of path following methods} \label{sec: problem}
\subsection{Vehicle kinematic models}
\begin{figure}[h]
	\centering
	\includegraphics[width=100mm]{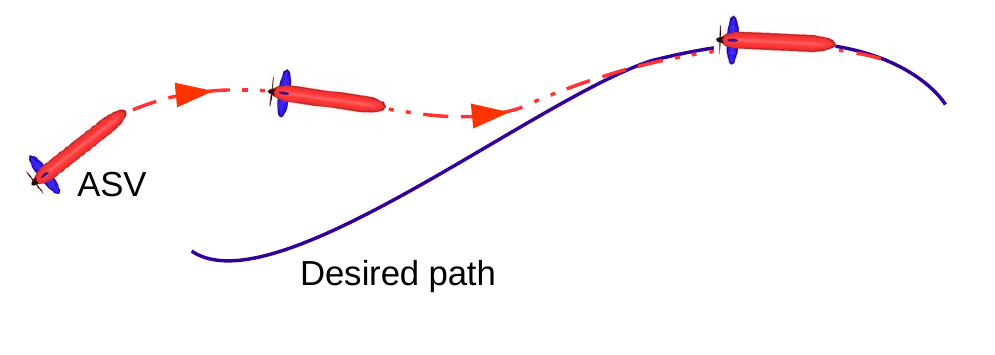} 
	\caption{Geometric illustration of the path following problem. }
	\label{fig: 3D presentation}
\end{figure}
\begin{figure}[h]
	\centering
	\includegraphics[width=120mm]{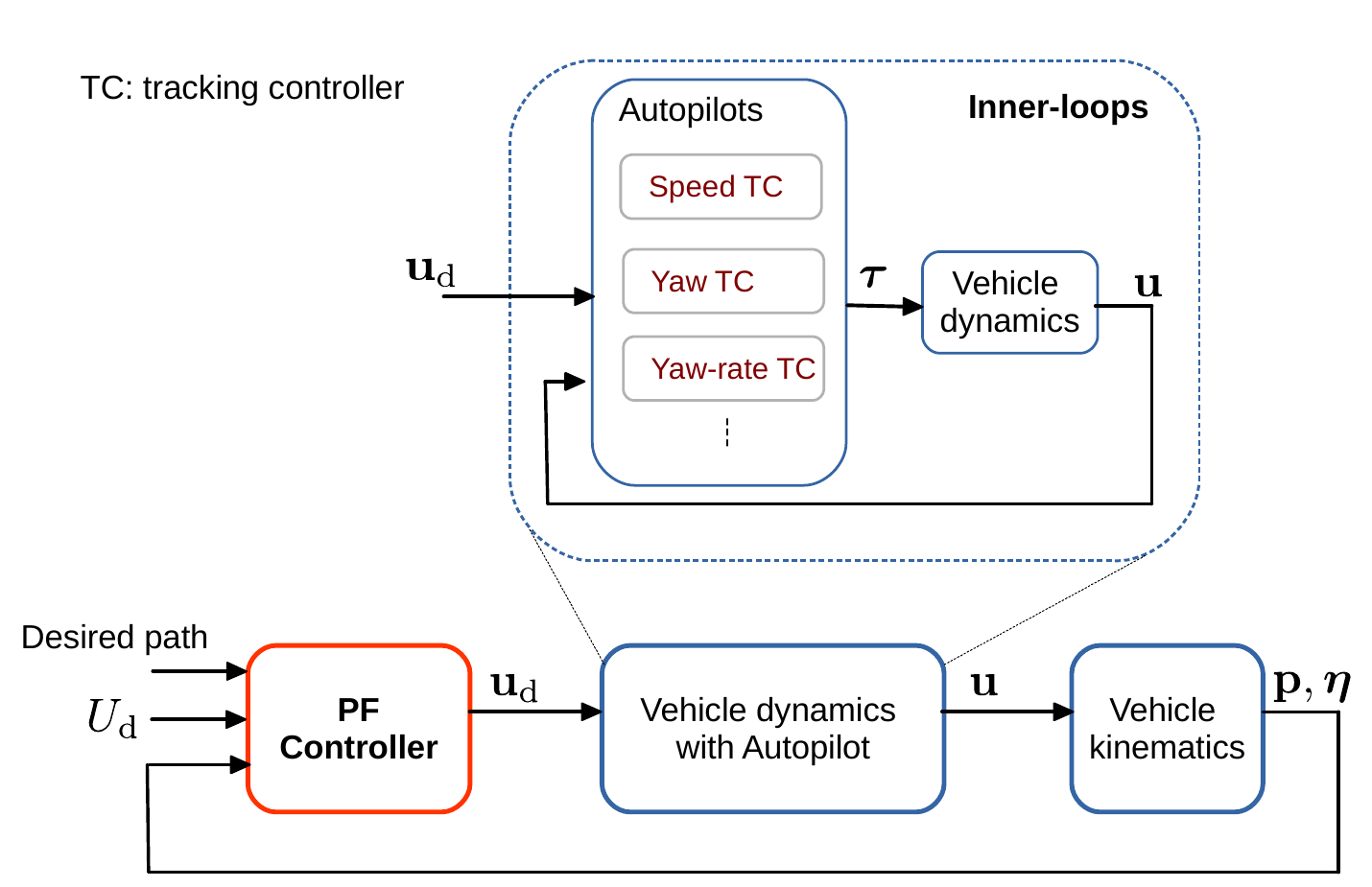} \\
	a) A complete path following control system ~\\~\\
	\includegraphics[width=120mm]{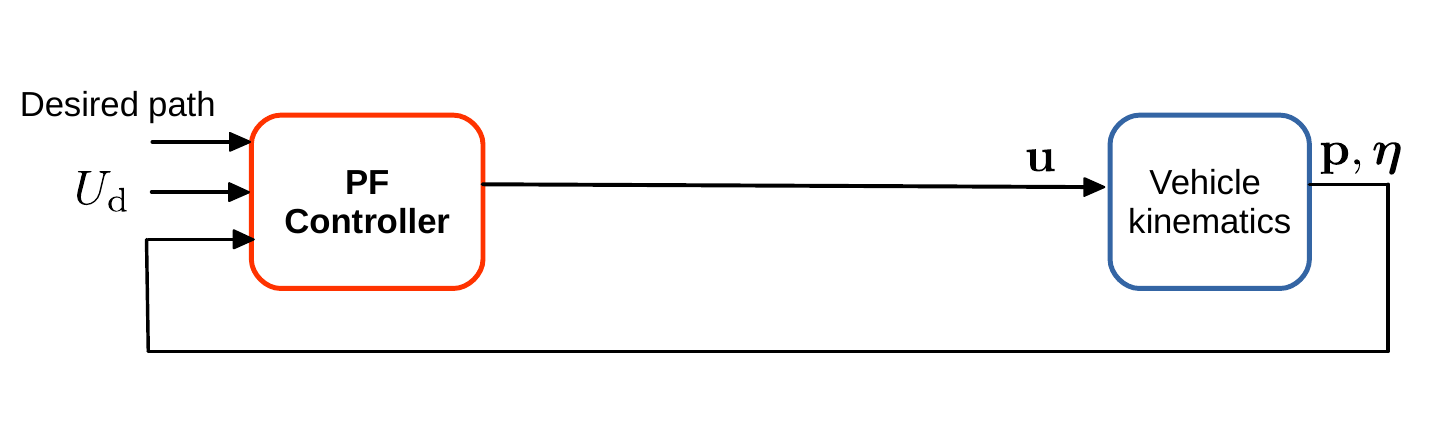} \\
	b) A simplified path following system for PF controller design
	\caption{Path following control systems; ${\bf u}_{\rm d}$: reference inputs (e.g. desired linear and angular speeds, orientations) for the autopilots;  $\p$ and $\bs{\eta}$: the vehicle's position and orientation, respectively,  ${\boldsymbol \tau}$: force and torque, $U_{\rm d}$: desired speed profile that the vehicle must track. }
	\label{fig: path following system}
\end{figure}

Path following refers to the problem of making a vehicle converge to and follow a spatial path while asymptotically tracking a desired speed profile along the path; see Fig. \ref{fig: 3D presentation} that shows an ASV executing a path following maneuver. From a control system standpoint, the structure of a complete path following system is captured in Fig. \ref{fig: path following system}(a). In this architecture, the outer-loop path following controller implements a guidance strategy, in charge of computing desired references (e.g. linear and angular speeds, or  orientations) to steer  the vehicle along the path with a desired speed profile $U_d$. These references act as inputs to autopilots that play the role of inner-loop controllers, in charge of generating suitable forces and torque for the vehicle in order to track the desired references, thus achieving the path following objectives. \\
In practice, to simplify the design of a path following controller, it is commonly assumed that the responses the inner-loops are sufficiently fast so that the influence of the latter in the complete system can be neglected. Under these ideal conditions, the path following control system can be simplified by considering the kinematics model only, as shown in Fig. \ref{fig: path following system}(b). Our purpose in the present paper is to review the core ideas behind existing path following methods in the literature, therefore we primarily focus on those designed for the vehicle kinematics only. We then treat the effect of the inner-loops as an internal disturbances and analyze the robustness of the path following methods under these disturbances accordingly. \\
\begin{figure}[h]
	\centering
	\includegraphics[width=100mm]{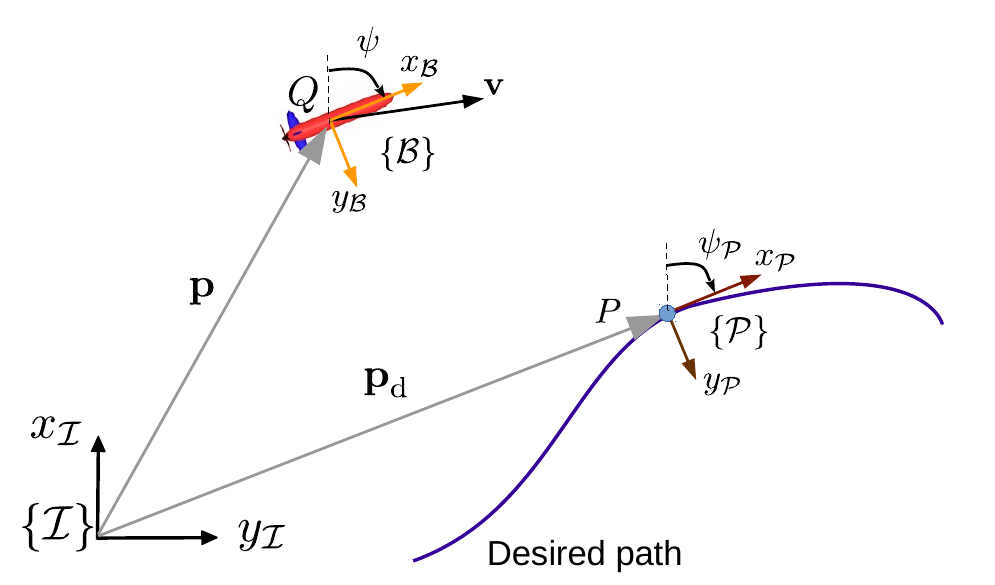} 
	\caption{Geometric illustration of the path following problem. $\{I\}, \{\B\}$, and $\{\P\}$ denote inertial/global, the vehicle body, and path frames, respectively. The symbols $\p$ and $\p_{\rm d}$ represent the position vectors of the vehicle and a generic point on the path, respectively expressed in $\{\I\}$. }
	\label{fig: 2D illustration with frame}
\end{figure}
The vehicle kinematic models that will be used to derive path following control laws in the present paper will be described next. The vehicle's motion with respect to a path is illustrated Fig.\ref{fig: 2D illustration with frame}. The following notation and nomenclature will be used. The symbol $\{\I\} = \{x_\I, y_\I\}$ denotes an inertial (global) North-East (NE) frame, where the axis $\xI$ points to the North and the axis $\yI$ points to the East. Let $Q$ be the center of mass of the vehicle and denote by ${\bf p}=[x, y]^{\top} \in \R^{2}$ the position of $Q$ in  $\Iframe$. Let also $\{\B\} = \{x_\B, y_\B\}$ be a body-fixed frame whose origin is located at $Q$. In addition, denote by ${\bf v}=[u,v]^{\top} \in \R^{2}$ the vehicle's velocity vector with respect to the fluid, measured in $\Bframe$, where $u,v$ are the surge/longitudinal and sway/lateral speeds, respectively. With the above notation, the \emph{``general"} 3-DOF vehicle's kinematic model is described by 
\begin{equation} \label{eq: kinematics in 2D general form}
\begin{split}
\dot{x}&=u\cos(\psi) - v\sin(\psi) + v_{\rm{cx}} \\
\dot{y}&=u\sin(\psi) + v\cos(\psi) + v_{\rm{cy}}				\\
\dot{\psi}&=r,				
\end{split}
\end{equation}   
where $\psi$ is the vehicle's heading/yaw angle and $r$ is its heading/yaw rate, $v_{\rm cx}$ and $v_{\rm cy}$ are components of vector ${\bf v}_{c} = [v_{\rm cx},v_{\rm cy}]^{\top}\in \R^2$ that represents the effect of external unknown disturbances (e.g. ocean current in the case of marine vehicles and wind in the case of aerial vehicles) in $\Iframe$. For the sake of clarity, in the present paper we consider the following three separate subcases of \eqref{eq: kinematics in 2D general form}.  
\subsubsection{Scenario 1: under-actuated vehicle without external disturbances} 
In this scenario we assume that the vehicle is under-actuated and that the vehicle's lateral motion and external disturbances are so small so that they can be neglected, that is, making $v,v_{\rm cx}, v_{ \rm cy}$ zero we obtain   
\begin{equation} \label{eq: kinematics in 2D}
\begin{split}
\dot{x}&=u\cos(\psi) \\
\dot{y}&=u\sin(\psi)					\\
\dot{\psi}&=r.				
\end{split}
\end{equation}
   
\begin{figure}
	\centering
	\includegraphics[width=.47\textwidth]{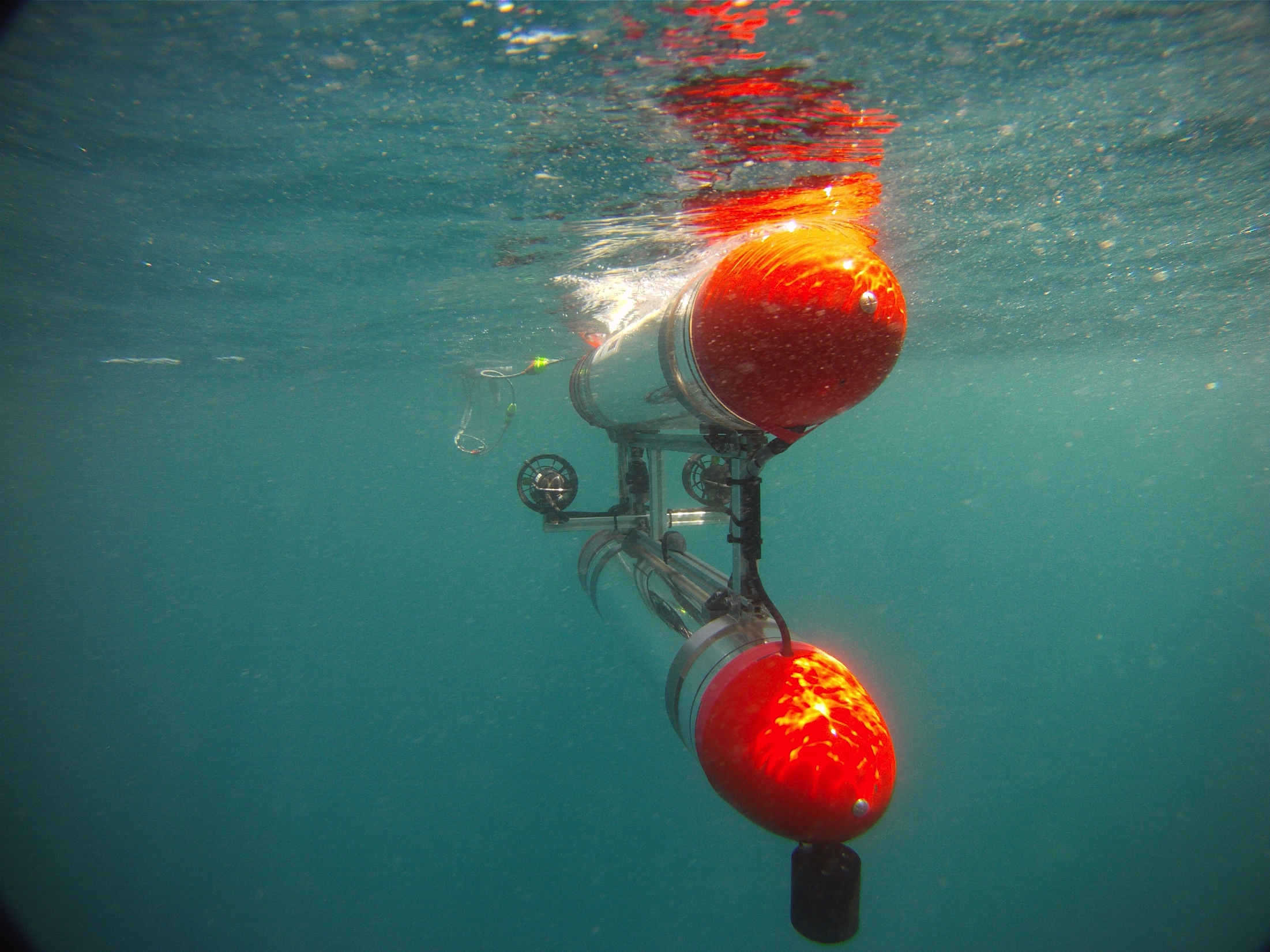}\hfill
	\includegraphics[width=.47\textwidth]{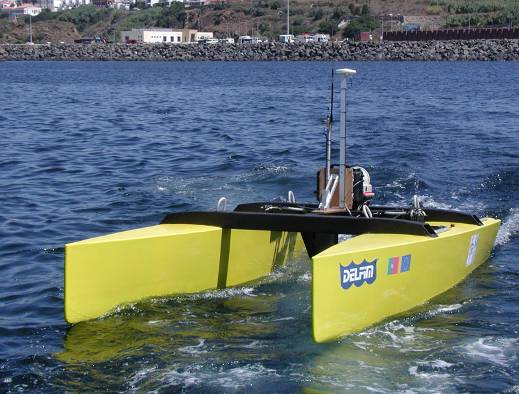} \vfill 
	a) Under-actuated Medusa \cite{abreu2016medusa}            \hspace{40pt}          b) Delfim \cite{DELFIM2006}
	\\[\bigskipamount]
	\includegraphics[width=.47\textwidth]{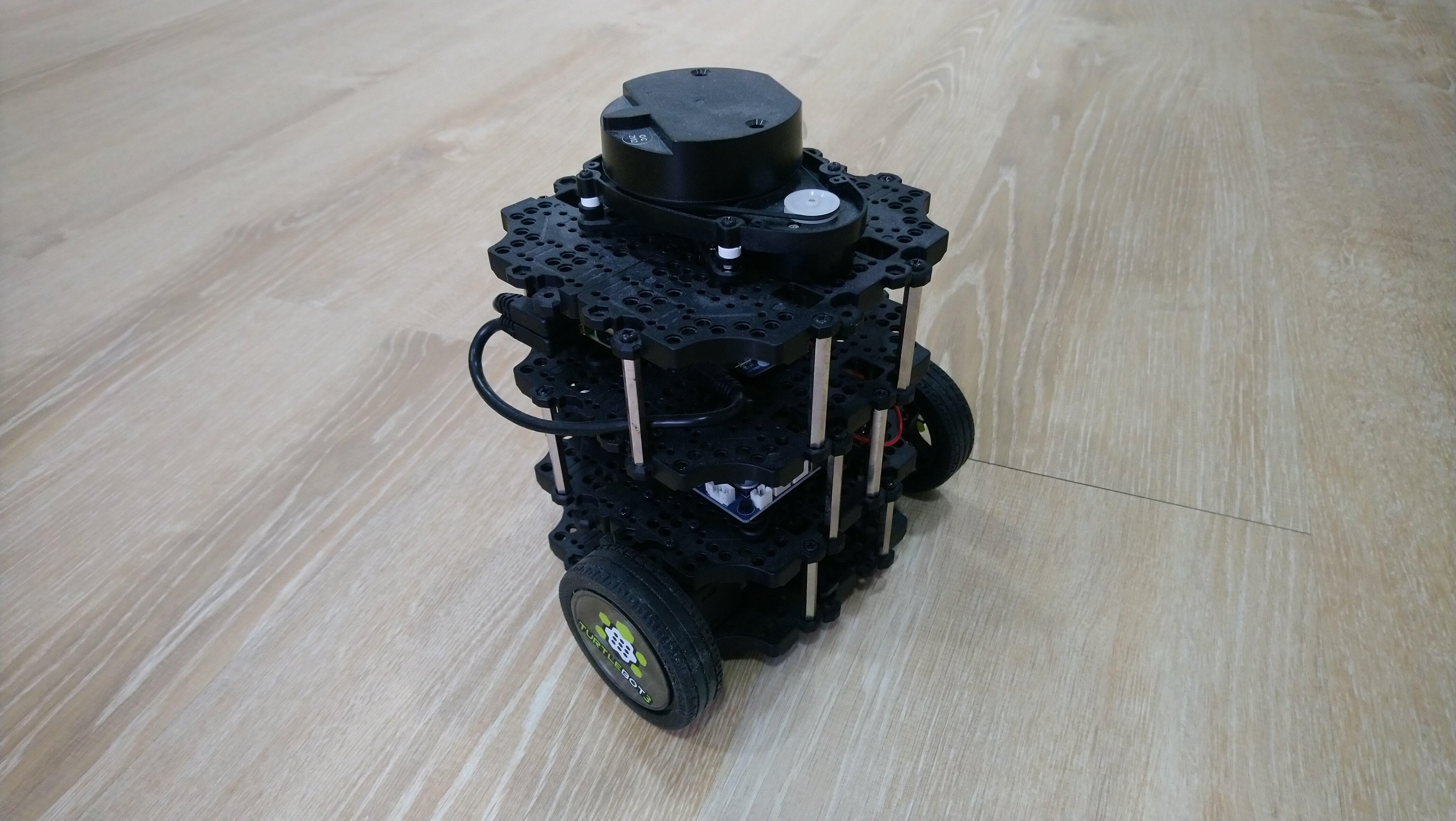}\hfill
	\includegraphics[width=.47\textwidth]{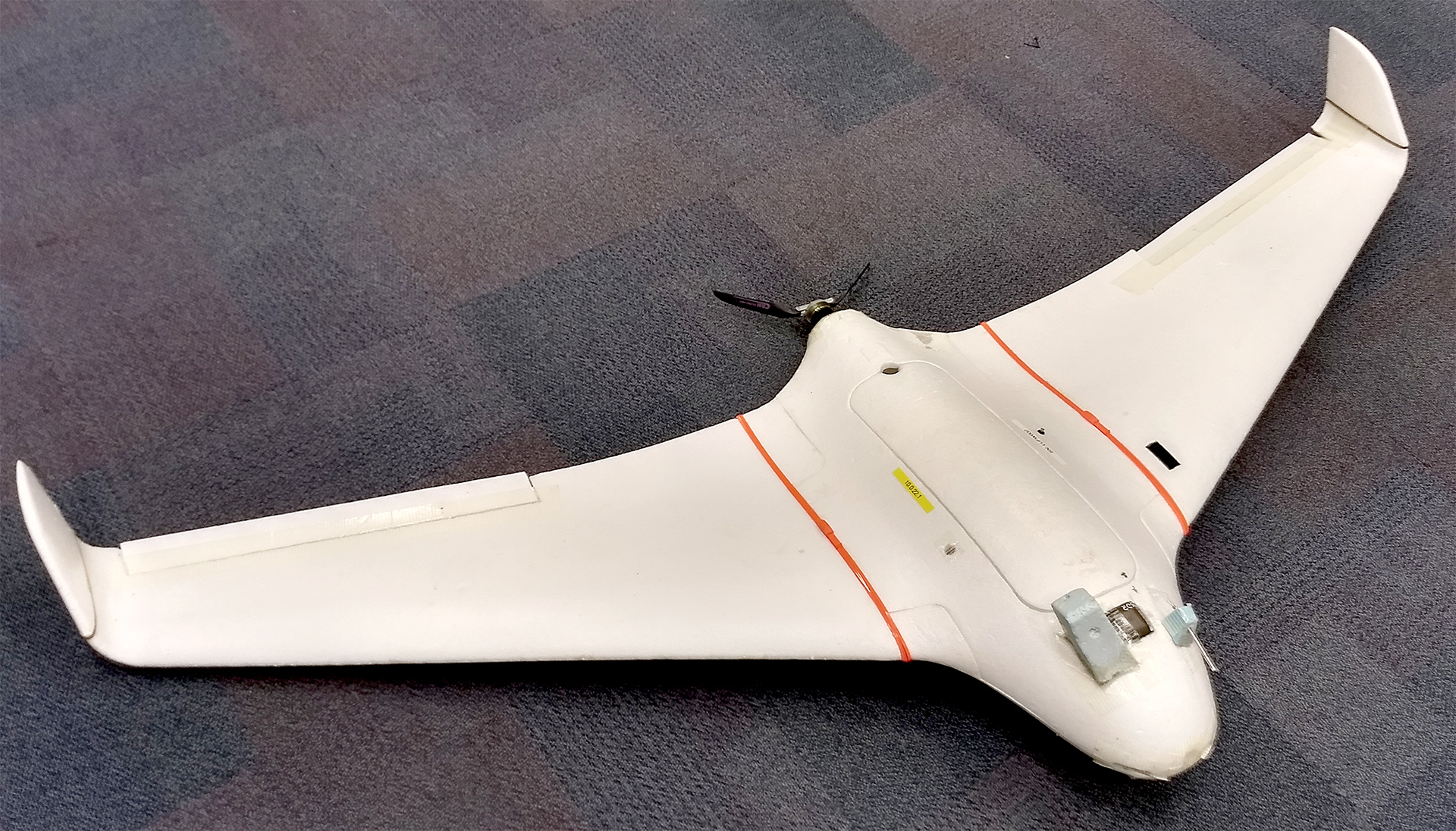}\hfill
	\vfill 
	c) Turtlebot Burger \cite{rosprogramming2017}          \hspace{70pt}          d) X8 fixed-wing UAV \cite{Yang2021}
	\caption{Under-actuated robotic vehicles}
	\label{fig: underactuated vehicles}
\end{figure}

We also assume that the longitudinal speed ($u$) and the heading ($\psi$) or heading rate ($r$) can be tracked with good accuracy by inner-loop controllers. Although \eqref{eq: kinematics in 2D} is a significant simplification of \eqref{eq: kinematics in 2D general form} it still captures sufficiently well the behavior of a large class of under-actuated vehicles, including unicycle mobile robots \cite{Samson1993,rosprogramming2017}, fixed-wing UAVs  undergoing planar motion \cite{Nelson2007,zhao2020cooperative,Liu2013,Yang2021,rucco2015model}, and a wide class of under-actuated autonomous marine vehicles (AMVs). The later include the Medusa and Delfim \cite{abreu2016medusa} and Charlie \cite{ROB:ROB20303} vehicles, for which the sway speed is in practice so small that it can be neglected. Snapshots of several vehicles mentioned above are shown in Fig.\ref{fig: underactuated vehicles}.
In the literature, most path following methods are proposed for the kinematic model \eqref{eq: kinematics in 2D}; in accordance, a large part of this paper presented in Section \ref{sec: pf method in 2D} is devoted to their review.

\subsubsection{Scenario 2: under-actuated vehicle with external disturbances} 
In this scenario the vehicle motion is influenced significantly by external disturbances that can not be neglected; however, the lateral sway is still small enough to be ignored (i.e. $v = 0$). Given these assumptions, the vehicle kinematics model is given by  
\begin{equation} \label{eq: kinematics in 2D no sway}
\begin{split}
\dot{x}&=u\cos(\psi) + v_{cx}\\
\dot{y}&=u\sin(\psi) + v_{cy}					\\
\dot{\psi}&=r.				
\end{split}
\end{equation}   
Note also that similar to \emph{Scenario 1}, the vehicle is under-actuated. Path following methods developed for this model will be discussed in Section \ref{section: extension with unknown disturbance}.
\subsubsection{Scenario 3: fully or over-actuated vehicle}
\begin{figure}
	\centering
	\includegraphics[width=.45\textwidth]{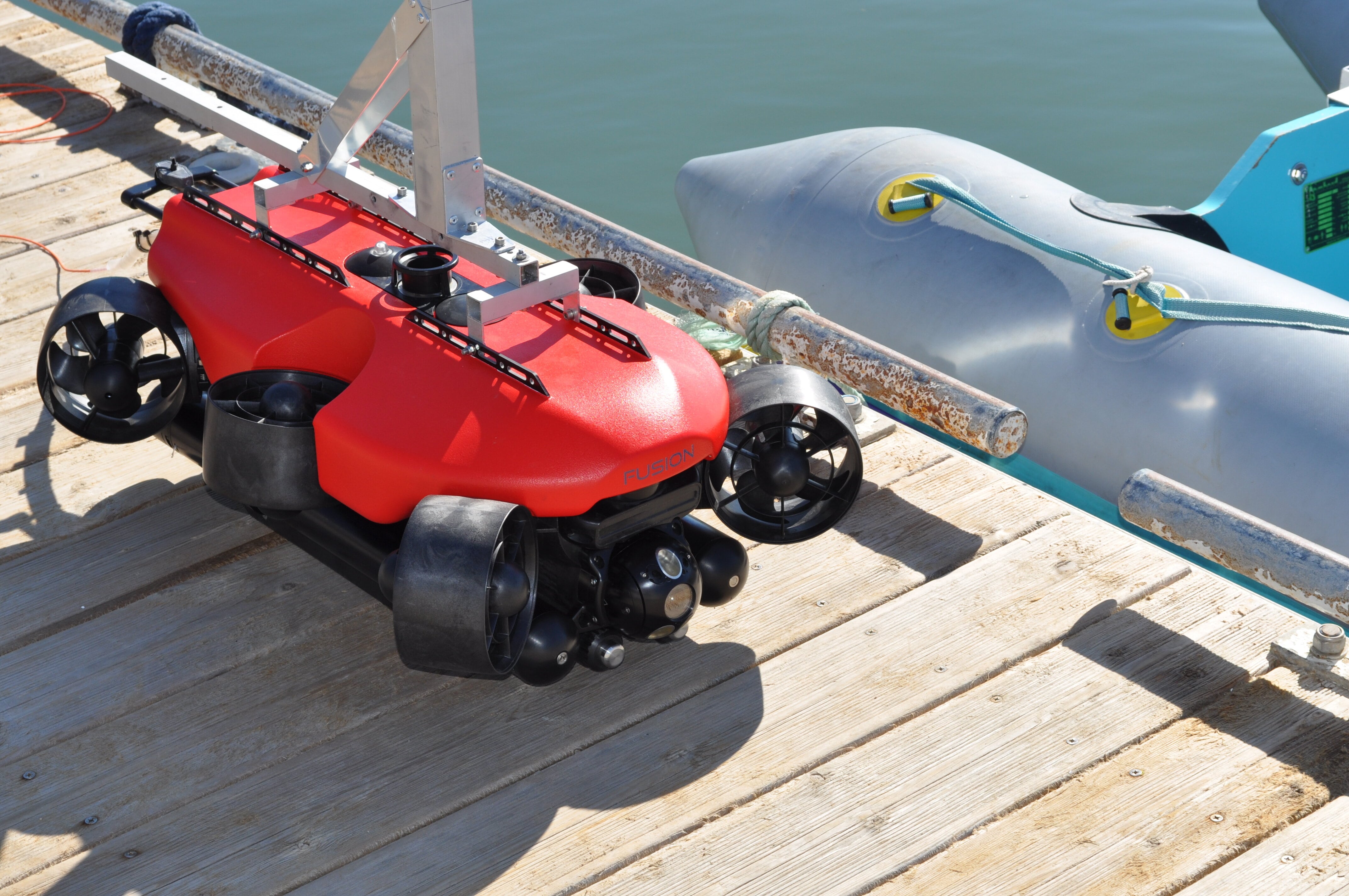} \hfill
		\includegraphics[width=.45\textwidth]{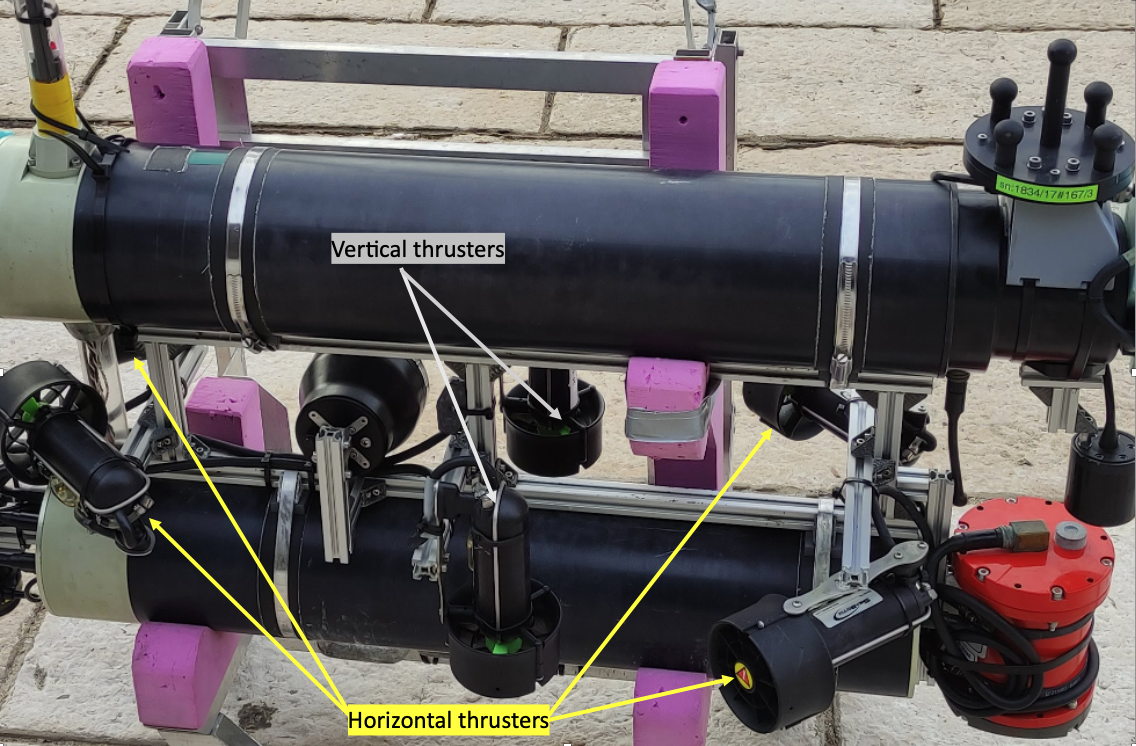} \vfill
	\caption{Over-actuated robotic vehicles. Left: Fusion vehicle, produced by Strategic Robot Systems company equipped with the Guidance, Navigation, and Control systems developed by IST Lisbon; Right: M-Vector vehicle, developed by IST Lisbon. Notice in both vehicles the existence of 4 thrusters in the horizontal plane (2 at the bow and 2 at the  stern, installed at slant angles), capable of imparting directly forces along the longitudinal and lateral axis and torque about the vertical axis. }
	\label{fig: fully actuated-vehicle}
\end{figure}

In this scenario we assume that the lateral speed is significant and can not be neglected. However, for simplicity we neglect the effect of external disturbances. With these assumptions, the vehicle kinematic model is given by 
\begin{equation} \label{eq: kinematics in 2D no current}
\begin{split}
\dot{x}&=u\cos(\psi) - v\sin(\psi)\\
\dot{y}&=u\sin(\psi) + v\cos(\psi)					\\
\dot{\psi}&=r.				
\end{split}
\end{equation}   
In this scenario we assume further that the vehicle is fully or over-actuated in that its longitudinal and lateral speeds and heading rate can be controlled simultaneously. Fig.\ref{fig: fully actuated-vehicle} shows snapshots of several fully-actuated vehicles that meet these assumptions. For this scenario we are interested the path following problem in which the vehicle is not only required to follow a predefined path but also to maneuver such that its heading tracks an arbitrary heading reference. This scenario will be presented in Section \ref{sec: fully actuated}.
\subsection{Path parameterization and path frames} \label{section: path frames}
Let $\mathcal{P}$ be a spatial path defined in an inertial frame and parametrized by a scalar variable $\gamma$ (eg. arc-length of the path). Normally, $\gamma \in \Omega:= [a, b]$ where $a,b \in \R$ are values of $\gamma$ corresponding to the points at beginning and end of the path. The position of a generic point $P$ on the path in the inertial frame $\{\I\}$
 is described by vector
\begin{equation} \label{eq: path in 2D}
\pd(\gamma)=[x_{\rm d}(\gamma),y_{\rm d}(\gamma)]^{\top} \in \R^2.
\end{equation}
 At $P$, there are two frames adopted in the literature to formulate the path following problem, that is, to describe the position error between the vehicle and the path. Namely, the \emph{Frenet–Serret} (F-S) and the \emph{Parallel Transport} (P-T) frames that are described next.  
 \subsubsection{Frenet–Serret frame, \cite{gray2006modern}}
The F-S frame is used in the path following methods described in \cite{Samson1993,Lapierre2003,Isaac07}. A detailed description of this frame for 3D curves can be found in \cite{gray2006modern}. In the present paper, because we only consider path following in 2D we simplify the frame for 2D curves as follows, see Fig.\ref{fig: S-F and P-T frame presentation}. Formally, let
 \begin{equation} \label{eq: frenet frame}
 {\bf t}(\gamma)=\frac{\pd'(\gamma)}{\Enorm{\pd'(\gamma)}}, \quad {\bf n}(\gamma)=\frac{{\bf t}'(\gamma)}{\Enorm{{\bf t} '(\gamma)}}
 \end{equation}
 be the basis vectors defining the F-S frame at the point $\pd(\gamma)$ where, for every differentiable $\f(x)$, $\f'(x)\triangleq \partial \f(x)/\partial x$. These vectors define the unit tangent and \emph{principle} unit normal respectively to the path at the point determined
 by $\gamma$. The curvature $\kappa(\gamma)$ of the path at that point is given by 
 \begin{equation} \label{eq: curvature}
 \kappa(\gamma)=\Enorm{{\bf t}'(\gamma)}.
 \end{equation} 
 As can be seen in the formula for computing the normal vector in \eqref{eq: frenet frame}, the main technical problem with the F-S frame is that it is not well-defined for paths that have a vanishing second derivative (i.e. zero curvature) such as straight lines or non-convex curves. The other alternative frame, called Parallel-Transport frame, overcomes this limitation and is presented next.
  \subsubsection{Parallel–Transport frame, \cite{Hanson95paralleltransport}}
\begin{figure}[h]
  	\centering
  	\includegraphics[width=100mm]{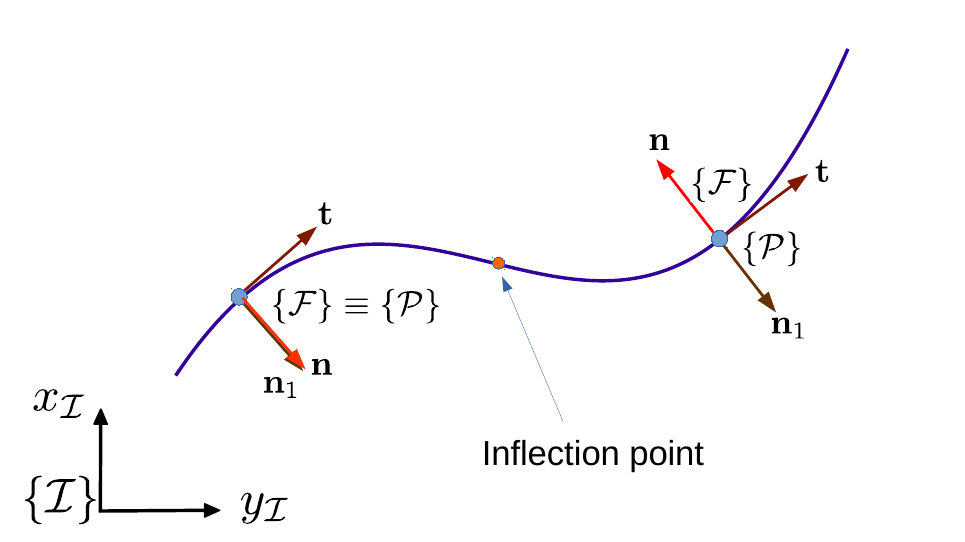} 
  	\caption{An illustration of Frenet-Serret $\{\mathcal{F}\}$ and Parallel Transport $\{\mathcal{P}\}$ frames in 2D. Notice how at the inflection point the normal vector $\bf n$ of the Frenet-Serret frame inverts its direction.}
  	\label{fig: S-F and P-T frame presentation}
  \end{figure}

  This frame was introduced in \cite{Hanson95paralleltransport} and used for the first time in the path following method of \cite{Isaac2010}. The P-T frame is based on the observation that, while the tangent vector for a given curve is unique, we may choose any convenient arbitrary normal vector so as to make it perpendicular to the tangent and vary smoothly throughout the path regardless of the curvature \cite{Hanson95paralleltransport}. \\
  In 2D, a simple way to define the P-T frame is as follows. First, specify the tangent basic vector {\bf t} as in \eqref{eq: frenet frame}. The second basic vector, called normal vector ${\bf n}_1$, is obtained by rotating the tangent vector $90$ degree clockwise. This, as shown in \cite{Breivik2005}, is equivalent to translating $\Iframe$ to the ``\textit{reference point}" $P$ and then rotating it about the z-axis by the angle 
  \begin{equation}
  \psi_{\rmP} = \arctan\left(\frac{y'_{\rm d}(\gamma)}{x'_{\rm d}(\gamma)}\right).
  \end{equation}       
 The difference between the F-S frame and the P-T frame is illustrated in Fig. \ref{fig: S-F and P-T frame presentation}. With the F-S frame the normal component always points to the center of curvature thus, its direction switches at inflection points, while the P-T frame has no such discontinuities. From a path following formulation standpoint, with the F-S frame, the path following error is not well-defined at inflection points because the cross-track error (the position error projected on the normal vector) switches sign, which is not the case with the P-T frame.    
\begin{remark} 
	Another way of propagating a P-T frame along the path is to use the algorithm proposed in \cite{Hanson95paralleltransport}. While this algorithm is general and efficient for 3D, it is unnecessarily complicated for 2D curves. 
\end{remark}

\subsection{Path following formulation}
 \label{section: path following problem statement}
With the concepts and notation described above, the path following problem is stated next, see also Fig.\ref{fig: 2D illustration with frame}.\\
{\bf Path following problem in 2D}: \textit{Given the 2-D spatial path $\mathcal{P}$ described by \eqref{eq: path in 2D} and a vehicle with the kinematics model described by  \eqref{eq: kinematics in 2D}, derive a feedback control law for the vehicle's inputs $(u,r)$ and possibly for $\dot{\gamma}$ or $\ddot{\gamma}$ so as to fulfill the following tasks:
		\begin{enumerate}[i)]
			\item Geometric task: steer the position error $\e \triangleq \p - \p_{\rm d}$ s.t. 
\begin{equation} \label{eq: gemetric task}
\lim_{t\to \infty} \e(t)={\bf 0},
\end{equation}
where $\p_{\rm d}$ is the inertial position of a \emph{``reference point"} $P$ on the path, the temporal evolution of which can be chosen in a number of ways as discussed later. 
			\item Dynamic task: ensure that the vehicle's forward speed tracks a positive desired speed profile $U_{\rm d}=U_{\rm d}(\gamma,t)$, i.e. 
			\begin{equation} \label{eq: dynamic task}
         \lim_{t\to \infty} u(t)- U_{\rm d}(t) = 0.
			\end{equation}	
		\end{enumerate}
}
In what follows, let $u_{\rmP}$ be the speed of the \emph{``reference point"} $P$ with respect to the inertial frame, that is,
\begin{equation} \label{eq: uP}
u_{\rmP}=\frac{\rm d{\p}_{\rm d}}{{\rm d} t}=\Enorm{\p'_{\rm d}(\gamma) }\dot{\gamma}.
\end{equation}
Should the vehicle achieve precise path following, both the vehicle and the point $P$ will move with the desired speed profile $U_{\rm d}$, i.e. $u=u_{\rmP}=U_{\rm d}$. In this case the dynamics task in \eqref{eq: dynamic task} is equivalent to requiring 
\begin{equation} \label{eq: dynamic task for gamma}
\lim_{t\to \infty} \dot{\gamma}(t)- v_{\rm d}(\gamma,t) = 0,
\end{equation}	
where $v_{\rm d}$ is the desired speed profile for $\dot{\gamma}$, defined by
\begin{equation} \label{eq: vd}
v_{\rm d}(\gamma,t) \triangleq \frac{1}{\Enorm{\p'_{\rm d}(\gamma)}}U_{\rm d}(\gamma,t).
\end{equation}
In path following, the point $P$ on the path plays the role of a \emph{``reference point"} for the vehicle to track. This point can be chosen as the nearest point to the vehicle \cite{Samson1993}, i.e. the orthogonal projection of the vehicle on the path (in case it is well defined), or can be initialized arbitrarily anywhere on the path with its evolution controlled through $\dot{\gamma}$ \cite{Lapierre2003,Breivik2005} or $\ddot{\gamma}$ \cite{aguiar2007trajectory} to achieve the path following objectives. In the latter cases, $\dot{\gamma}$ or $\ddot{\gamma}$ are considered as the controlled input of the dynamics of the \emph{``reference point"}, affording an extra freedom in the design of path following controllers.
\subsection{Common principles of path following methods}
Although there are a variety of path following methods described in the literature, most of them can be categorized as in Table \ref{tab: principle of path following method}. The first category includes the methods that aimed to stabilize the position error in a frame attached to a \emph{reference point} point moving along the path (e.g. F-S or P-T frames), whereas the second category consists of the methods that aim to stabilize the position error in the vehicle's body frame. The origin of the first method can be traced back to the work of \cite{Samson1993} addressing the path following problem of unicycle-type and two-steering-wheels mobile robots. The core idea in this work was then adopted to develop more advanced path following algorithms in \cite{Lapierre2003,YuMPC2015,HungCAMS2018}. Later, we shall see that Line-of-Sight (LOS), a well-known path following method and widely used in marine craft \cite{FOSSEN2003} can be categorized in this group as well. The second approach was proposed by \cite{aguiar2007trajectory} and further developed in \cite{alessandretti2013trajectory} to handle the vehicle's input and state constraints. 
\begin{table}[hbt!] 
	\centering
	\caption{Principles of path following methods }
	\vspace{-20pt}
		\begin{minipage}{\textwidth}
	{\begin{tabular*}{32pc}{@{\extracolsep{\fill}}clcc@{}}\\ \toprule
    &  \multirow{2}{*}{ ${\bf u}$}	 &\multicolumn{2}{c}{\bf Principles}\\ 		  \cmidrule{3-4}
			&   & Stabilizing  $\e$ in path frames	& Stabilizing $\e$ in the body frame \\	\midrule
	\multirow{2}{*}{\bf 2D}		&$u,\psi$	& \cite{PAPOULIAS1991,FOSSEN2003,Breivik2005,Pramod2009}								&    						\\ 
			&$u,r$	    &  	\cite{Samson1993,Lapierre2003} \cite{YuMPC2015,HungCAMS2018}							&   	\cite{aguiar2007trajectory}\cite{alessandretti2013trajectory}					\\ \midrule
	\end{tabular*}}{}
\end{minipage}
	\label{tab: principle of path following method}
\end{table}

%% file: GUIDANCE_IN_2D.tex
\section{Basic path following methods} \label{sec: pf method in 2D}
In this section we study path following methods for vehicles whose motion is described by the kinematic model \eqref{eq: kinematics in 2D}. In Section \ref{section: extension with unknown disturbance} we will extend the methods to the cases when the vehicle motion is subjected to external disturbances.   


\subsection{Methods based on stabilizing the path following error in the path frame} \label{section: methods in P}


In this section we present a \emph{``unified formulation"} that is simple but general enough to cover the path following methods in  \cite{Samson1993,Lapierre2003,lapierre2006nonsingular,Breivik2005,Fossen2015,HungCAMS2018}. The common principle behind these methods can be summarized in two steps: 
\begin{itemize}
	\item step 1: derive the dynamics of the path following error between the vehicle and the path in a path frame (e.g. F-S or P-T frame)
	\item step 2: drive these errors to zero using nonlinear control techniques to achieve path following, i.e. make the vehicle converge to and move along the path with a desired speed profile. 
\end{itemize}
 In order to have a \emph{unified formulation}, instead of using the F-S frame as in \cite{Samson1993,Lapierre2003,lapierre2006nonsingular} we use the P-T frame. The main advantage of the P-T frame in comparison with the F-S frame was discussed in Section \ref{section: path frames}, i.e. it avoids the singularity when the path has a vanishing second derivative, e.g. concave paths. Furthermore, the approach that we describe here is different from the one in \cite{Samson1993,Lapierre2003,lapierre2006nonsingular} in that the formulation in this section applies to any path that is not necessarily parameterized by the arc-length. \\
  \begin{figure}[h]
	\centering
	\includegraphics[width=100mm]{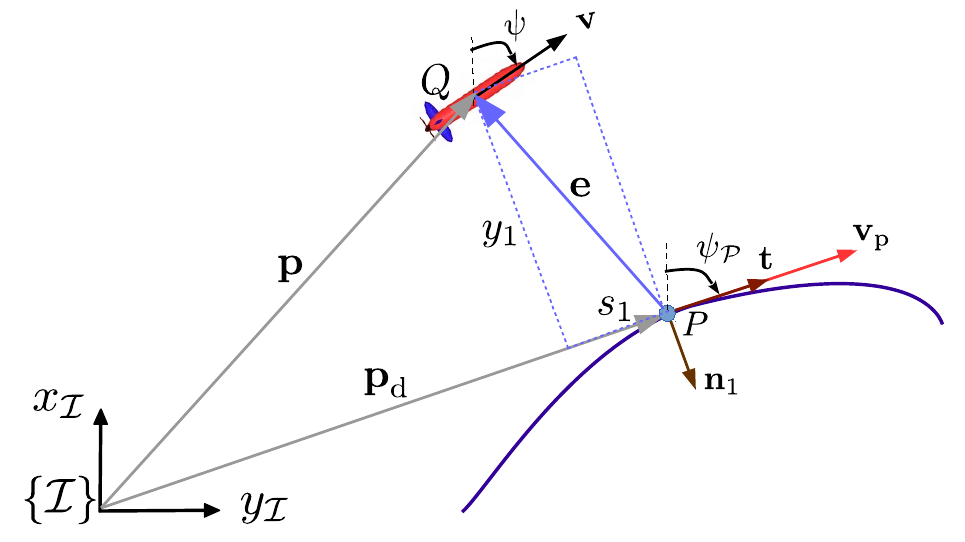} \\
	
	\caption{A geometric illustration of the methods in Section \ref{section: methods in P}. $P$ is the ``\emph{reference point}" that the vehicle must track to achieve path following. }
	\label{fig: Geometry illustration Path Following Method 1}
\end{figure}
\subsection{Derivation of the path following error} 
\label{section: derivation of path following error in path frame}
We now derive the dynamics of the path following errors (position and possibly orientation errors) between the vehicle and the path to be stabilized in order to achieve path following.  The formulation is inspired by the work in \cite{Samson1993,Lapierre2003,Breivik2005} and is presented next. Let $P$ be a point moving along the path that plays the role of a ``\emph{reference point}" for the vehicle to track so as to achieve path following. Let $\{\P\}$ be the P-T frame attached to this point defined by rotating the inertial frame by angle $\psi_{\rmP}$, where $\psi_{\rmP}$ is the angle that the tangent vector at $P$ makes with $\xI$; see Fig.	\ref{fig: Geometry illustration Path Following Method 1}. Let $\eP\triangleq[s_1,y_1]^{\top} \in \R^{2} $ be a vector defining the position error  between the vehicle and the \emph{referene point} $P$, where $s_1$ and $y_1$ are called \emph{along-track} and \emph{cross-track} errors, respectively. This vector can be viewed as the position vector of the vehicle expressed in $\{\P\}$. According to this definition, it is given by
\begin{equation} \label{eq: eP}
\eP=\RIP(\psi_{\rmP})(\p - \pd),
\end{equation}
where $\p=[x,y]\in\R^2$ is the position of the vehicle, $\pd$ is given by \eqref{eq: path in 2D} and $\RIP \in SO(2)$ is the rotation matrix from $\Iframe$ to $\{\P\}$, defined as
\begin{equation} \label{eq: RIP}
\RIP(\psi_{\rmP})=\begin{bmatrix}
\cos(\psi_{\rmP})  & \sin(\psi_{\rmP}) \\
-\sin(\psi_{\rmP})  & \cos(\psi_{\rmP}) 
\end{bmatrix}. 
\end{equation}
Note that $\RIP(\psi_{\rmP})=[\RPI(\psi_{\rmP})]^{\top}$. It is obvious that if $\eP\to \bf{0}$, then the geometrical task in the path following problem will be solved. Taking the time derivative of \eqref{eq: eP} yields
\begin{equation}  
\dot{\e}_{\subscript{\P}}=[\dot{R}^{\subscript{\I}}_{\subscript{\P}}(\psi_{\rmP})]^{\top}(\p - \pd)+\RIP(\psi_{\rmP})(\dot{\p} - \dot{\p}_{\rm d}).  
\end{equation}
Applying Lemma \ref{lemma: rotation matrix differential equation} (in the Appendix) for the first term of the previous equation we obtain 
\begin{equation} \label{eq: dynamics of eP}
\begin{aligned}
\dot{\e}_{\subscript{\P}}=-S(\bs{\omega}_{\rmP})\eP + \RIP(\psi_{\rmP})\dot{\p} - \RIP(\psi_{\rmP})\dot{\p}_{\rm d},   
\end{aligned}
\end{equation}
where $S(\bs{\omega}_{\rmP}) \in \R^{2\times 2} $ is a skew symmetric matrix parameterized by $\bs{\omega}_{\rmP} = [r_{\rmP},0 ]^{\top}\in \R^2 $ which is the angular velocity vector of $\{\P\}$  respect to $\Iframe$, expressed in $\{\P\}$. Note that $r_{\rmP}$ satisfies the relation 
\begin{equation} \label{eq: angular velocity vector of P in I} 
r_{\rmP}= 
\kappa(\gamma)u_{\rmP},  
\end{equation} 
where $u_{\rmP}$ is the total speed of $P$ given by \eqref{eq: uP} and $\kappa(\gamma)$ is the ``\emph{signed}" curvature of the path at $P$, given by 
\begin{equation} \label{eq: signed curvature formular }
 \kappa(\gamma)=\frac{\xd'(\gamma)\yd''(\gamma)-\xd''(\gamma)\yd'(\gamma) }{\Enorm{\pd'(\gamma)}^{3}}. 
\end{equation}
Note also that if $\gamma$ is the arc-length of the path then $\Enorm{\p'_{\rm d}(\gamma)}=1$. In this case, $u_{\rmP} = \dot{\gamma}$, i.e. the speed of the ``\emph{reference point}" equals the rate of change of the path length. Define 
\begin{equation} \label{eq: psie}
\psi_{\rm e}\triangleq \psi - \psi_{\rmP} 
\end{equation}
as the orientation error between the vehicle's heading and the tangent to the path. Then,
\begin{equation} \label{eq: difference in heading }
\RIP(\psi_{\rmP})\dot{\p} \stackrel{\eqref{eq: kinematics in 2D},\eqref{eq: RIP},\eqref{eq: psie}}{=} \begin{bmatrix}
u\cos(\psi_{\rm e})\\
u\sin(\psi_{\rm e})
\end{bmatrix}.
\end{equation}
Furthermore, letting ${\bf v}_{\rmP}\triangleq[u_{\rmP},0]^{\top} \in \R^{2}$ be the velocity of $P$ with respect to $\Iframe$, expressed in $\{\P\}$, yields  
\begin{equation} \label{eq: velocity of P in P}
\RIP(\psi_{\rmP})\dot{\p}_{\rm d}  = {\bf v}_{\rmP}.  
\end{equation}
Substituting \eqref{eq: difference in heading } and \eqref{eq: velocity of P in P} in \eqref{eq: dynamics of eP} we obtain the dynamics of the position error as
\begin{equation} \label{eq: dot eP}
\dot{\e}_{\subscript{\P}}=-S(\bs{\omega}_{\rmP})\eP +\begin{bmatrix}
u\cos(\psi_{\rm e}) \\
u\sin(\psi_{\rm e})
\end{bmatrix} - \begin{bmatrix}
u_{\rmP} \\
0
\end{bmatrix}.
\end{equation}
Furthermore, from \eqref{eq: psie} the dynamics of the orientation error are given by
\begin{equation} \label{eq: dote_psi}
\dot{\psi}_{\rm e} \stackrel{\eqref{eq: kinematics in 2D},\eqref{eq: angular velocity vector of P in I} }{=} r-\kappa(\gamma)u_{\rmP}.
\end{equation}  
At this point, it should be clear that the geometric task in the path following problem, stated in Section \ref{section: path following problem statement}, is equivalent to the problem of stabilizing the position error system \eqref{eq: dot eP}, i.e. making $\eP(t)\to {\bs 0}$ as $t\to \infty$. In what follows we will describe a number of path following methods available in the literature that solve this problem. These methods are categorized in Table \ref{tab: method in path frame}. In \emph{Methods 1} and \emph{3}, the ``\emph{reference point}" is chosen as the orthogonal projection of the center of mass of the vehicle on the path, thus the \emph{along-track} error $s_1(t)=0$ for all $t$. In this case, only the \emph{cross-track} $y_1$ needs to be stabilized to fulfill the geometrical task. In contrast, in the other methods the ``\emph{reference point}" is initialized arbitrarily anywhere on the path and its evolution is controlled by assigning a proper law for $\dot{\gamma}$ so as to make the \emph{cross-track} and \emph{along-track} errors converge to zero. In all of the methods, in order to fulfill the dynamics task in the path following problem (see \eqref{eq: dynamic task}), the linear speed of the vehicle is assigned with the desired speed profile, i.e. $u=U_d$. 

\begin{table}[hbt!] 	
	\begin{minipage}{\textwidth}
	\centering
	\caption{Methods proposed to stabilize $\eP$ to zero}
	{\begin{tabular*}{28pc}{@{\extracolsep{\fill}}lll@{}}\toprule
		PF Methods	& Drive $\eP$ to zero by & References \\	\midrule 
		Method 1	&  $u,r\textcolor{white}{\dot{\gamma},}$ 	&  \cite{Samson1993} 				\\ 
		Method 2	&  $u,r,\dot{\gamma}$					&  \cite{Lapierre2003}						\\ 
		Method 3\footnote{Line-of-Sight methods}	&  $u,\psi\textcolor{white}{\dot{\gamma},}$			& \cite{PAPOULIAS1991,FOSSEN2003,Fossen2015}  						\\ 
		Method 4	&  $u,\psi,\dot{\gamma}$				&  	\cite{Breivik2005}					\\ 
		Method 5\footnote{NMPC-based methods}	&  $u,r,\dot{\gamma}$					&  \cite{YuMPC2015,HungJRNC2019}						\\ \hline
	\end{tabular*}}{}
\label{tab: method in path frame}
\end{minipage}
\end{table}

\begin{remark}{
	In the formulation above, if $\gamma$ is the arc-length of the path, then in \eqref{eq: signed curvature formular } $\Enorm{\p'_{\rm d}(\gamma)}=1$ and thus $u_{\rmP}=\dot{\gamma}$. In this case, the path following error system composed by  \eqref{eq: dot eP} and \eqref{eq: dote_psi} resembles the path following error system developed in \cite{Lapierre2003}; see equation (5) in \cite{Lapierre2003}.
	Notice that although parameterizing a path by its arc-length is convenient, this is not always possible; elliptical and sinusoidal curves are typical examples, \cite{gray2006modern}. In the set-up above, the path parameter $\gamma$ is not necessarily the arc-length, thus making the formulation presented above applicable to any path. }	 
\end{remark}
\begin{remark} {
	It is very important to note that in order to obtain  $\dot{\psi}_{\rm e}$ in \eqref{eq: dote_psi}
	 ${\psi}_{\rm e}$ must be differentiable. This implies that $\psi$ and $\psi_{\rmP}$ must be differentiable as well. Although from a formulation standpoint there is no problem with this, in practice however, the heading sensors that measure $\psi$ normally return values in $[-\pi,\pi]$ or $[0, 2\pi]$, thus the discontinuities happen when the angle changes between $-\pi$ and $\pi$ or between $0$ and $2\pi$. In this situation, to make $\psi$ and $\psi_{\rmP}$ continuous and differentiable we should open the domain of these angles to $\R$ before using them in path following algorithms involving in using $\psi_{\rm e}$. We suggest a simple algorithm to deal with this situation in the path following toolbox that shall be presented in Section \ref{sec:Implementation}. }
\end{remark}
\subsubsection{Method 1 \cite{Samson1993}: Achieve path following by controlling $(u,r)$} \label{Section: path following method 1}
In this section we will describe the first method named as \emph{Method 1}, proposed in \cite{Samson1993}, to solve the path following problem. The main objective is to derive a control strategy for $(u,r)$ to drive the position error in the systems \eqref{eq: dot eP} and \eqref{eq: dote_psi} to zero. In this method, the ``\emph{reference point}" is chosen as the orthogonal projection of the real vehicle on the path, that is, the point on the path closest to the vehicle (if it is well-defined). With this strategy the \emph{along-track} error is always zero, i.e. $s_1(t)=0$ for all $t$. Thus, we only need to stabilize the \emph{cross-track} $y_1$ to zero to fulfill the geometry task. The dynamics of the \emph{cross-track} error can be written explicitly from \eqref{eq: dot eP} as 
\begin{equation} \label{eq: y1dot}
\dot{y}_1 = -r_{\rmP}s_1 + u\sin(\psi_{\rm e})  \stackrel{ s_1=0}{=}u\sin(\psi_{\rm e}).
\end{equation}
Define 
\begin{equation} \label{eq: tildepsi }
\tilde{\psi} = \psi_{\rm e}- \delta(y_1,u),
\end{equation}
 where $\delta(y_1,u)$ is a time differentiable design function that can be used to shape the manner in which the vehicle approaches the path. The design of $\delta(y_1,u)$ must satisfy the following condition.
\begin{condition}[for the design of $\delta(y_1,u)$  \cite{Samson1993,Lapierre2003}] ~ \\
	\begin{enumerate}[i)]
		\item $\delta(0,u)=0$ for all $u$.
		\item $y_1u\sin(\delta(y_1,u))\le 0$ for all $y_1,u$.
	\end{enumerate}	
	\label{con: for delta}
\end{condition}
Taking the time derivative of \eqref{eq: tildepsi } yields 
\begin{equation} \label{eq: dot_tilde_psi}
\dot{\tilde{\psi}}\stackrel{\eqref{eq: dote_psi}}{=}r-\kappa(\gamma)u_{\rmP}-\dot{\delta}.
\end{equation}
Notice also that because ${s}_1(t)=0$ for all $t$, $u_{\rmP}$ in the above equation is obtained by solving the first equation in \eqref{eq: dot eP}, yielding
\begin{equation} \label{eq:v}
u_{\rmP}=u\frac{\cos(\psi_{\rm e})}{1-\kappa(\gamma)y_1}.
\end{equation}
We obtain the following theorem.
	\begin{theorem} \label{theorem: samson method}
		Consider a system composed by the dynamics of the cross-track error in \eqref{eq: y1dot} and the orientation error in \eqref{eq: dot_tilde_psi}. Let 
		\begin{equation} \label{eq: u samson}
		u=U_{\rm d},
		\end{equation} 
		where $U_{\rm d}$ is the positive desired speed profile for the vehicle to track. Further let
		\begin{equation} \label{eq: r samson}
		r= \kappa(\gamma)u_{\rmP}+\dot{\delta}-k_1\tilde{\psi}-k_2y_1u\frac{\sin(\psi_{\rm e})-\sin(\delta)}{\tilde{\psi}}, 
		\end{equation}
		where $k_1,k_2 >0$ are tuning parameters, $\kappa(\gamma)$ is defined in \eqref{eq: signed curvature formular }, $\psi_{\rm e}$ and $\tilde{\psi}$ are given by \eqref{eq: psie}  and \eqref{eq: tildepsi }, respectively, $u_{\rmP}$ given by \eqref{eq:v}. Then, $y_1(t)$ and $\tilde{\psi}(t)$ converge to zero as $t\to \infty$.
	\end{theorem} 
Proof: See the Appendix - in Section \ref{Proof of theorem: samson method}. \\\\
Notice that because of Condition \ref{con: for delta} once $y_1,\tilde{\psi}\to 0$, then $\psi_e\to 0$ as well. This is true because we are assuming that 
the vehicle is under-actuated and has negligible lateral motion (that is, the total velocity vector is aligned with the body's
longitudinal axis). In this situation, the vehicle velocity's vector will align itself with the tangent vector of the P-T frame. To satisfy Condition \ref{con: for delta}, $\delta(\cdot)$  can be chosen as 
\begin{equation} \label{eq: for delta}
\delta(y_1,u)=-\theta\tanh(k_\delta y_1u),
\end{equation}
where $\theta \in (0,\pi/2)$ and $k_{\delta} > 0$ \cite{lapierre2006nonsingular}. In summary, this path following method can be implemented with Algorithm \ref{alg: simon method}.
\begin{algorithm}[!hbt] 
	\caption{PF algorithm using Method 1} 	
	\begin{algorithmic}[1]
		\State For every time $t$ do:
		\Procedure{PF CONTROLLER }{}	
		\State Find $P$ - the point on the path closest to the vehicle
		\State Compute $y_1$ by \eqref{eq: eP}, $\psi_{\rm e}$ by \eqref{eq: psie}, and $\tilde{\psi}$ by \eqref{eq: tildepsi }
		\State For inner-loop controllers (see Fig. \ref{fig: path following system}):
		\State \qquad - Compute the desired vehicle's forward speed by \eqref{eq: u samson}
		\State \qquad - Compute the desired vehicle's yaw rate  given by \eqref{eq: r samson}
		\EndProcedure
	\end{algorithmic}
	\label{alg: simon method}
\end{algorithm}
The implementation of line 3 in the algorithm is equivalent to solving an optimization problem to find $\gamma^{*}$, where
\begin{equation} \label{eq: gamma star}
\gamma^{*}= ~\underset{\gamma~ \in ~\Omega}{\operatorname{argmin}} \Enorm{\p - \pd(\gamma)},
\end{equation}
where $\pd(\gamma)$ is given by \eqref{eq: path in 2D}.
Note also that the computation in Algorithm \ref{alg: simon method} associated with $\gamma$ uses $\gamma^{*}$, which is obtained by solving \eqref{eq: gamma star}. In many cases, e.g. straight-line or circumference paths, $\gamma^{*}$ can be found analytically. 
\begin{remark}
Because the \emph{``reference point"} on the path for the vehicle to track is chosen closest to the vehicle, there exists a singularity when $y_1=1/\kappa(\gamma)$ which stems from \eqref{eq:v}. This happens for example when the path is a circumference and the vehicle goes through the center of it. 
\end{remark}
\subsubsection{Method 2 \cite{Lapierre2003}: Achieve path following by controlling $(u,r,\dot{\gamma})$} \label{Section: path following method 2}
In this section we will describe the second method, named as \emph{Method 2}, proposed by \cite{Lapierre2003} to solve the path following problem. The main objective is to derive a control law for $(u,r,\dot{\gamma})$ to drive the position and orientation errors in \eqref{eq: dot eP} and \eqref{eq: dote_psi} to zero in order to achieve path following. In this method, instead of choosing the ``\emph{reference point}" on the path that is closest to the vehicle as in \emph{Method 1}, this point can be initialized anywhere on the path, and its evolution is controlled by $\dot{\gamma}$ (to be defined later) in oder to eliminate the \emph{along-track} error \cite{Lapierre2003}. In order to eliminate the \emph{cross-track} and orientation errors, this method uses the same controller for $u,r$ as in \emph{Method 1}. We now state the main result of this method, followed by a discussion on the intuition behind this result. 
\begin{theorem} \label{theorem: Lapier method}
Consider the path following error system composed by \eqref{eq: dot eP} and \eqref{eq: dote_psi}. Let $u=U_d$ (as given by \eqref{eq: u samson}), $r$ be given by \eqref{eq: r samson}, and
\begin{equation} \label{eq: v lapier} 
	u_{\rmP}=u\cos(\psi_{\rm e})+k_3s_1,    
\end{equation}
where $k_3 > 0$. Then, $\e_{\rmP}(t), \psi_{\rm e}(t)\to 0$ as $t\to \infty$. 
\end{theorem} 
Proof: See in Appendix - Section \ref{Proof of theorem: Lapier method}.\\\\
Recall that in \eqref{eq: v lapier} $s_1$ is the \emph{along-track} error defined in \eqref{eq: eP}.
Because of the relation in \eqref{eq: uP}, the control law for $\dot{\gamma}$ is given by
\begin{equation} \label{eq: dotgamma Lapieer} 
	 \dot{\gamma}= u_{\rmP}/\Enorm{\p'_{\rm d}(\gamma)},  
\end{equation}
 where $u_{\rmP}$ is given by \eqref{eq: v lapier}. In summary, the path following strategy of this method can be implemented as specified by Algorithm \ref{alg: PF method 2 Lapier}.
\begin{algorithm}[!hbt] 
	\caption{PF algorithm using Method 2} 	
	\begin{algorithmic}[1]
		\State Initialize $\gamma(0)$
		\State For every time $t$ do:
		\Procedure{PF CONTROLLER }{}	
		\State Compute the position and the orientation errors using \eqref{eq: eP} and \eqref{eq: psie}.
		\State For inner-loop controllers (see Fig. \ref{fig: path following system}):
		\State \qquad - Compute the desired vehicle's forward speed using \eqref{eq: u samson} 
		\State \qquad - Compute the desired vehicle's yaw rate  given by \eqref{eq: r samson} 
		\State Compute $\dot{\gamma}$ in \eqref{eq: dotgamma Lapieer}, then integrate it to update the value of $\gamma$
		\EndProcedure
	\end{algorithmic}
	\label{alg: PF method 2 Lapier}
\end{algorithm}~\\
The control law for $u_{\rmP}$ in \eqref{eq: v lapier} implies that if the vehicle is behind/ahead of the ``\emph{reference point}" $(s_1<0/s_1>0)$ then the ``\emph{reference point}" decreases/increases its speed. Intuitively, it aims to adjust the speed of the \emph{``reference point"} to coordinate with the vehicle along the tangent axis of the P-T frame so as to reduce the \emph{along-track error} to zero.  Recall that in \emph{Method 1} proposed in \cite{Samson1993} this error is always zero because the ``\emph{reference point}" is chosen as the point on the path that is closest to the vehicle. Compared with \emph{Method 1} this strategy has the following advantages: i) it doesn't require an algorithm to find a point on the path that is closest to the vehicle and ii) it avoids the singularity that occurs in the method of \cite{Samson1993} when $y_1=1/\kappa(\gamma)$. 
\subsubsection{Method 3 \cite{Fossen2015}: Line-of-Sight path following}
\label{section: LOS method} 
\begin{figure}[h]
	\centering
	\includegraphics[width=100mm]{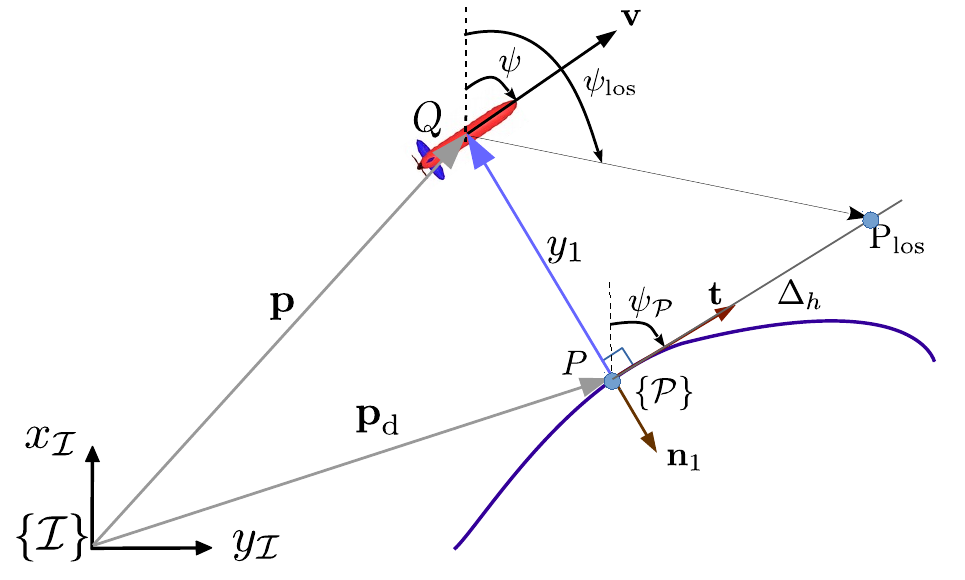} 
	\caption{Geometric illustration of LOS method, where $y_1$ is the cross-tracking error. }
	\label{fig: Geometry illustration LOS}
\end{figure}

\vspace{10pt}
We now present the third method, named \emph{Method 3}, to solve the path following problem. The main objective is to derive a control law for $(u,\psi)$ to drive the position error in the system \eqref{eq: dot eP} to zero in order to achieve path following. In the literature, this method is known as line-of-sight (LOS) method and was described in \cite{Fossen2015}. Earlier work on LOS methods for straight-lines can be found in \cite{PAPOULIAS1991,FOSSEN2003,Lekas2013}. This method can be also found in \cite{LIU2016799}. The LOS method is similar to \emph{Method 1} in the sense that the ``\emph{reference point}" is chosen as the orthogonal projection of the vehicle onto the path. Thus, the along \emph{track-error} $s_1(t)=0$ for all $t$. However, it is different from \emph{Method 1} in that it derives a control law for the vehicle's heading $\psi$ to achieve path following, instead of the heading rate $r$ as in \emph{Method 1}. We recall from \emph{Method 1} that because $s_1(t)=0$ the dynamics of \emph{cross-track} are given by \eqref{eq: y1dot}. In the present method $\psi_{\rm e}$ is viewed as a \emph{``control input"} whose control law must be chosen to stabilize the \emph{cross-track} error to zero.  
\begin{theorem} \label{theorem: method 3}
		Consider the cross track error system described by \eqref{eq: y1dot}. Let the vehicle's forward speed $u$ be given by \eqref{eq: u samson}. Let the control law for $\psi_e$ is given by    
	\begin{equation} \label{eq: psie in PF method 3}
	\psi_{\rm e} = \arctan\left(-\frac{y_1}{\Delta_{\rm h}}\right),
	\end{equation}
	where $\Delta_{\rm h}>0$ is a tuning parameter. Then, $y_1(t)\to 0$ as $t\to \infty$. Because of the relation in \eqref{eq: psie}, the control law for the vehicle's heading is given by
	\begin{equation} \label{eq: psi control law method 3}
	\psi=   \underbrace{\psi_{\rmP}+\arctan\left(-\frac{y_1}{\Delta_{\rm h}}\right)}_{\psi_{\rm los}}.
	\end{equation}
\end{theorem}
Proof: See Section \ref{Proof of theorem: method 3} of the Appendix. \\~\\
 In the literature, $\Delta_{\rm h}$ is often referred
 as the \emph{look-ahead distance} \cite{Lekas2013,Fossen2015} and $\psi_{\rm los}$ is called the light-of-sight angle that the heading of the vehicle should reach to achieve path following. An illustration of the relation between these variables is shown in Fig. \ref{fig: Geometry illustration LOS}. Notice that $\Delta_{\rm h}$ can be time varying and used to shape the convergence behavior towards
 the tangent (longitudinal) axis of $\{\P\}$. The larger value of $\Delta_{\rm h}$, the slower will the convergence be, but this in turn will require less aggressive turning maneuvers in order for the vehicle to reach the path. In summary, the path following \emph{Method 3} can be implemented using Algorithm \ref{alg: PF method 3}.

\begin{algorithm}[!hbt] 
	\caption{PF algorithm using Method 3} 	
	\begin{algorithmic}[1]
		\State For every time $t$ do:
		\Procedure{PF CONTROLLER }{}	
		\State Find $P$ -  the point on the path closest to the vehicle 
		\State For inner-loop controllers (see Fig. \ref{fig: path following system}):
		\State \qquad - Compute the desired vehicle's forward speed using \eqref{eq: u samson} 
		\State \qquad - Compute the desired vehicle's heading angle using \eqref{eq: psi control law method 3}.
		\EndProcedure
	\end{algorithmic}
	\label{alg: PF method 3}
\end{algorithm}
\begin{remark} {
The control law for the steering angle of the vehicle in \eqref{eq: psi control law method 3} can be rewritten in the Body frame as
 	\begin{equation} \label{eq: psi control law method 3 in body}
 \psi_{\B} = \psi_{e}+\arctan\left(-\frac{y_1}{\Delta_{\rm h}}\right),
 \end{equation}
 where $\psi_e$ is given by \eqref{eq: psie}.}
\end{remark}
It is interesting that this control law for the steering angle is equivalent to the one used in the autonomous car called Stanley who won
the DARPA Grand Challenge - the competition for autonomous driving in unrehearsed off-road terrain in 2005 \cite{Thrun2007}. 
\begin{figure}[h]
	\centering
	\includegraphics[width=100mm]{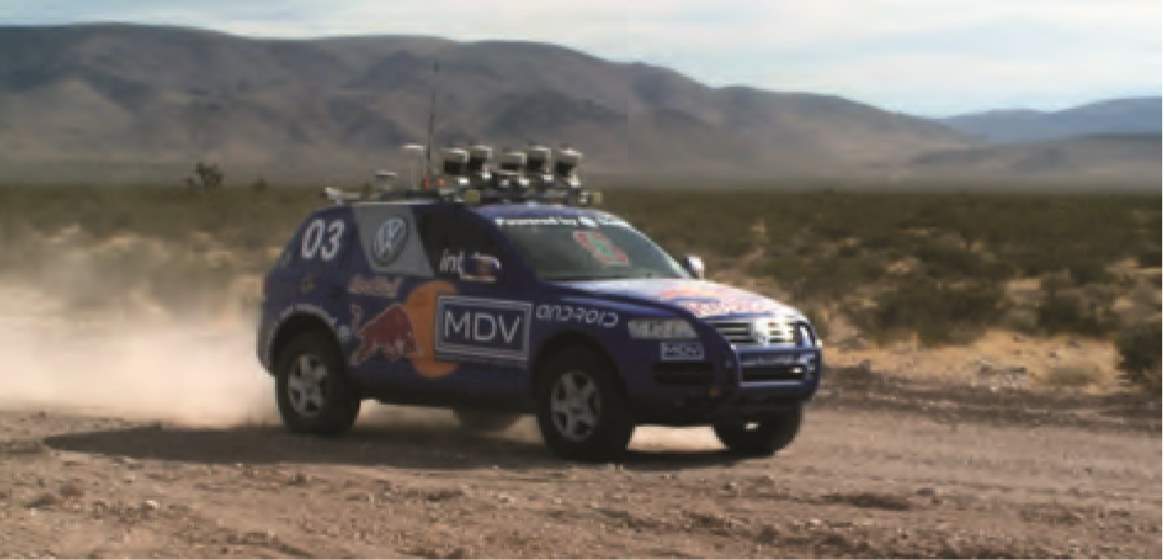} 
	\caption{Stanley robotic car \cite{Thrun2007} }
	\label{fig: Standley robot}
\end{figure} 

\begin{remark} {
The cross track error $y_1(t)$ in system \eqref{eq: y1dot} can also be driven to zero with the control law $\psi_{\rm e}$ given by
	\begin{equation} \label{eq: pramod law without ocean}
	\psi_{\rm e}  = \arcsin\left({\rm sat}\left(-\frac{1}{u}k_1y_1   \right)\right)
	\end{equation} 
	with any $k_1>0$. 
	In \eqref{eq: pramod law without ocean}, $\rm sat(\cdot)$ is a saturation function that returns values in $[-1,1]$, thus guaranteeing that the $\arcsin$ function is well defined. With \eqref{eq: pramod law without ocean}, the dynamics of the cross track error are given by $$\dot{y}_1 = -u~{\rm sat}\left(-\frac{1}{u}k_1y_1   \right),$$
	from which it is easy to see that $y_1(t)$ converges to zero as $t \to \infty$,   \cite{Pramod2009}.  
}
\end{remark}
\subsubsection{Method 4  \cite{Breivik2005}: Achieve path following by controlling $(u,\psi,\dot{\gamma})$} \label{section:  method 4}
This path following method is described in \cite{Breivik2005} and has a similar principle with that in \cite{Lapierre2003,Rysdyk2003}. The idea behind this method is quite similar to that in \emph{Method 2}; however, instead of achieving path following by controlling the heading rate $r$, the authors propose a control law for $\psi$.   
\begin{theorem} \label{theorem: method 4}
Consider the position error system described by \eqref{eq: dot eP}. Let the vehicle's forward speed $u$ be given by \eqref{eq: u samson}, the speed of the ``\emph{reference point}" $P$ $u_{\rmP}$ be given by \eqref{eq: v lapier}, and $\psi_{\rm e}$ given by \eqref{eq: psie in PF method 3}. Then, $\eP={\bf 0}$ is UGAS.
\end{theorem}
Proof: See Section \ref{Proof of theorem: method 4} in in the Appendix. \\~\\
Because of relation \eqref{eq: psie} the control laws for the vehicle's heading and $\dot{\gamma}$ are given by \eqref{eq: psi control law method 3} and \eqref{eq: dotgamma Lapieer}, respectively. The path following method can be implemented as specified in Algorithm \ref{alg: PF method 4}. It can be seen from Algorithms \ref{alg: PF method 2 Lapier} and \ref{alg: PF method 4} that the two are quite similar except that in the latter the vehicle is guided by controlling its heading, whereas in the former this is done by controlling its heading/yaw rate. 
\begin{algorithm}[!hbt] 
	\caption{PF algorithm using Method 4} 	
	\begin{algorithmic}[1]
		\State Initialize $\gamma(0)$
		\State For every time $t$ do:
		\Procedure{PF CONTROLLER }{}	
		\State Compute the position errors $s_1,y_1$ using \eqref{eq: eP}
		\State For inner-loop controllers (see Fig. \ref{fig: path following system}):
		\State \qquad - Compute the desired vehicle's forward speed using \eqref{eq: u samson} 
		\State \qquad - Compute the desired heading angle using \eqref{eq: psi control law method 3}.
		\State Compute $\dot{\gamma}$ using \eqref{eq: dotgamma Lapieer}, then integrate it to update the value of $\gamma$
		\EndProcedure
	\end{algorithmic}
	\label{alg: PF method 4}
\end{algorithm}
\begin{remark} {
	If the vehicle's heading in Theorem \ref{theorem: method 4} is replaced by \eqref{eq: pramod law without ocean}, then the path following error $\e_{\rmP}$ converges to zero asymptotically as well. This idea was presented in \cite{Miguelthesis} and can be proved simply as in the proof of Theorem \ref{theorem: method 4}. }
\end{remark}

\subsubsection{Method 5 \cite{YuMPC2015,HungJRNC2019}: NMPC-based path following} \label{section: Method 5}
The path following methods we have studied so far are conceptually simple in terms of design and implementation; however, they do not take into account the vehicle's physical constraints (e.g. maximum and minimum yaw rate). As a consequence, proper care must be taken to ensure that the resulting systems end up operating in a small region where the control law for the unconstrained system does not violate the constraints. In order to deal with the vehicle's constraints explicitly, in this section we will present an optimization based control strategy called model predictive control to solve the path following problem \cite{YuMPC2015,HungCAMS2018,HungJRNC2019}.\\
First, define 
 $${\bf x}_{\rmP}\triangleq [\eP^{\top} ,\psi_{\rm e},\gamma]^{\top}\in \R^{4}
$$  
as the state of the complete path following system where the position error $\eP$ is defined by \eqref{eq: eP} and the orientation error is defined by \eqref{eq: psie}. 
If we assign to the vehicle the desired speed profile, i.e. $u=U_{\rm d}$  then the dynamics of the complete path following system can be rewritten from \eqref{eq: dot eP} and \eqref{eq: dote_psi} as
\begin{equation} \label{eq: pf system for mpc 1}
\begin{split}
\dot{\bf x}_{\rmP}&=\f({\x}_{\rmP},\u_{\rmP})\triangleq\left[\begin{matrix}
-\Enorm{\p'_{\rm d}(\gamma)}v_{\gamma}\left(1-\kappa(\gamma)y_1\right)+U_{\rm d}\cos(\psi_{\rm e})\\
-\kappa(\gamma)s_1 \Enorm{\p'_{\rm d}(\gamma)}v_{\gamma} + U_{\rm d}\sin(\psi_{\rm e})   \\r-\kappa(\gamma)\Enorm{\p'_{\rm d}(\gamma)}v_{\gamma}\\
v_{\gamma}
\end{matrix}\right] ,\\
\end{split}
\end{equation} 
where $\u_{\rmP}\triangleq[r,v_{\gamma}]^{\top} \in \R^{2}$ is the input of the system, which is constrained in the set $\U_{\rmP}$ given by
\begin{equation} \label{set: UP}
\U_{\rmP} \triangleq \{ (r,v_{\gamma}): v_{\min}\le v_{\gamma} \le v_{\max},  \abs{r} \le r_{\max} \}.
\end{equation}
Here, $r_{\max}$ arises from the physical limitations of the vehicle while $v_{\min}$ and $v_{\max}$ are design parameters. In order for the path following problem to be solvable, the bounds on $v_{\gamma}$ can be chosen such that $v_{\min}\le 0$ and  
\begin{equation} \label{eq: dotgammamax}
v_{\max}>  \vd^{*}\triangleq\max_{}{(\vd)},
\end{equation} 
where $\vd$ is given by  \eqref{eq: vd}. Intuitively, the conditions on the bounds of $v_{\gamma}$ ensure that the ``\emph{reference point}" on the path that the vehicle must track has enough speed to coordinate with the vehicle in order to achieve path following and also to track the desired speed profile $\vd$. The main objective now is to find an MPC control law for $\u_{\rmP}$ to stabilize the position  orientation errors in the system \eqref{eq: pf system for mpc 1} to zero. To this end, we define a finite  horizon open loop optimal control problem (FOCP) that the MPC solves at every sampling time as follows:
\begin{definition} FOCP-1:
	\begin{equation}
	\displaystyle{\min_{\ubar_{\rmP}{}(\cdot)}}~J_{\rmP} \left({\x}_{\rmP} (t), \ubar_{\rmP}{} (\cdot)\right)
	\end{equation}  
	subject to
	\begin{subequations} \label{constraint of OPC}
		\begin{align}
		& {\dot{\xbar}}_{\rmP}(\tau)=\f_{\rmP}(\xbar_{\rmP}{(\tau)},\ubar_{\rmP}{(\tau)}),\tau \in \left[t,t+\Tp\right],\quad \xbar_{\rmP}{(t)}=\x_{\rmP}(t), \label{error dy}\\
		& \ubar_{\rmP}{(\tau)} \in \U_{\rmP}, \qquad \qquad \tau \in \left[t,t+T_{\rm p}\right], \label{input constraint}\\
		& \left(\bar{\e}_{\rmP}(\tau),\bar{\psi}_{\rm e}(\tau) \right) \in \mathbb{E}_{\rmP},  \quad \tau \in \left[t,t+T_{\rm p}\right]
		\end{align}
	\end{subequations}
	with $$J_{\rmP} \left(\x_{\rmP} (t),\ubar_{\rmP}{}(\cdot)\right)
\triangleq \int_{t}^{t+\Tp}\Enorm{\begin{bmatrix}
	\bar{\e}_{\rmP}(\tau) \\
	\bar{\psi}_{\rm e}(\tau)
	\end{bmatrix}}_{Q}+\Enorm{\bar{\u}_{\rm a}(\tau)}_{R} ~ {\rm d} \tau + F_{\rmP}\left(\bar{\e}_{\rmP}(t+T_{\rm p}),\bar{\psi}_{\rm e}(t + T_{\rm p})\right),$$ 
where $Q \in \R^{3\times 3} , R\in  \R^{2\times 2}$ are positive definite matrices, and the notation $||\x||_Q = \x^{\rm T}Q\x $ for any $\x \in R^{n}$ and $Q \in \R^{n\times n} $. The cost associated with the input is defined by 
\begin{equation} \label{eq: ua}
\u_{\rm a}=\begin{bmatrix}
U_{\rm d}\cos(\psi_{\rm e})-\Enorm{\p'_{\rm d}(\gamma)}v_{\gamma} \\
r-\kappa(\gamma)\Enorm{\p'_{\rm d}(\gamma)}v_{\gamma}
\end{bmatrix}.
\end{equation}
\end{definition}
In the FOCP-1, we use the bar notation to denote the predicted variables and to differentiate them from the actual variables which do not have a bar. Specifically, $\xbar_{\rmP}(\tau)$ is the predicted trajectory of the state $\x_{\rmP}$, computed using the dynamic model \eqref{eq: pf system for mpc 1} and the initial condition at the time $t$, driven by the input $\ubar_{\rmP}{(\tau)} ~ \text {with} ~ \tau \in [t,t+\Tp]$ over the prediction horizon $[t,t+\Tp]$. The first cost of $J_{\rmP}(\cdot)$ is associated with the geometrical task, whereas the choice of $\u_{\rm a}$ in the argument of the cost function to be minimized is motivated by the fact that once $\eP$ and $\psi_{\rm e}$ $\to {0}$, $\u_{\rm a}\to \bs{0}$ as well. $\mathbb{E}_{\rmP}$ and $F_{\rmP}$ represent the terminal constraints (the terminal set and the terminal cost, respectively), that should be designed appropriately to guarantee ``\emph{recursive feasibility}"\footnote{An MPC scheme is called recursive feasibility if its associated finite optimal control problem is feasible for all $t$ \cite{mayne2000constrained}.} and ``\emph{stability}" of the MPC scheme \cite{YuMPC2015}.   \\
In the MPC scheme, the FOCP-1 is repeatedly solved at every discrete sampling instant $t_{i}=iT_{\rm s}$, $i \in \N_{+}$, where $T_{\rm s}$ is a sampling interval. Let $\ubar_{\rmP}{}^{*}(\tau), \tau \in [t,t+T_{\rm p}],$ be the optimal solution of the FOCP-1. Then, the MPC control law $\u_{\rmP} (\cdot)$ is defined by
\begin{equation} \label{MPC control law}
\u_{\rmP}(t) = \ubar_{\rmP}^{*}{(t)},  ~ t \in [t_{i},t_{i}+T_{\rm s}].
\end{equation}
In summary, the MPC strategy for the path following problem can be implemented in Algorithm \ref{alg: PF method 5}.
\begin{algorithm}[!hbt] 
	\caption{PF algorithm using Method 5} 	
	\begin{algorithmic}[1]
		\State Initialize $\gamma(0)$
		\State For every time $t$ do:
		\Procedure{PF CONTROLLER }{}	
		\State Compute the path following errors $s_1,y_1, \psi_{\rm e}$ using \eqref{eq: eP} and \eqref{eq: psie}
		\State Solve the FOCP-1 and apply the MPC control law \eqref{MPC control law} to obtain optimal $r,v_{\gamma}$
		\State For inner-loop controllers (see Fig. \ref{fig: path following system}):
		\State \qquad -  Compute the desired vehicle's forward speed using \eqref{eq: u samson} 
		\State \qquad - The optimal $r$ is used as the desired vehicle's heading rate
		\State  Iterate $\gamma$ with the optimal input $v_{\gamma}$ to update $\gamma$
		\EndProcedure
	\end{algorithmic}
	\label{alg: PF method 5}
\end{algorithm} \\
	 Another way of ensuring stability for the path following error system \eqref{eq: pf system for mpc 1} without using the terminal constraints is to impose a ``contractive constraint" in the FOCP-1, see \cite{HungJRNC2019}. The ``contractive constraint" in the reference is designed based on the knowledge of an existing global nonlinear stablizing controller for the path following error system, and is advantageous over the NMPC scheme in \cite{YuMPC2015} in the sense that the origin of path following error is globally asymtotical stable. However, locally, with the NMPC scheme in \cite{YuMPC2015}, the path following error might converge to zero faster.
 \begin{remark}
 	While using the terminal constraints in \cite{YuMPC2015} and the contractive constraint in \cite{HungJRNC2019} are appealing from a theoretical standpoint, for the simplicity in design and implementation in practice they are normally excluded from the finite optimal control problem. In this situation, it is well-known that the
 	convergence of the path following error is guaranteed if the prediction horizon $T_{\rm p}$ is
 	chosen sufficiently large \cite{Jadbabaie2005,mayne2000constrained}. However, it requires effort in tuning the prediction horizon to achieve stability.
 \end{remark}
 \begin{remark}
Nowadays there are a variety of tools that can support solving the nonlinear optimization problem like FOCP-1. Typical tools that are widely used in MPC's work include Casadi \cite{Andersson2019}, which was used in the work of \cite{HungJRNC2019} and in the simulation toolbox of the present paper, ACADO \cite{Houska2011a}, which was used in the work of \cite{Batkovic2019}, or MATLAB optimization toolbox (fmincon function).   	 
\end{remark}
 
\subsection{Methods based on stabilizing the path following error in the body frame   }
\subsubsection{Method 6 \cite{aguiar2007trajectory}: Achieve path following by controlling $(u,r,\ddot{\gamma})$} \label{Section: Method 6 Aguiar}
\begin{figure}[h]
	\centering
	\includegraphics[width=100mm]{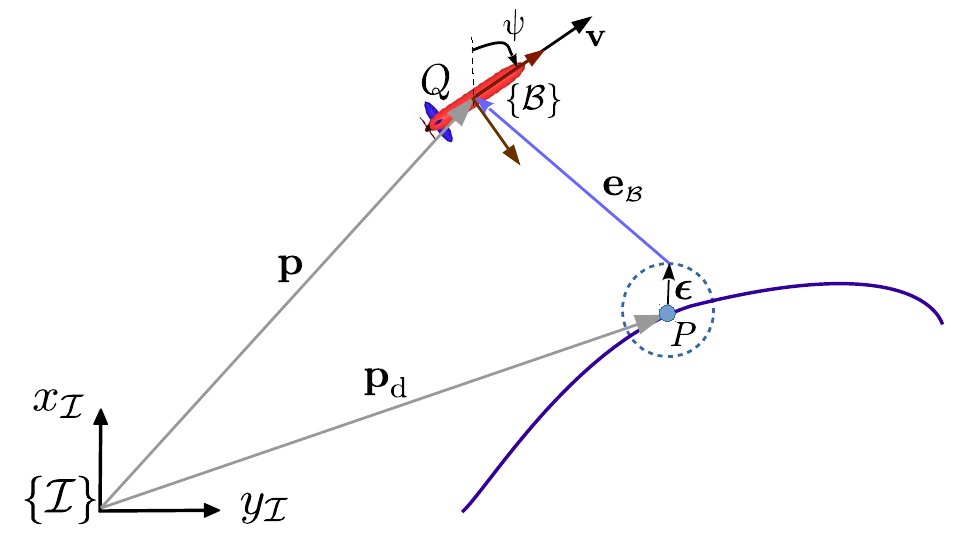} 
	\caption{Geometric illustration of path following Method 6  \cite{aguiar2007trajectory} }
	\label{fig: Geometry illustration Aguiar method}
\end{figure}

We now describe the second approach to the path following problem. It is different from the methods proposed in Section \ref{section: methods in P} in that the position error between the vehicle and the path is formulated in the vehicle's body frame $\{\mathcal{B}\}$, instead of a path frame, \cite{aguiar2007trajectory,Vannithesis}. A geometric illustration of this path following method is shown in Fig \ref{fig: Geometry illustration Aguiar method}.
First let $P$, whose coordinate is specified by $\pd(\gamma)$, be the ``\emph{reference point}" on the path that the vehicle should track to achieve path following. Define
\begin{equation} \label{eq: eB}
\begin{split}
\e_{\subscript{\B}}=\RIB(\psi)(\p-\p_{\rm d}) - {\bs \epsilon} 
\end{split}
\end{equation}
as the position error between the vehicle and the path resolved in the vehicle's body frame $\Bframe$, where ${\bs \epsilon}$ is an arbitrarily small non-zero constant vector and 
\begin{equation} \label{eq: rotation matrix from B to I}
\RIB(\psi)=\begin{bmatrix}
\cos(\psi)& \sin(\psi) \\
-\sin(\psi)& \cos(\psi) \\
\end{bmatrix}
\end{equation}
is the rotation matrix from the inertial frame $\Iframe$ to the body frame $\Bframe$. The reason for introducing ${\bs \epsilon}$ will become clear later. By definition, if $\eB$ can be driven to zero then the vehicle will converge to the ball centered at the point $P$ with radius $\Enorm{\bs \epsilon}$, which implies that the vehicle will converge to a neighborhood of the path that can be made arbitrarily small by choosing the size of $\boldsymbol{\epsilon}$. Taking the time derivative of \eqref{eq: eB} and using \eqref{eq: kinematics in 2D} and the fact that $\RBI(\psi)=[\RIB(\psi)]^{\top}$ yields 
\begin{equation*} 
\dot{\e}_{\subscript{\B}}=[\dot{R}^{\subscript{\I}}_{\subscript{\B}}(\psi)]^{\top}(\p - \pd)+\RIB(\psi)(\dot{\p} - \dot{\p}_{\rm d}) 
\end{equation*}
Using Lemma \ref{lemma: rotation matrix differential equation} in the Appendix for the first term, and expanding the previous equation we obtain
\begin{equation} \label{eq: following error aguiar method}
\begin{split}
\dot{\e}_{\subscript{\B}}&=  −S(\bs{\omega}){\e}_{\subscript{\B}} −S({\bs{\omega}}){\bs \epsilon} + {\bf v} − \RIB(\psi)\p'_{\rm d}(\gamma)\dot{\gamma} \\
&= −S(\bs \omega){\e}_{\subscript{\B}} +\Delta {\bf u} −  \RIB(\psi)\p'_{\rm d}(\gamma)\dot{\gamma},
\end{split}
\end{equation}
where ${\bs \omega}=[r,0]^{\top} \in \R^{2} $ is the angular velocity vector of $\Bframe$ respect to $\Iframe$, expressed in $\Bframe$, $\u=[u,r]$ and 
$
{\Delta}= \begin{bmatrix}
1 &\epsilon_2\\
0 &\minus\epsilon_1
\end{bmatrix}
$. To address the dynamic task stated in  \eqref{eq: dynamic task for gamma}, define  \begin{equation} \label{eq: e gamma}
e_{\gamma}=\dot{\gamma} - v_{\rm d}
\end{equation} 
as the tracking error for the speed of the path parameter. By definition, if $e_{\gamma}$ can be driven to zero then the dynamics task is fulfilled. The dynamics of this error is given by
\begin{equation} \label{eq: gamma dot}
\dot{e}_{\gamma}=\ddot{\gamma}-\dot{v}_{\rm d}.
\end{equation} 
Let ${\bf x}=[\e^{\top}_{\rmB}, e_\gamma ]^{\rmT}\in \R^{3} $ be the complete path following error vector. From \eqref{eq: following error aguiar method} and \eqref{eq: gamma dot}, the dynamics of the path following error vector can be expressed as
\begin{equation} \label{eq: complete the path following error system Aguiar method}
\begin{aligned}
\dot{\bf x}
= \left[\begin{matrix}
−S(\bs \omega){\e}_{\subscript{\B}} +\Delta {\bf u} −  \RIB(\psi)\p'_{\rm d}(\gamma)\dot{\gamma}\\
\ddot{\gamma}-\dot{v}_{\rm d}
\end{matrix}\right].
\end{aligned} 
\end{equation}
%
%
%
\begin{theorem} \label{theorem: Aguiar method}
	Consider the path following error system described by \eqref{eq: complete the path following error system Aguiar method}. Then, the control law for $\u$ and $\ddot{\gamma}$ given by
\begin{subequations} \label{eq: Aguiar controller}
\begin{align}
{\bf u}&=\bar{\Delta}\left(\RIB(\psi)\p'_{\rm d}(\gamma)v_{\rm d} -K_{\rm p}{\e}_{\rmB} \right) \label{eq: Aguiar control law for u} \\
\ddot{\gamma}&=-k_{\gamma}e_\gamma + \e^{\top}_{\rmB}\RIB({\psi})\p'_{\rm d}(\gamma) +\dot{v}_{\rm d},  \label{eq: Aguiar control law for gammaddot}
\end{align}
\end{subequations} 
render the origin of $\x$ GES, where $\bar{\Delta}=\Delta^{\top}(\Delta\Delta^{\top})^{-1}$, $K_{\rm p}$ is a positive definite matrix with appropriate dimension, and $k_{\gamma} >0$.
\end{theorem}
In summary, this path following method can be implemented as described in Algorithm \ref{alg: PF method 3 Aguiar}.
\begin{algorithm}[!hbt] 
	\caption{PF Algorithm using Method 6} 	
	\begin{algorithmic}[1]
		\State Initialize $\gamma(0)$ and $\dot{\gamma}(0)$
		\State For every sampling interval, repeat the following procedure:
		\Procedure{PF CONTROLLER }{}	
		\State Compute the position error $\eB$ using \eqref{eq: eB} and the tracking error $e_{\gamma}$ using \eqref{eq: e gamma}.    
		\State For inner-loop controllers (see Fig. \ref{fig: path following system}):
		\State 	\qquad - Compute the desired vehicle's forward speed and yaw-rate using \eqref{eq: Aguiar control law for u} 
		\State Compute $\ddot{\gamma}$ using \eqref{eq: Aguiar control law for gammaddot}, then integrate it to update the value of $\gamma$
		\EndProcedure
	\end{algorithmic}
	\label{alg: PF method 3 Aguiar}
\end{algorithm}

Proof: See Section \ref{Proof of theorem: Aguiar method} in the Appendix.
\begin{remark}
	In \eqref{eq: Aguiar controller} we can control the evolution of the ``\emph{reference point}" by assigning $\dot{\gamma}=\vd$, instead of using the control law for $\ddot{\gamma}$, as in \eqref{eq: Aguiar control law for gammaddot}. It can be easily shown that the origin of the path following error system is still GES as well. However, making $\dot{\gamma}=\vd$ implies that the ``\emph{reference point}" moves without taking into consideration the state of the vehicle. In other words, in this case the vehicle tracks a pure trajectory and this might demand more aggressive manoeuvres. 
\end{remark}

\begin{remark}
	It is worth noticing that this method is not only applicable to the kinematic model \eqref{eq: kinematics in 2D} but also to the dynamics model of quad-rotor drone. For further details we refer the reader to the work presented in \cite{Marcelothesis}.
\end{remark}
\subsubsection{Method 7 \cite{alessandretti2013trajectory}: NMPC-based path following}
In this section, we will describe the path following NMPC scheme in \cite{alessandretti2013trajectory}. This scheme borrows from the formulation in Section \ref
{Section: Method 6 Aguiar}. Using this formalism, path following is achieved by deriving an NMPC control law for $(u,r,\dot{\gamma})$ to i) drive the position error $\eB$ whose dynamics described by \eqref{eq: following error aguiar method} to zero and also to ii) ensure that $\dot{\gamma}$ converges to $\vd$ to fulfill the dynamics task in \eqref{eq: dynamic task for gamma}. For this purpose, define
\begin{equation} \label{eq: xB and uB}
\x_{\rmB}\triangleq[\e^{\top}_{\rmB},\psi,\gamma]^{\top} \in \R^{4}  , \quad \u_{\rmB}\triangleq [u,r,v_{\gamma}]^{\top}\in \R^{3} 
\end{equation}
as the state and input of the complete path following system, respectively. Note that similar to the NMPC scheme in \emph{Method 5} we let $v_{\gamma}=\dot{\gamma}$ for convenience of presentation. Note also that in the present NMPC scheme, the vehicle's speed is naturally optimized through the NMPC scheme, whereas with the NMPC scheme in \emph{Method 5}, it is assigned directly by the desired speed profile $U_{\rm d}$. Because the vehicle's speed is constrained due to the vehicle's physical limitations, we define a constraint set for $\u_{\rmB}$ in \eqref{eq: xB and uB} as
\begin{equation} \label{eq: U in B}
\U_{\rmB} \triangleq \U_{\rmP}\times \{u: u_{\min} \le u \le u_{\max}\},
\end{equation}
where $\U_{\rmP}$ given by \eqref{set: UP} and the bounds on the vehicle's speed vary according to the physical constraints of the vehicle. 
 The dynamics of the complete path following state can be rewritten from \eqref{eq: kinematics in 2D} and \eqref{eq: following error aguiar method} as
\begin{equation} \label{eq: pf system for mpc 2}
\dot{\x}_{\rmB}= {\f}_{\rmB}(\x_{\rmB},\u_{\rmB} ) \begin{bmatrix}
 −S(\bs \omega){\e}_{\subscript{\B}} +\Delta {\bf u} −  \RIB(\psi)\p'_{\rm d}(\gamma)v_{\gamma}\\
 r\\
 v_{\gamma}
\end{bmatrix},
\end{equation}
The objective of the NMPC scheme is to find an optimal control strategy for $\u_{\rmB}$ to drive the position error $\e_{\rmB}$ and the speed tracking error $(v_{\gamma} - v_{\rm d})$ to zero. To this end, we now define a FOCP that the NMPC solves at every sampling time as follows:
\begin{definition} FOCP-2:
	\begin{equation}
	\displaystyle{\min_{\ubar_{\rmB}{}(\cdot)}}~J_{\rmB} \left({\x}_{\rmB} (t), \ubar_{\rmB}{} (\cdot)\right)
	\end{equation}  
	subject to
	\begin{subequations} \label{constraint of OPC2}
		\begin{align}
		& {\dot{\bar{\x}}_{\rmB}}(\tau)=\f_{\rmB}(\bar{\x}_{\rmB}{(\tau)},\ubar_{\rmB}{(\tau)}),\tau \in \left[t,t+\Tp\right],\quad {\bar{\x}}_{\rmB}{(t)}={\x}_{\rmB}(t), \label{error dy 1}\\
		& \ubar_{\rmB}{(\tau)} \in \U_{\rmB}, \quad \tau \in \left[t,t+T_{\rm p}\right], \label{input constraint2} \\
		& \bar{\e}_{\rmB}(\tau) \in \mathbb{E}_{\rmB}, \quad \tau \in \left[t,t+T_{\rm p}\right].
		\end{align}
	\end{subequations}
	with $$J_{\rmB} \left(\x_{\rmB} (t),\ubar_{\rmB}{}(\cdot)\right)
	\triangleq \int_{t}^{t+\Tp}\Enorm{\bar{\e}_{\rmB}(\tau)}_{Q}+\Enorm{\u_{\rm b}(\tau)}_{R} + \Enorm{v_{\gamma}(\tau) - v_{\rm d}(\tau) }_{O} ~ {\rm d} \tau +  F_{\rmB}(\bar{\e}_{\rmB}(t+T_{\rm p}))$$  
	where $Q, R,O   \succ 0$ and 
	\begin{equation} \label{eq: ub}
	\u_{\rm b}=
	\Delta {\bf u} −  \RIB(\psi)\p'_{\rm d}(\gamma)v_{\gamma}.
	\end{equation}
\end{definition}
In the FOCP, we use the bar notation to denote the predicted variables, to differentiate them from the actual variables which do not have a bar. Specifically, $\xbar_{\rmB}(\tau)$ is the predicted trajectory of $\x_{\rmB}$, using the dynamic model \eqref{eq: pf system for mpc 2} and their perspective initial condition at time $t$, driven by the input $\ubar_{\rmB}{(\tau)} ~ \text {with} ~ \tau \in [t,t+\Tp]$ over the prediction horizon $T_{\rm p}$. 
$\mathbb{E}_{\rmB}$ and $F_{\rmB}$ are called terminal set and terminal cost, respectively, to be designed appropriately to guarantee ``\emph{recursive feasibility}" and ``\emph{stability}" of the NMPC scheme \cite{alessandretti2013trajectory}. The choice of $\u_{\rm b}$ in the cost function to be minimized is motivated by the fact that once $\e_{\rmB} \to \bs{0}$, $\u_{\rm b}\to \bs{0}$ as well.  \\
In the NMPC scheme, the FOCP-2 is repeatedly solved at every discrete sampling instant $t_{i}=iT_{\rm s}$, $i \in \N_{+}$, where we recall that $T_{\rm s}$ is a sampling interval. Let $\ubar_{\rmB}{}^{*}(\tau), \tau \in [t,t+T_{\rm p}],$ be the optimal solution of the FOCP-2. Then, the NMPC control law $\u_{\rmB} (\cdot)$ is defined as
\begin{equation} \label{MPC control law 2}
\u_{\rmB}(t) = \ubar_{\rmB}^{*}{(t)},  ~ t \in [t_{i},t_{i}+T_{\rm s}].
\end{equation}
In summary the NMPC scheme can be implemented as in Algorithm \ref{alg: PF method 7}.
\begin{algorithm}[!hbt] 
	\caption{PF algorithm using Method 7} 	
	\begin{algorithmic}[1]
		\State Initialize $\gamma(0)$
		\State For every time $t$ do:
		\Procedure{PF CONTROLLER }{}	
		\State Compute the path following errors $s_1,y_1, \psi_{\rm e}$ using \eqref{eq: eP} and \eqref{eq: psie}
		\State Solve the FOCP-2 and apply the NMPC control law \eqref{MPC control law} to obtain optimal values of $u,r,v_{\gamma}$ 
		\State For inner-loop controllers (see Fig. \ref{fig: path following system}):
		\State \qquad - the optimal $u$ is used as the desired vehicle's forward speed.  
		\State \qquad - the optimal $r$ is used as the desired vehicle's heading rate.  
		\State  Iterate $\gamma$ with the optimal input $v_{\gamma}$ to update the value of $\gamma$
		\EndProcedure
	\end{algorithmic}
	\label{alg: PF method 7}
\end{algorithm}\\
	Note that for simplicity of design and implementation we can exclude the terminal constraints above from the finite optimal control problem. In this situation, we recall that the convergence of the path following error is guaranteed if the prediction horizon $T_{\rm p}$ is chosen sufficiently large \cite{Jadbabaie2005,mayne2000constrained}. 
\section{Path following methods in the presence of external disturbances}
\label{section: extension with unknown disturbance}

In the previous section, we described a number of path following methods for the nominal kinematic model \eqref{eq: kinematics in 2D}. We now discuss how these methods can be extended to the cases when the vehicle maneuvers in an environment where \emph{unknown constant external disturbances} e.g. wind in the case of UAVs or ocean currents in the case of AMVs are present. In this situation, the vehicle kinematic model in \eqref{eq: kinematics in 2D} can be extended as 
\begin{equation} \label{eq: kinematics in 2D extended}
\begin{split}
\dot{x}&=u\cos(\psi) + v_{\rm cx} \\
\dot{y}&=u\sin(\psi) + v_{\rm cy}					\\
\dot{\psi}&=r	\\		
\dot{v}_{\rm cx} &= 0, \quad \dot{v}_{\rm cy} = 0,    	
\end{split}
\end{equation}  
where $v_{\rm cx}, v_{\rm cy}$ are two components of ${\bf v}_{\rm c}$, i.e. ${\bf v}_{\rm c} = [v_{\rm cx},v_{\rm cy}]^{\top} \in \R^2$  - the vector that describes the influence of the \emph{external unknown disturbance} in the inertial frame. It is important to note that in \eqref{eq: kinematics in 2D} $u$ is the longitudinal/surge speed measured with respect to the fluid. In the literature, the effect of a \emph{constant disturbance} can be eliminated using two approaches. The first is to add an integral term in the path following control laws. This is simple but only applicable for straight-line paths. The second uses estimates of the disturbances and is applicable to every path. These approaches are presented next.
\begin{remark} 
	Note that in the present paper we only address the cases where the external disturbance is a constant. For more general cases where the disturbance is an unknown sinusoidal signal the reader is referred to the work in \cite{Mohamad2017} and \cite{Mohamad2021}. The idea behind this work is to use an adaptive internal model to estimate the disturbance and then use it in a combination with the path following \emph{Method 6} to cancel out the disturbance effect. 
\end{remark}
\subsection{Path following with integral terms}
The LOS type path following methods presented in Section \ref{section: LOS method}  can be extended in a simple manner to handle constant external disturbances. For this purpose, we first re-derive the dynamics of the path following error in the presence of an external disturbance. Given the kinematic model in \eqref{eq: kinematics in 2D extended} and following the procedure described in Section \ref{section: derivation of path following error in path frame}, it can be shown that the dynamics of the position error $\e_{\rmP}$ (expressed in the P-T frame) are given by
\begin{equation} \label{eq: dot eP with current}
\dot{\e}_{\subscript{\P}}=-S(\bs{\omega}_{\rmP})\eP +\begin{bmatrix}
u\cos(\psi_{\rm e}) \\
u\sin(\psi_{\rm e})
\end{bmatrix} - \begin{bmatrix}
u_{\rmP}  \\
0
\end{bmatrix} + \RIP(\psi_{\rmP})\vc  .
\end{equation}
Recall that $\e_{\rmP}=[s_1,y_1]^{\top}$ is defined by \eqref{eq: eP} while the orientation error $\psi_{\rm e}$ is defined by \eqref{eq: psie}. Note that in the above equation, the last term represents the influence of the external disturbance $\vc$. We will show that this disturbance can be eliminated by adding an integral term in the LOS methods provided that the following assumptions are satisfied 
\begin{assumption} \label{ass: for ILOS}
	\textit{
	 The ``reference point" on the path for the vehicle to track is chosen as the closet point to the vehicle, i.e. the along-track error $s_1(t)=0$ for all $t$.  
}	
\end{assumption}
With the above assumption the dynamics of the \emph{cross track} can be rewritten from \eqref{eq: dot eP with current} as
\begin{equation} \label{eq: dynamcis of y_1 with disturbance}
\dot{y}_1 = u\sin(\psi_{\rm e}) + \vcy^{\rmP}, 
\end{equation}
where 
\begin{equation} \label{eq: vcyP}
\vcy^{\rmP} \triangleq -\sin(\psi_{\rmP})v_{\rm cx} + \cos(\psi_{\rmP})v_{\rm cy}.
\end{equation}
is the external disturbance acting along the coordinate $x_{\rmP}$ of the P-T frame. 
 If $\vc$ and $\psi_{\rmP}$ are constant, then the unknown disturbance $\vcy^{\rmP}$ is constant as well. Thus, by adding an integral terms in the LOS methods presented in Section \ref{section: LOS method}, the effect of $\vcy^{\rmP}$ on the \emph{cross-track} error can be eliminated. 
\begin{theorem}
\textit{	
Consider the dynamics of the cross-track error given by \eqref{eq: dynamcis of y_1 with disturbance} where $\vcy^{\rmP}$ is assumed to constant (i.e. the path is a straight-line). The control law for $\psi_{\rm e}$ given by
\begin{equation} \label{eq: psie control law method 3 ILOS}
\begin{split}
\psi_{\rm e} & = \arctan\left(-\frac{y_1 + \sigma y_{\rm int}}{\Delta_{\rm h}}\right) \quad \text{and}\\
\dot{y}_{\rm int} & = \frac{\Delta_{\rm h}y_1}{(y_1 + \sigma y_{\rm int})^2 + \Delta_{\rm h}^2} 
\end{split}
\end{equation}
with $\Delta_{\rm h}, \sigma>0$ are tuning parameters drives $y_1(t)$ to zero as $t\to \infty$. Because of the relation in \eqref{eq: psie}, the control law for the vehicle's heading $\psi$  is given by
\begin{equation} \label{eq: psi control law method 3 ILOS}
\psi =  \psi_{\rm e} + \psi_{\rmP}. 
\end{equation}
}
where in this situation $\psi_{\rm e}$ is given by \ref{eq: psie control law method 3 ILOS}.
\end{theorem} 
The guidance law in the theorem is referred as integral line-of-sight (ILOS). The convergence of the \emph{cross-track} error is proved in \cite{Caharija2016}. \\
Another way to reject the disturbance is by adding an integral term to \eqref{eq: pramod law without ocean}, yielding a control law for $\psi_e$, given by
\begin{equation} \label{eq: pramod law with ocean}
\psi_{\rm e} = \arcsin\left({\rm sat} \left( -\frac{1}{u}k_1y_1 - {k_2\int_{0}^{t}y_1(\tau){\rm d}\tau}\right) \right),
\end{equation} 
where $k_1,k_2$ are design parameters. The rational behind this design is that without saturation the resulting \emph{cross-track} error system
\eqref{eq: dynamcis of y_1 with disturbance} is given by
\begin{equation} \label{eq: dynamcis of y_1 with disturbance closeloop}
\dot{y}_1 = -k_1y_1 - {k_2\int_{0}^{t}y_1(\tau){\rm d}\tau} + \vcy^{\rmP}. 
\end{equation}
It is well-known that the integral term in \eqref{eq: dynamcis of y_1 with disturbance closeloop} is capable of canceling the effect of the constant disturbance $\vcy^{\rmP}$. Further,  
 $k_1,k_2$ can be chosen so as to obtain a desired natural frequency and damping factor for the above second order system, \cite{Pramod2009}. \\
Although adding an integral terms in the LOS guidance methods is a natural and intuitive way to reject a constant disturbance, the main limitation of this approach is that it can only reject the disturbance completely if the path is a straight-line.       
\subsection{Path following with estimation of external disturbances}
An alternative approach to eliminate the effect of an external disturbance is to estimate it and then use the estimate in the path following algorithms. 
\subsubsection{Disturbance estimation}
The disturbance can be estimated using the underlying model described in  \eqref{eq: kinematics in 2D extended} \cite{Aguiar2002Dynamic}. Suppose that the navigation system of the vehicle provides estimate of its position $\p = [x,y]^{\top}$, longitude/surge speed $u$ wrt. the fluid, and heading $\psi$. Define
$$
\x_c \triangleq [\p^{\top}, {\bf v}^{\top}_c ]^{\top} \in \R^{4},  \quad \u \triangleq [u\cos(\psi), u\sin(\psi)]^{\top} \in \R^2,    \quad  {\bf y}_c \triangleq \p.
$$
From \eqref{eq: kinematics in 2D extended} we obtain the linear time invariant system
\begin{equation} \label{eq: ocean current estimation model}
\begin{split}
\dot{\x}_c  &= A_c\x_c  + B_c \u \\
{\bf y}_c  &= C_c\x_c,  
\end{split}
\end{equation}
where
$$A_c = \begin{bmatrix}
0_{2\times 2} & I_{2\times 2}\\
0_{2\times 2} & 0_{2\times 2} 
\end{bmatrix}  \qquad B_c = \begin{bmatrix}
I_{2\times 2} \\
0_{2\times 2}  
\end{bmatrix} \qquad C_c = \begin{bmatrix}
I_{2\times 2} & 0_{2\times 2}
\end{bmatrix}.  $$
It is easy to check that the observability matrix of the above system computed from the pair $(A_c, C_c)$ is full rank, hence the system \eqref{eq: ocean current estimation model} is observable, implying that the external disturbance ${\bf v}_c$ can be estimated using model \eqref{eq: ocean current estimation model}. Let $\hat{\x}_c$ denote the estimate of ${\x}_c$. In order to estimate $\vc$ (through estimating $\x_c$), one can adopt the estimator given by
\begin{equation} \label{eq: dynamic of xc hat}
\begin{split}
\dot{\hat{\x}}_c  &= A_c\hat{\x}_c  + B_c \u + K_c ({\bf y}_c - \hat{\bf y}_c)\\
\hat{\bf y}_c &= C_c\hat{\x}_c, 
\end{split}
\end{equation} 
where $K_c\in \R^{2\times 2}$ is the designed matrix, to be defined. Let $\tilde{\x}_c\triangleq \x_c - \hat{\x}_c$ be the estimation error. Then, it follows from \eqref{eq: ocean current estimation model} and \eqref{eq: dynamic of xc hat} that
\begin{equation}
\dot{\tilde{\x}}_c = (A_c + K_cC_c)\tilde{\x}_c. 
\end{equation}
We obtain the following result.
\begin{lemma} \label{lemma: ocean current estimation}
	Consider the estimator \eqref{eq: dynamic of xc hat}. Then, the origin of the estimation error $\tilde{\x}$ is GES if $K$ is chosen such that $(A_c + K_cC_c)$ is Hurwitz (i.e. the real part of all its eigenvalues are negative).   
\end{lemma}
We now revisit the path following methods in the previous section and modify them to eliminate the effect of the disturbance by assuming that the disturbance is estimated using the above described method. We show how to achieve this with Method 3 (Section \ref{section: LOS method}) and Method 6 (Section \ref{Section: Method 6 Aguiar}). The other methods can be extended similarly to deal with external disturbances. 
\subsubsection{Method 3 with estimation of external disturbances}
Recall that in Method 3 (Section \ref{section: LOS method}), the \emph{``reference point"} on the path for the vehicle to track is chosen as the one closest to the vehicle, and therefore the along track error $s_1(t)=0$ for all $t$. As a consequence, under the effect of the external disturbance the dynamics of the \emph{cross-track} error satisfy \eqref{eq: dynamcis of y_1 with disturbance}. In order to eliminate the effect of $v_{\rm cy}^{\rmP}$ in \eqref{eq: dynamcis of y_1 with disturbance} we can add the estimate of the disturbance to the control law \eqref{eq: pramod law without ocean}, yielding
  \begin{equation} \label{eq: pramod law with disturbance}
\psi  = \psi_{\rmP} +  \arcsin\left({\rm sat}\left(-\frac{1}{u}k_1y_1 - \frac{1}{u}\hat{v}_{\rm cy}^{\rmP}  \right)\right)
\end{equation} 
where 
\begin{equation} \label{eq: vcyP hat}
\hat{v}_{\rm cy}^{\rmP} \triangleq -\sin(\psi_{\rmP})\hat{v}_{\rm cx} + \cos(\psi_{\rmP})\hat{v}_{\rm cy}.
\end{equation}
%
\subsubsection{Method 6 with estimation of the external disturbances}
This approach was described in \cite{Vannithesis} and is summarized as follows. With the kinematic model in \eqref{eq: kinematics in 2D extended} and following the procedure described in Section \ref{Section: Method 6 Aguiar}, we can show that the dynamics of the position error in the vehicle body frame $\e_{\rmB}$ (defined by \eqref{eq: eB}) are given by
\begin{equation} \label{eq: following error aguiar method with current}
\dot{\e}_{\subscript{\B}}= −S(\bs \omega){\e}_{\subscript{\B}} +\Delta {\bf u} −  \RIB(\psi)\p'_{\rm d}(\gamma)\dot{\gamma} + \RIB(\psi)\vc. 
\end{equation}
where $\Delta$ is a design matrix defined in Section \ref{Section: Method 6 Aguiar}. The last term in the above equation represents the influence of the disturbance on the path following error. Clearly, if $\vc = 0$ then \eqref{eq: following error aguiar method with current} is the same as \eqref{eq: following error aguiar method} (the case without external disturbance). In order to reject the external disturbance in \eqref{eq: following error aguiar method with current} we modify the control law for the vehicle input in \eqref{eq: Aguiar control law for u} as
\begin{equation} \label{eq: Aguiar control law for u current}
{\bf u}=\bar{\Delta}\left(\RIB(\psi)\p'_{\rm d}(\gamma)v_{\rm d} -K_{\rm p}{\e}_{\rmB}  - \RIB(\psi)\hat{\bf v}_c \right) ,
\end{equation}
where $\hat{\bf v}_c$ is the estimate of the external disturbance obtained from estimator \eqref{eq: dynamic of xc hat}. Note that the control law for $\ddot{\gamma}$ remains the same as in \eqref{eq: Aguiar control law for gammaddot}. With this control law, we obtain the following result
\begin{lemma} \label{lemma: aguiar method with current}
	Let $\x = [\e_{\rmB}, e_{\gamma}]$ be the path following error whose dynamics are described by \eqref{eq: following error aguiar method with current} and \eqref{eq: gamma dot}. Let also $\e_{c} \triangleq {\bf v}_c - {\hat{\bf v}}_c$ be the estimation error of the external disturbance. Then, with the control laws \eqref{eq: Aguiar control law for u current} and  \eqref{eq: Aguiar control law for gammaddot} the path following error system composed by \eqref{eq: following error aguiar method with current} and \eqref{eq: gamma dot} is input-to-state stable (ISS) with respect to the state $\x$ and the input $\e_c$.
\end{lemma} 
Proof: see Section \ref{proof of lemma: aguiar method with current} in the Appendix. \\\\
For the definition of an ISS system, we refer the reader to Definition 7 in \cite{khalil2002}. The ISS property implies that as long as the estimation error $\e_c$ is bounded, then the path following error $\x$ is also bounded. Furthermore, if  $\e_c(t)\to \bf{0}$ as $t\to \infty$ then $\x(t)\to \bf{0}$ as $t\to \infty$.  \\
With the path following controller given in \eqref{eq: Aguiar control law for u current} where the ocean current is estimated by \eqref{eq: dynamic of xc hat}, the main result in this section is stated in the next theorem.
\begin{theorem}
	Consider the closed-loop path following system composed by the path following error system described by \eqref{eq: following error aguiar method with current} and \eqref{eq: gamma dot} and the disturbance estimation system described by \eqref{eq: dynamic of xc hat}.
	\begin{enumerate}[i.]
		\item Let $K_c$ be chosen such that $A_c +K_cC_c$ is Hurwitz
		\item Let $\u$ be given by \eqref{eq: Aguiar control law for u current} and $\ddot{\gamma}$ be given by \eqref{eq: Aguiar control law for gammaddot}.
	\end{enumerate}
Then, both the path following error $\x(t)=[\e_{\rmB}(t), e_{\gamma}(t)]$ and the disturbance estimation error $\e_c(t)$ converge to zero as $t\to \infty$.
\end{theorem}
Proof: This result is a consequence of lemmas \ref{lemma: ocean current estimation} and \ref{lemma: aguiar method with current}. \\

\section{Path-following of fully and over-actuated vehicles with arbitrary heading}
\label{sec: fully actuated}
In this section we consider path following for fully-actuated vehicles whose motion is described by the kinematics model \eqref{eq: kinematics in 2D no current}. We recall that the longitudinal and lateral speeds and heading rate of fully actuated vehicles can be controlled independently. As we shall see, this allows this class of vehicles to follow paths with arbitrary heading assignments. \\
We start by noticing that the thruster configuration of a vehicle dictates which kinematic variables can be tracked with appropriately designed  inner loop controllers. Consider only motions in the 2D plane and let the vehicle be equipped with $n$ horizontal thrusters. Let $\f:=[f_1, f_2, …, f_n]^{\top} \in \R^{n} $ denote the vector of forces, where $f_i \in \R$ is the force generated by thruster $i, i=1,...,n$. Following the SNAME convention, the resulting horizontal forces and torque that they impart to the vehicle can be represented by $F:=[F_{\rm x},F_{\rm y}]^{\top} \in \R^2$ and $N\in \R$, respectively. Let ${\bs\tau}=[F^\top, N]^{\top} \in \R^3 $ denote the total force and toque produced by all thrusters, then it can represented as ${\bs \tau} = K\f$, where $K \in \R^{3\times n}$ is the thrust allocation matrix that depends on the configuration (both in position and orientation) of the thrusters \cite{Fossen09}. In what follows we assume that $K$ is full row rank and $n \ge 3$, with $n=3$ and $n > 3$ corresponding to a fully actuated and an overactuated vehicle, respectively. In both cases, given a desired vector ${\bs \tau}$ requested by the vehicle's inner loop controllers, and assuming there are no physical limitations on the force vector $\f$, then $\f=K^*\bs{\tau}$ where $K^*$ is the either the inverse or the Moore pseudo-inverse of $K$ for $n=3$ and $n > 3$, respectively \cite{Fossen09,Tor2013Over}. The more realistic case where there are physical limitations on the force vector $\f$ can be dealt with by solving a constrained optimization problem. Finally, assuming that the force $f_i$ generated by each thruster is approximately given by  a quasi steady-state invertible map $f_i = g(u_i)$, where $u_i$ is the control input (e.g. speed of rotation of propeller $i$), then $u_i=g^{-1}(f_i)$. \\
For under-actuated vehicles, a proper thruster configuration will only allow for the generation of surge force $F_{\rm x}$ and yaw torque $N$, that is, all the elements of the second row of $K^*$ are zero. In this situation, the vehicle's inner loops can control the surge speed $u$ and the yaw $\psi$ or yaw rate $r$ but not sway speed $v$. For fully actuated vehicles in 2D, since $K$ is full rank, it is also possible to generate a force in sway, $Y$, which allows for the implementation of an inner loop to control sway speed $v$. Therefore, we can assign independent values for the total velocity vector ${\bf v}=[u,v]^{\top}$ and yaw $\psi$. The yaw $\psi$ can be assigned independently as a desired value $\psi_{\rm d}$ which may be constant or dependent on the path-following variable $\gamma$ or time. In view of this, path following with arbitrary heading can be achieved by commanding the total velocity vector in such a way as to make the evolution of $\p$ comply with the path following objectives, that is, converge to a desired path and track a desired velocity  profile. 
Based on the theory exposed in \emph{Method 6}, let $\pd(\gamma)$ be the \emph{``reference point"} on the path that the vehicle must track to achieve path following. Define
\begin{equation} \label{eq: eB fully}
\begin{split}
\e_{\subscript{\B}}=\RIB(\psi)(\p-\p_{\rm d}) 
\end{split}
\end{equation}
as the position error between the vehicle and the path in the vehicle's body frame $\Bframe$. Taking the time derivative of \eqref{eq: eB fully} and using \eqref{eq: kinematics in 2D no current} and the fact that $\RBI(\psi)=[\RIB(\psi)]^{\top}$ yields 
\begin{equation} \label{eq: following fully actuated}
\begin{split}
\dot{\e}_{\subscript{\B}}&=[\dot{R}^{\subscript{\I}}_{\subscript{\B}}(\psi)]^{\top}(\p - \pd)+\RIB(\psi)(\dot{\p} - \dot{\p}_{\rm d})\\
&= −S(\bs{\omega}){\e}_{\subscript{\B}} + {\bf v} − \RIB(\psi)\p'_{\rm d}(\gamma)\dot{\gamma}.
\end{split}
\end{equation}
To address the dynamic task stated in  \eqref{eq: dynamic task for gamma}, define the speed tracking error of the path parameter $e_{\gamma}$ as in \eqref{eq: e gamma}. If $e_{\gamma}$ can be driven to zero then the dynamics task is fulfilled. Let ${\bf x}=[\e^{\top}_{\rmB}, e_\gamma ]^{\top}\in \R^{3} $ be the complete path following error vector. From \eqref{eq: following fully actuated} and \eqref{eq: gamma dot}, the dynamics of the path following error vector are given by   
\begin{equation} \label{eq: complete path following error fully actuated}
\begin{aligned}
\dot{\bf x}
= \left[\begin{matrix}
−S(\bs \omega){\e}_{\subscript{\B}} +{\bf v} − \RIB(\psi)\p'_{\rm d}(\gamma)\dot{\gamma}\\
\ddot{\gamma}-\dot{v}_{\rm d}
\end{matrix}\right].
\end{aligned} 
\end{equation}
The main objective now is to derive path following control laws for the vehicle inputs $(u,v,\ddot{\gamma})$ to drive $\bf x$ to zero. 
\begin{theorem} \label{theorem: Fully actuated method}
	Consider the path following error system described by \eqref{eq: complete path following error fully actuated}. Then, the control law for ${\psi}$, ${\bf v}$, and $\ddot{\gamma}$ given by
	\begin{subequations} \label{eq: fully actuated controller}
		\begin{align}
		\psi &=\psi_{\rm d}, \label{eq: Fully actuated law for yaw} \\
		{\bf v}&=\RIB(\psi)\p'_{\rm d}(\gamma)v_{\rm d} -K_{\rm p}{\e}_{\rmB} \label{eq: Fully actuated law for v} \\
		\ddot{\gamma}&=-k_{\gamma}e_\gamma + \e^{\rm T}_{\rmB}\RIB({\psi})\p'_{\rm d}(\gamma) +\dot{v}_{\rm d},  \label{eq: Fully actuated law for gammaddot}
		\end{align}
	\end{subequations} 
	render the origin of $\x$ GES, where $K_{\rm p}$ is a positive definite matrix with appropriate dimensions, and $k_{\gamma} >0$.
\end{theorem}
Proof: See Appendix - in Section \ref{Proof of theorem: Fully Actuated}. \\
In summary, this path following method can be implemented using Algorithm \ref{alg: PF method with arbitrary heading}.
\begin{algorithm}[!hbt] 
	\caption{PF Algorithm for fully actuated vehicles} 	
	\begin{algorithmic}[1]
		\State Initialize $\gamma(0)$ and $\dot{\gamma}(0)$
		\State For every sampling interval, repeat the following procedure:
		\Procedure{PF CONTROLLER }{}	
		\State Set the yaw angle $\psi$ to the desired value $\psi_d$ as in \eqref{eq: Fully actuated law for yaw}.    
		\State Compute position error $\eB$ using \eqref{eq: eB fully} and tracking error $e_{\gamma}$ using \eqref{eq: e gamma}.    
		\State Compute desired vehicle's velocity $\textbf{v}$ using \eqref{eq: Fully actuated law for v}
		\State Compute $\ddot{\gamma}$ using \eqref{eq: Fully actuated law for gammaddot}, then integrate it to update new value of $\gamma$.
		\EndProcedure
	\end{algorithmic}
	\label{alg: PF method with arbitrary heading}
\end{algorithm} 

%% file: SIMULATIONS.tex
{
\color{black}
\section{Matlab Simulation Toolbox and ROS/Gazebo Implementation}
\label{sec:Implementation}
In the scope of this review paper, both a Matlab toolbox and a set of ROS packages were developed to test and analyze the performance of the path following algorithms described in the previous sections. The end goal is not only to have simulation and field-trial results included in the paper, but also to give the reader tools that allows testing quickly the path following algorithms and integrate them in guidance, navigation, and control systems.
\subsection{Matlab toolbox}
The path following Matlab toolbox can be accessed at \url{https://github.com/hungrepo/path-following-Matlab/tree/master/PF-toolbox}. The main purpose of this toolbox is to give the reader a simple tool for testing quickly the path following methods reviewed in the present paper. This toolbox implements the seven methods described in Section \ref{sec: pf method in 2D}. The source code is open and therefore anyone can join to use, modify or develop new functionalities on top of it. In this toolbox “PFtool.m” is the main file that initializes the simulation and runs the main-loop. In order to test the algorithms with different types of paths one can set values for the variables “path type” and “controller”, see an example in Fig. \ref{fig:pf_toolbox_matlab}.
\begin{figure}[hbt!]
    \centering
	\includegraphics[width=0.50\textwidth]{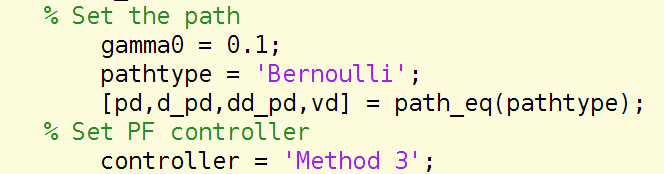}
	\caption{Path following toolbox main file: 'PFtool.m'}
    \label{fig:pf_toolbox_matlab}
\end{figure}
After running the ``PFtool.m" file, the toolbox will animate and plot results as illustrated in Fig. \ref{fig: example of pf toolbox}. See also animation videos at \url{https://youtu.be/XutfsXijHPE}.
\begin{figure}[hbt!]
		\centering
	\includegraphics[trim=100mm 10mm 100mm 18mm, clip,width=120mm]{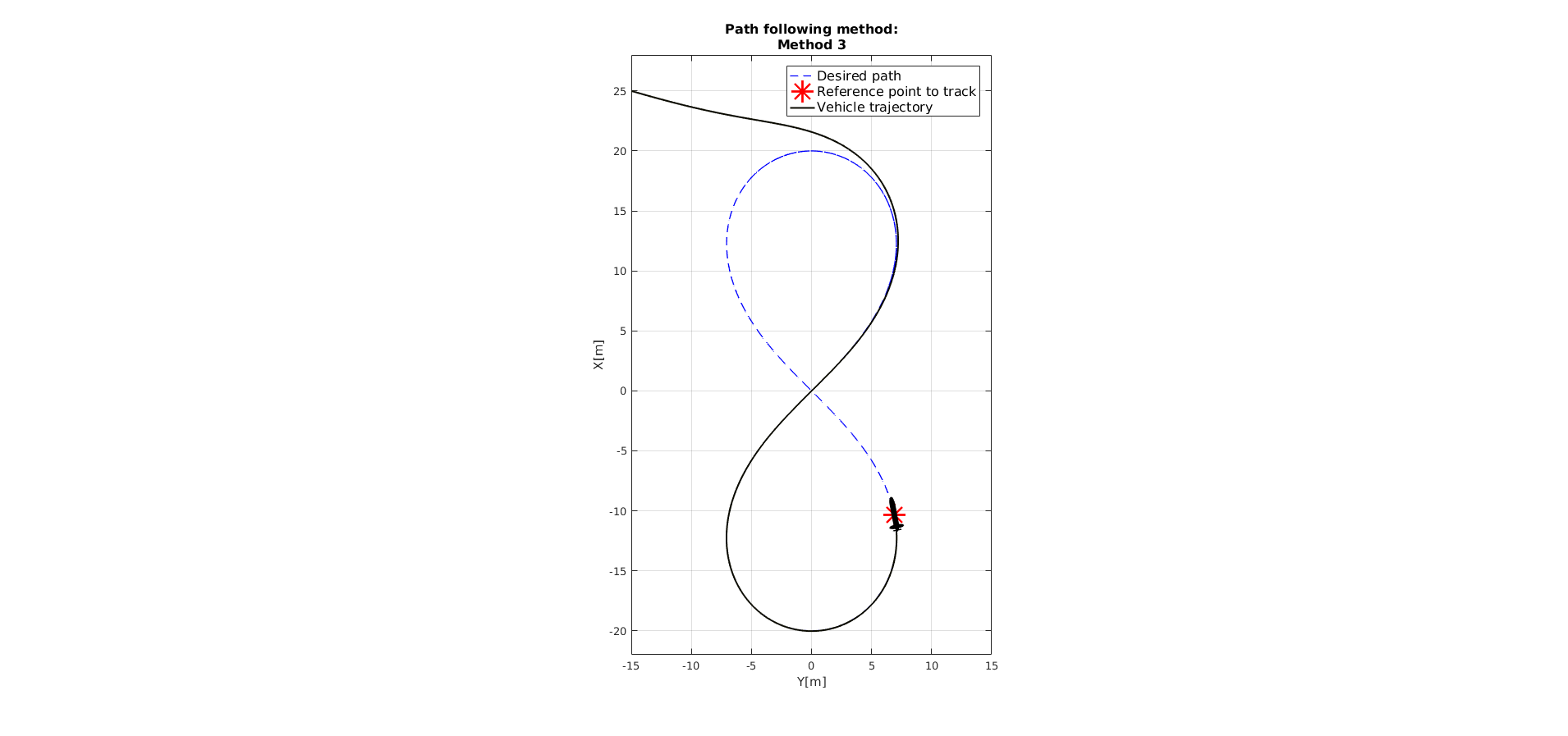} \\ \centering
	a) Example 1: Bernoulli path with Method 3	\\ ~\\
	\includegraphics [trim=50mm 40mm 40mm 40mm, clip,width=150mm]{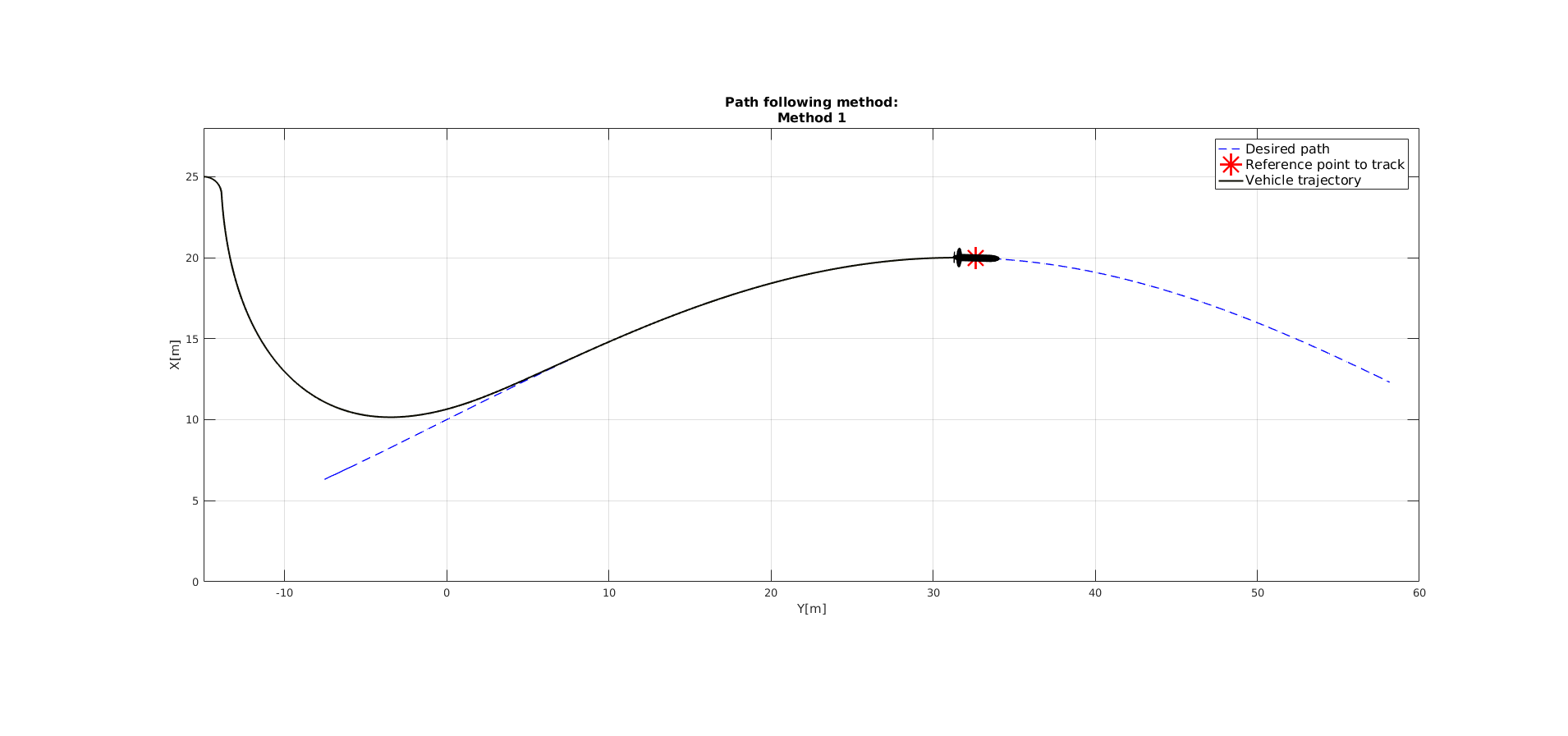}	\\~ \\
		b) Example 2: Sinusoidal path with Method 1 	
	\caption{Simulation example with different paths and path following methods}
	\label{fig: example of pf toolbox}
\end{figure}

\subsection{Gazebo/ROS packages}
In order to test the path following algorithms with the Medusa vehicles in realistic a environment. We developed two efficient ROS packages in C++ that are used to run the algorithms on the on-board computers of the vehicles, namely i) a paths package for the generation of conveniently parameterized planar paths and ii) a path following algorithm package. These code libraries communicate with each other according to Fig. \ref{fig:pf_ros_package}, using ROS topics. The paths library implements four simple parametric consisting of: arcs, lines, circles and Bernoulli lemniscates, as illustrated in Fig \ref{fig:paths_ros_package}.

\begin{figure}[hbt!]
    	\centering
		\includegraphics[width=0.5\textwidth]{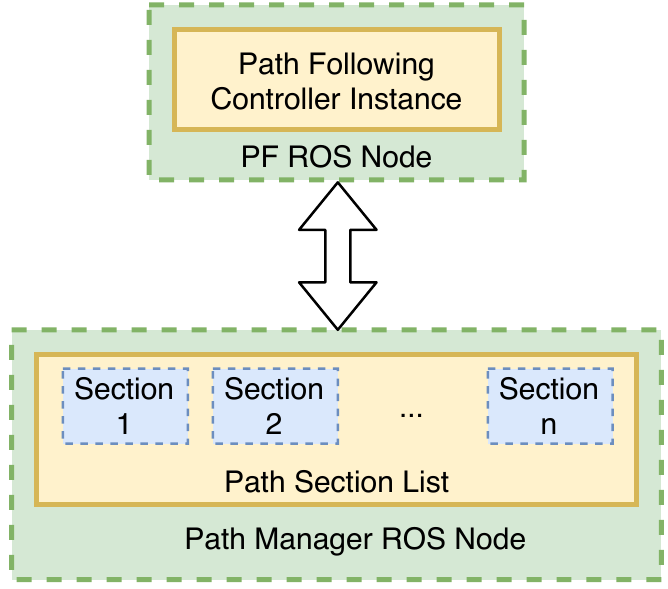}	
		\caption{ROS node abstraction}
    	\label{fig:pf_ros_package} 
\end{figure}
\begin{figure}[hbt!]
		\centering
		\includegraphics[width=0.6\linewidth]{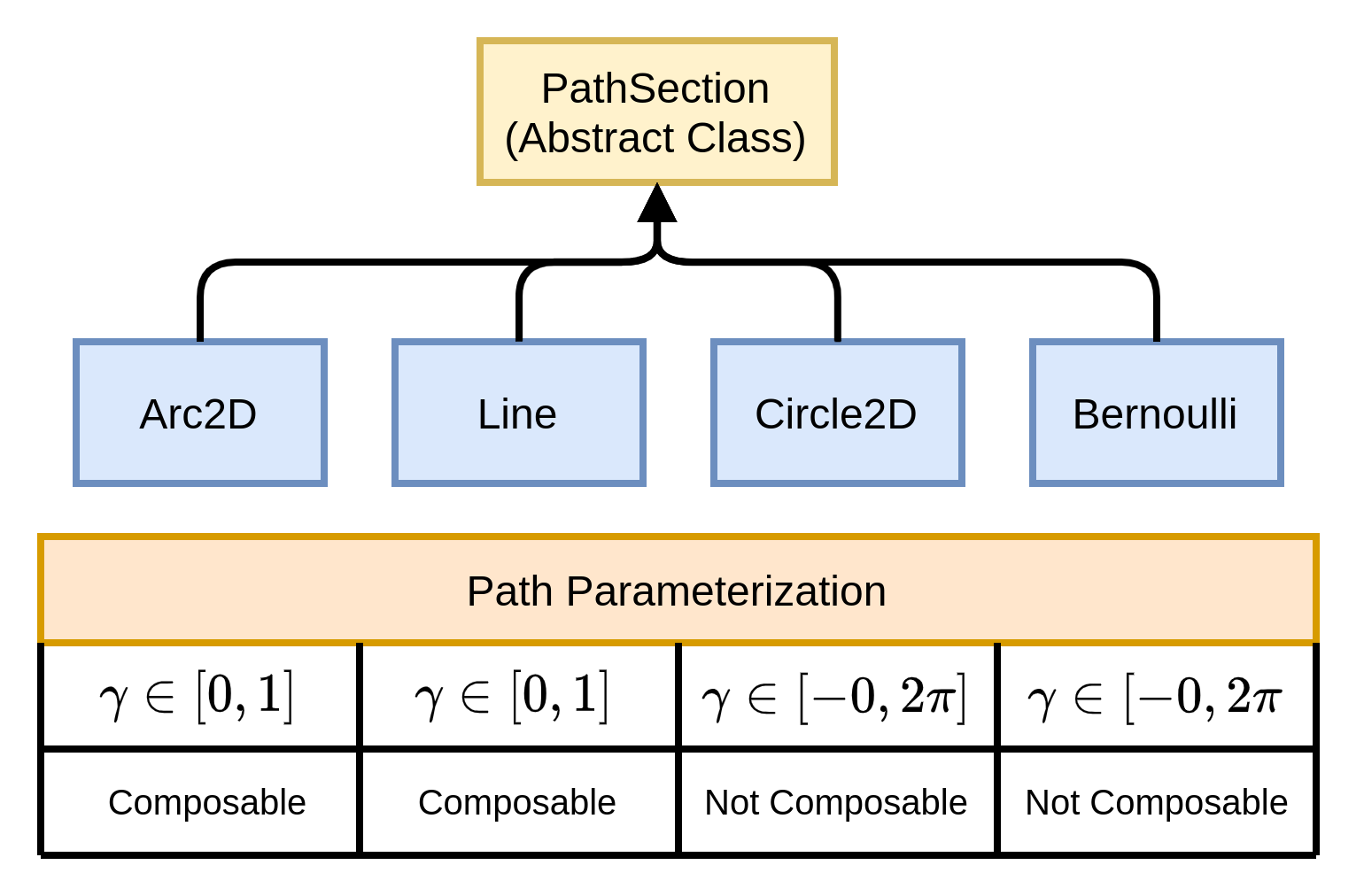}
		\caption{Paths Library}
    	\label{fig:paths_ros_package} 
\end{figure} ~\\
The paths library works as a state machine, such that multiple segments that are marked as ``composable" can be added together to form more complex shapes. This flexible structure allows for the switching of the path following methods in use and the paths being followed in real time. The complete path following system implemented based on this structure is illustrated in Fig. \ref{fig:vehicle_control_architecture}.
\begin{figure}[hbt!]
    \centering
	\includegraphics[width=0.97\textwidth]{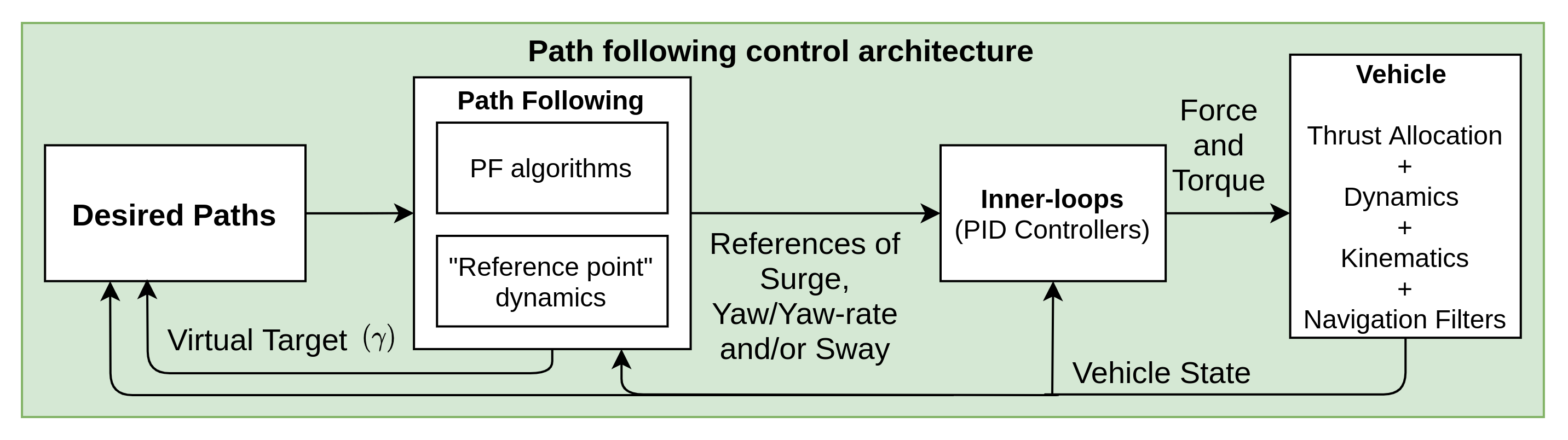}
	\caption{Complete path following control system implemented in Medusa vehicles}
    \label{fig:vehicle_control_architecture}
\end{figure} ~\\
To allow the reader to perform experiments similar to the ones presented in this paper we provide in \url{https://github.com/dsor-isr/Paper-PathFollowingSurvey} a realistic simulation environment, resorting to the UUVSimulator Plugin and Gazebo 11 simulator \cite{UUVSim}, widely used in the robotics community. This makes it possible for the reader to simulate the real conditions found a the Olivais Dock at Parque das Nações, Lisbon, during field trials.
\begin{figure}[hbt!]
    \centering
	\includegraphics[width=0.7\textwidth]{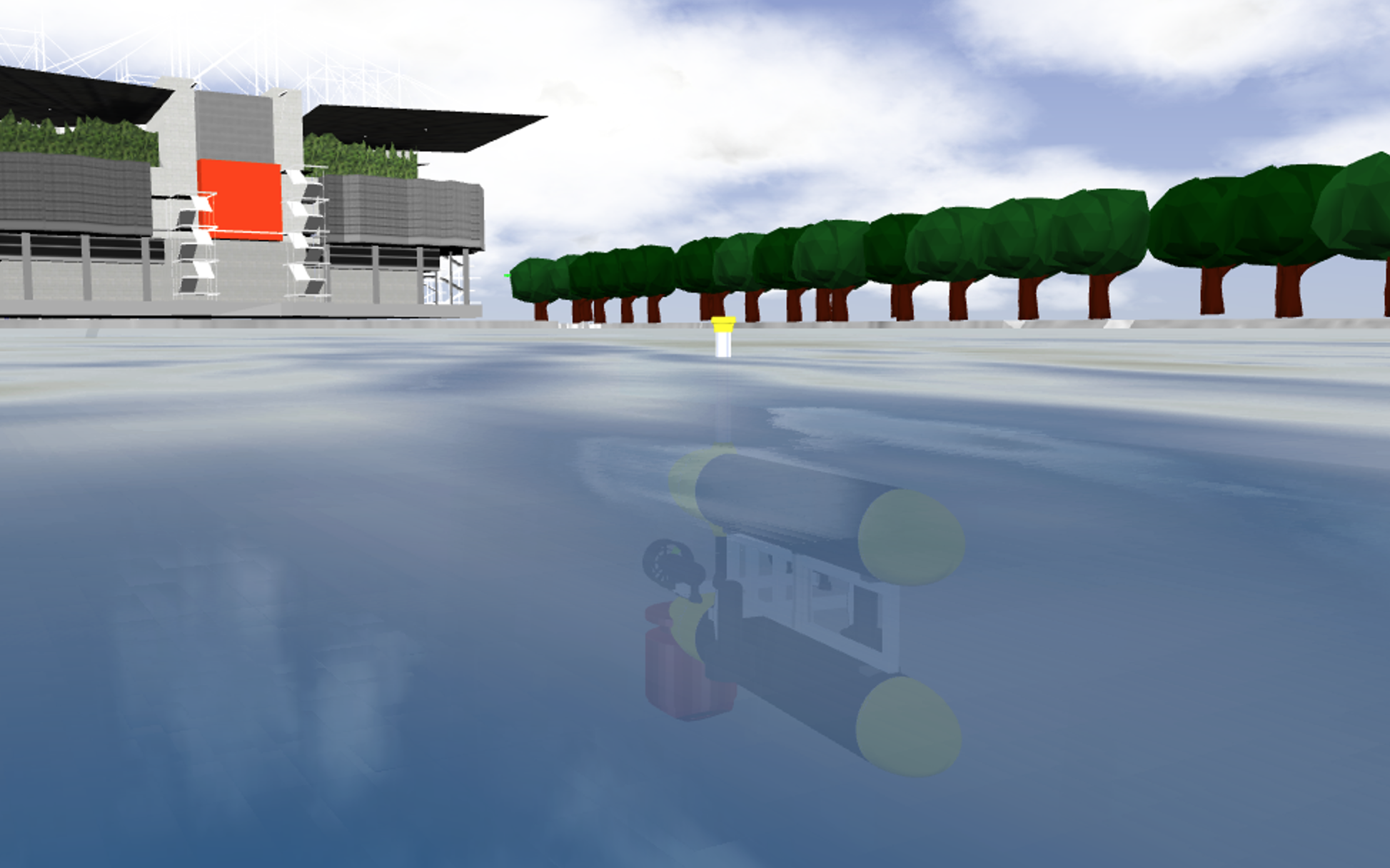}
	\caption{Medusa 3-D simulation using Gazebo}
    \label{fig:gazebo_screenshot}
\end{figure}
}

%% file: EXPERIMENTS.tex
\section{Simulation and experimental results} 
\label{sec: experiement}
In this section, we start by introducing the test set-up adopted for conducting trials with the Medusa class of marine vehicles. Following suit, Hardware-in-the-Loop (HIL) simulation results obtained by resorting to Gazebo simulator are presented for the case of a an over-actuated Medusa vehicle, followed by real field trials results. 

\subsection{Test set-up}
\begin{figure}[h!]
	\centering
	\includegraphics[width=.47\textwidth]{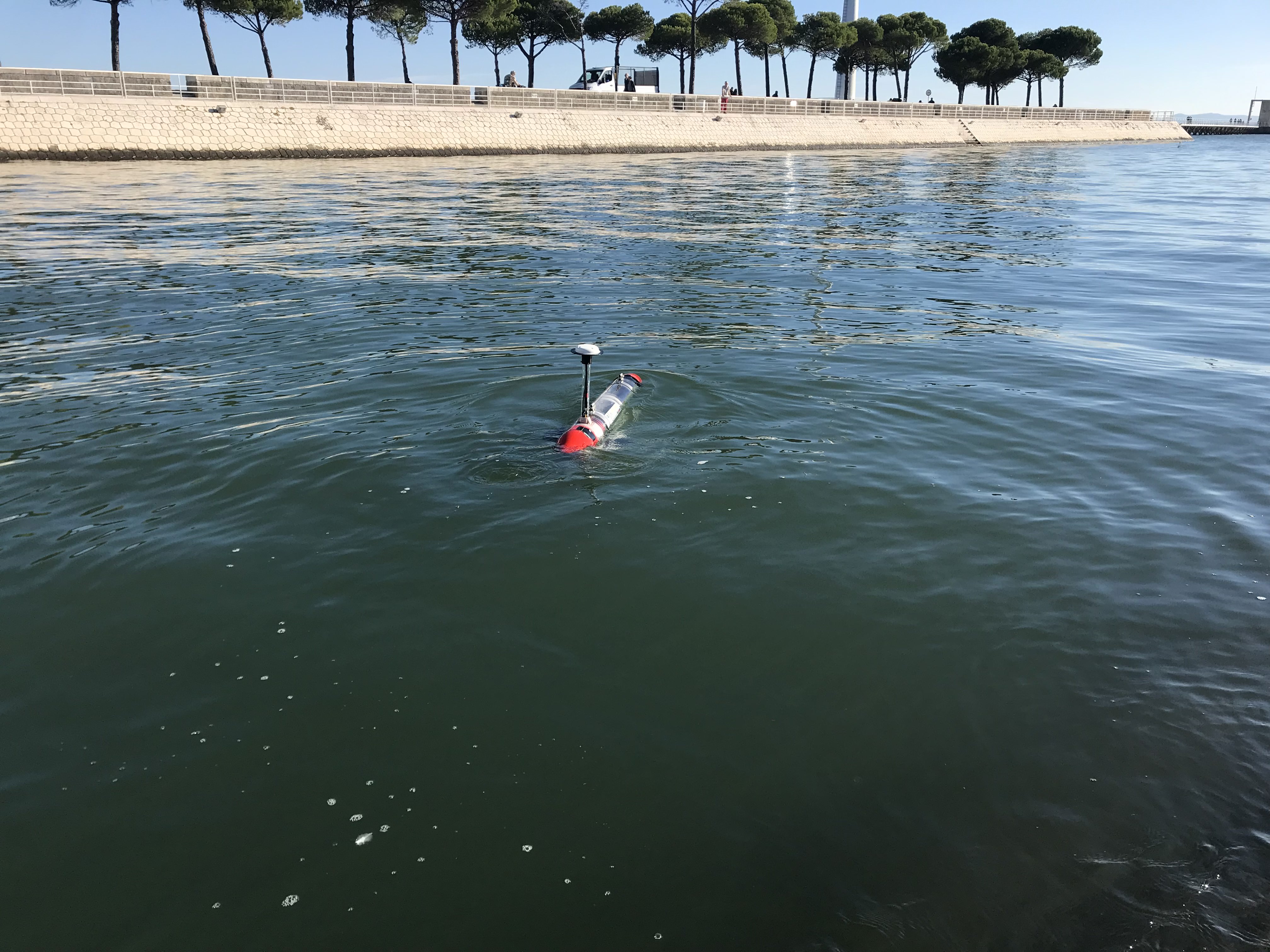} \hfill
	\includegraphics[trim=20.5cm 8.5cm 1.5cm 3cm,width=.47\textwidth,clip]{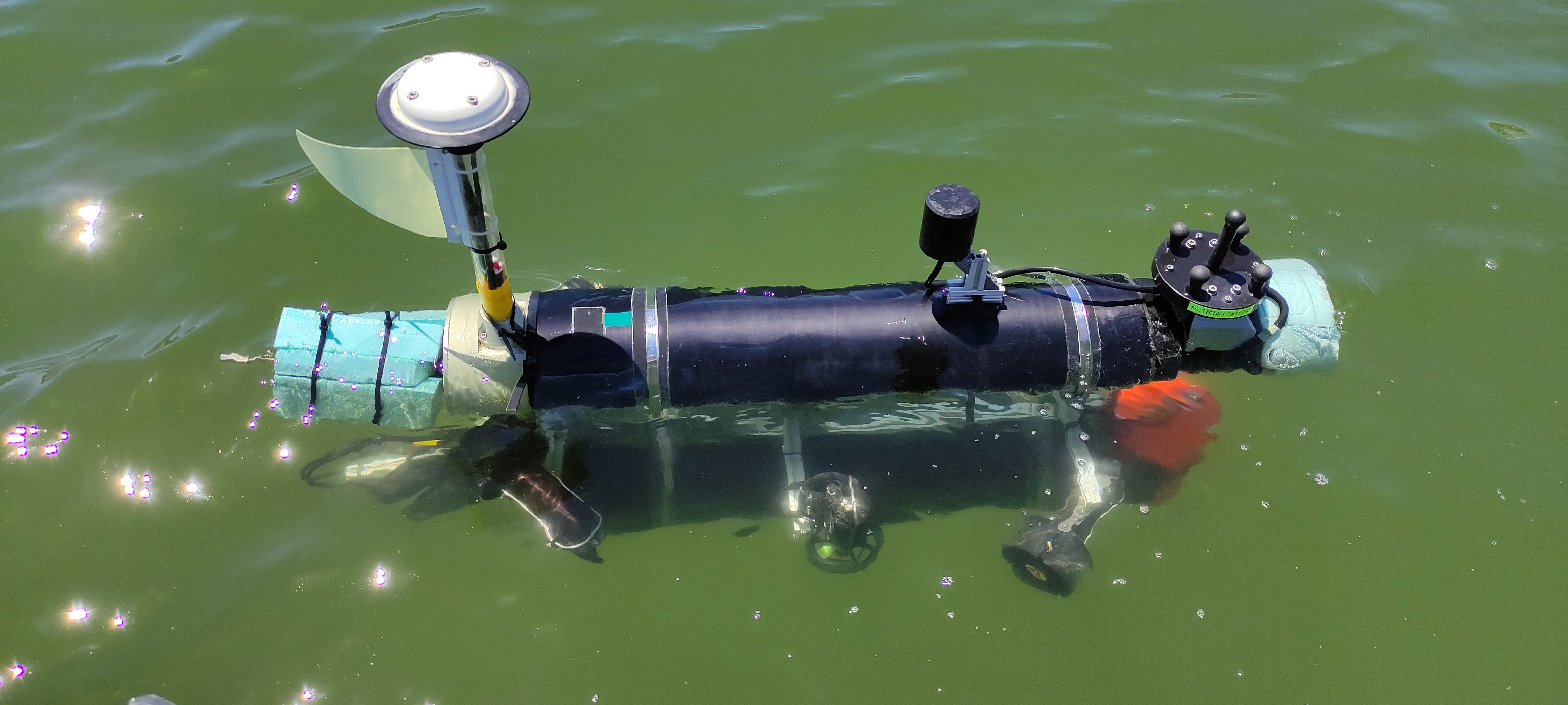} \vfill
	\caption{The Medusa vehicles performing path following during the trial. Left: under-actuated Medusa (a closer view can be seen in Fig. \ref{fig: underactuated vehicles}(a)), Right: over-actuated Medusa.  }
	\label{fig: medusa at trials}
\end{figure}

\begin{figure}[h!]
	\centering
	\includegraphics[trim=0.0cm 0.0cm 0.0cm 0cm,width=0.9\linewidth,clip]{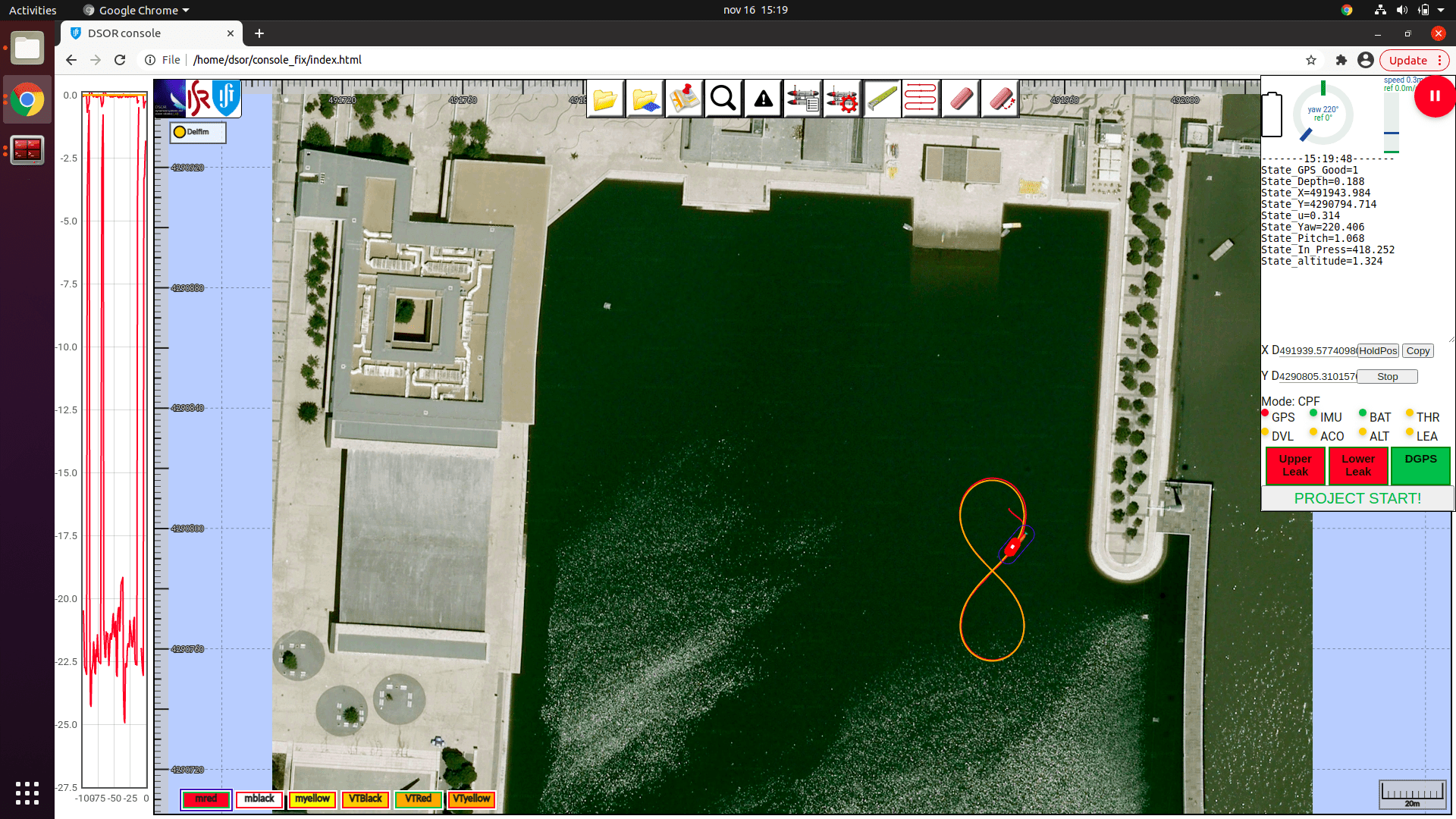} 
	\caption{Console used to operate the Medusa class vehicles}
	\label{fig:console}
\end{figure}

The field tests were conducted with two Medusa class autonomous surface vehicles built at IST. The first one is an under-actuated Medusa shown in Fig.\ref{fig: underactuated vehicles}. This type of vehicle has played central roles in many European research projects such as Morph\cite{PedroMorph} and Wimust \cite{PedroWimust} to support the research in marine science and also in the area of guidance, navigation and control of single and networked multiple autonomous vehicle. The under-actuated Medusa vehicle is equipped with two thrusters located at starboard and portside that allow for the generation of longitudinal forces and torques about the vertical axis. The second Medusa that was used in the trial is over-actuated that is shown in Fig.\ref{fig: fully actuated-vehicle}. This vehicle contain six thrusters. Two thrusters are mounted vertically and four thrusters are mounted horizontally. The thruster configuration in the over-actuated vehicle allows for the generation of longitudinal and lateral forces and torque about the vertical axis. \\
Each vehicle's navigation system builds upon GPS-RTK and AHRS sensors that allow for the computation of its position and orientation. All software modules for navigation and control were implemented using the Robot Operating System (ROS, Melodic version), programmed with C++, and ran on an EPIC computer board (model NANO-PVD5251). More information about the specification of the Medusa class vehicle can be found in \cite{abreu2016medusa}. The implementation of the path following controllers follows the structure depicted in Fig. \ref{fig: path following system}. The vehicle's inner-loop controllers include four proportional integrator and derivative (PID)-type controllers to track the references in linear speed, heading, heading rate, and sway (only for over-actuated vehicles) that are generated by the path following controllers. \\
Snapshots showing the vehicle performing path following can be seen in Fig.\ref{fig: medusa at trials} while its operation can be monitored online with a console shown in \ref{fig:console}. 

\subsection{Simulation results}
\noindent Prior to operating any vehicle in a real world environment, it is common practice to resort to HIL simulations to analyse the performance of the controllers developed. In Figures \ref{fig:medusa_gazebo_xy} and \ref{fig:medusa_gazebo_vector_errors} we can observe a lawn-mowing mission executed by an over-actuated Medusa vehicle, in the Gazebo simulator, using the path following method proposed in Section \ref{sec: fully actuated}. In particular, Figure \ref{fig:medusa_gazebo_xy} shows the X-Y trajectory of the vehicle, that was requested to keep its yaw angle offset by $45^{\circ}$ from the tangent to the path. \\
\begin{minipage}{.48\textwidth}
  \captionsetup{width=.9\linewidth}
  \centering
  \includegraphics[width=0.95\linewidth]{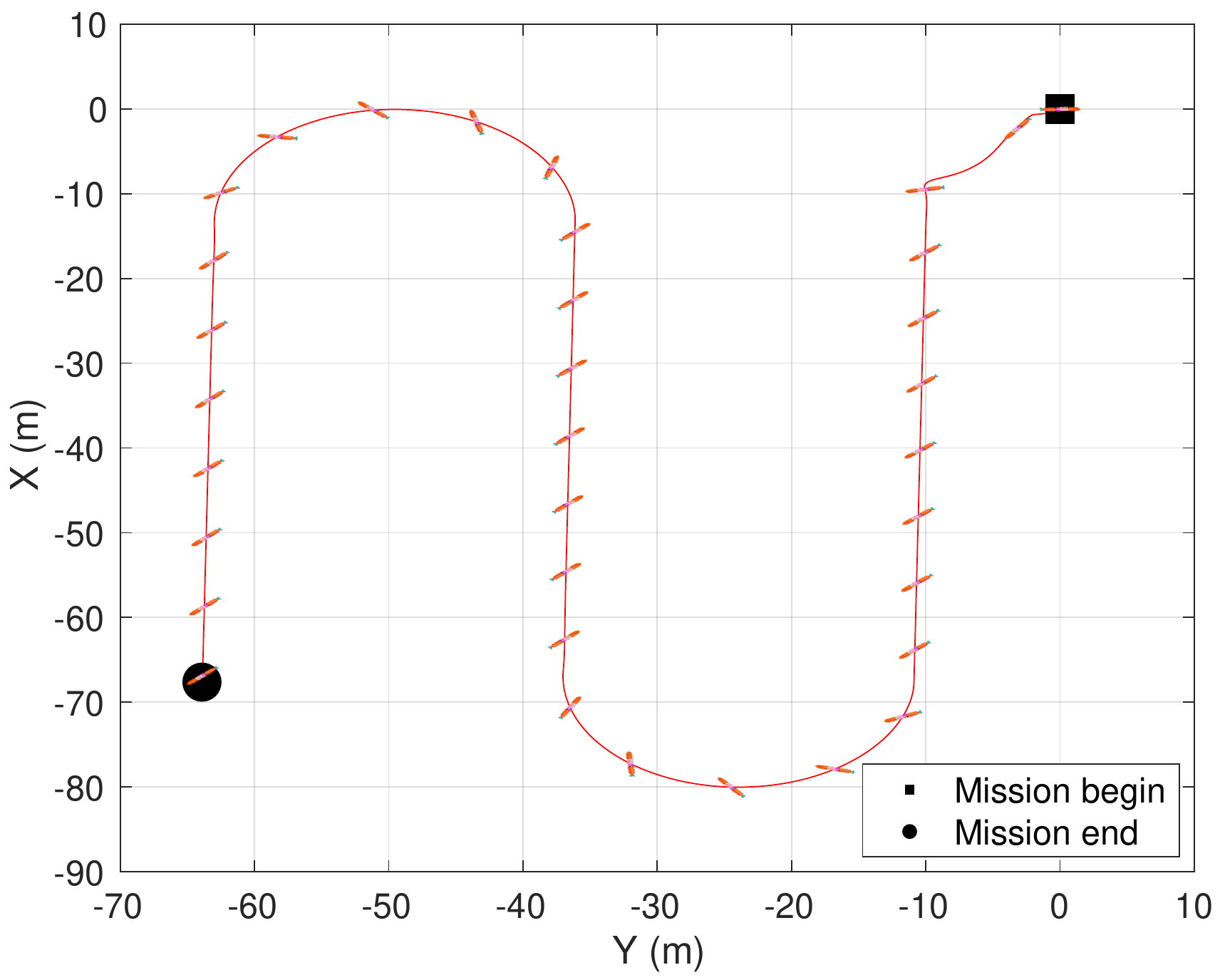} 
  \captionof{figure}{Vehicle trajectory using the method described in Section \ref{sec: fully actuated}.}
  \label{fig:medusa_gazebo_xy}
\end{minipage}
\begin{minipage}{.48\textwidth}
  \centering
  \captionsetup{width=.9\linewidth}
  \includegraphics[width=0.95\linewidth]{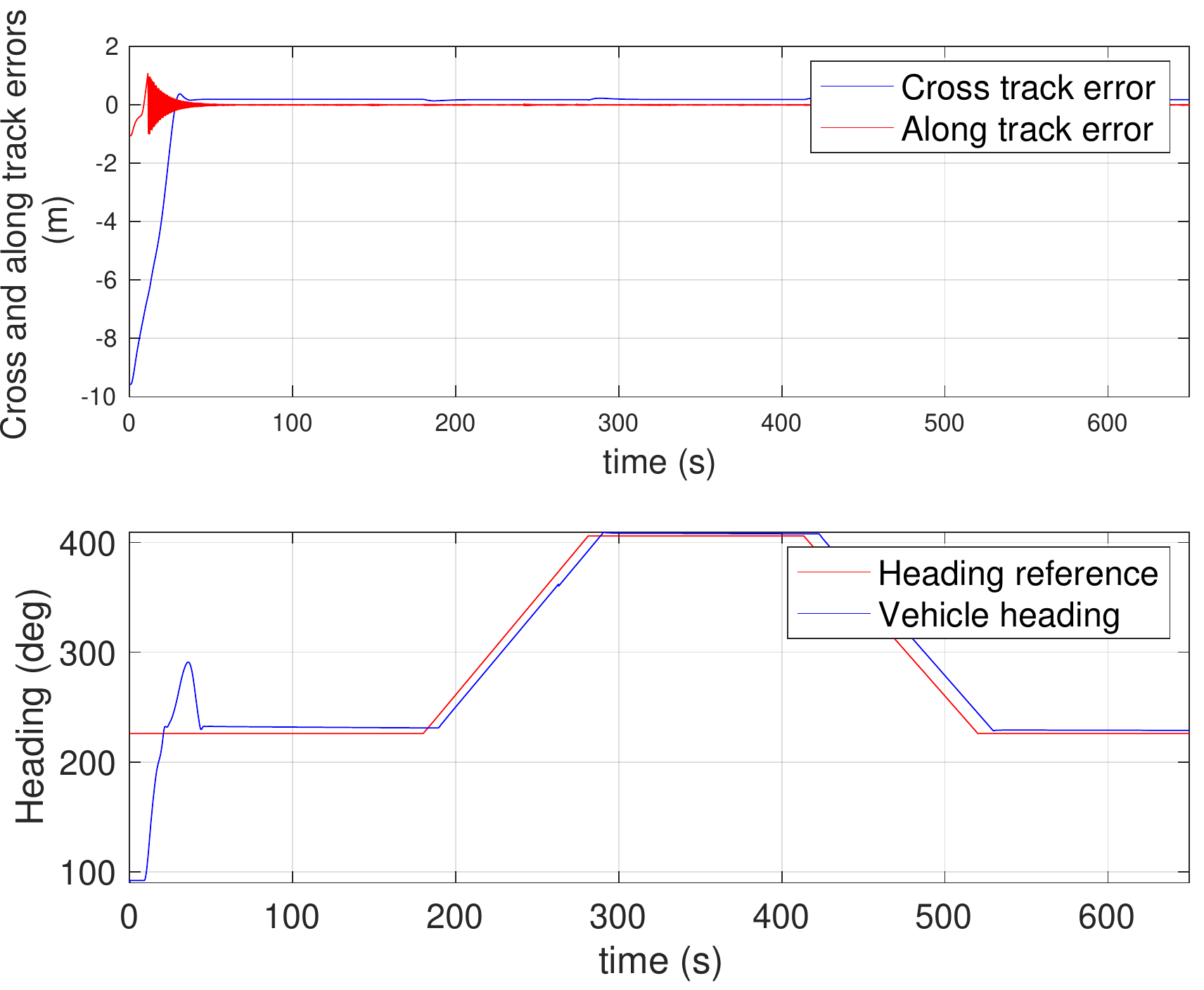} 
  \vfill
  \captionof{figure}{\emph{Cross-track and along-tracks} errors and vehicle heading.}
  \label{fig:medusa_gazebo_vector_errors}
\end{minipage} ~ \\

In Figure \ref{fig:medusa_gazebo_vector_errors} it can be observed that the along-track error converges to zero, while the cross-track and heading errors exhibit small static errors in the neighbourhood of zero. These errors result from the linear inner-loops implemented, a simplification in the design process which does not take into consideration the cross-terms in surge, sway and yaw in the dynamics of the vehicle. 

\subsection{Experimental results}
\subsubsection{Results with an under-actuated vehicle}
\begin{figure}[h!]
	\centering
	\includegraphics[trim=1.5cm 7.5cm 1.5cm 8cm,width=0.7\linewidth,clip]{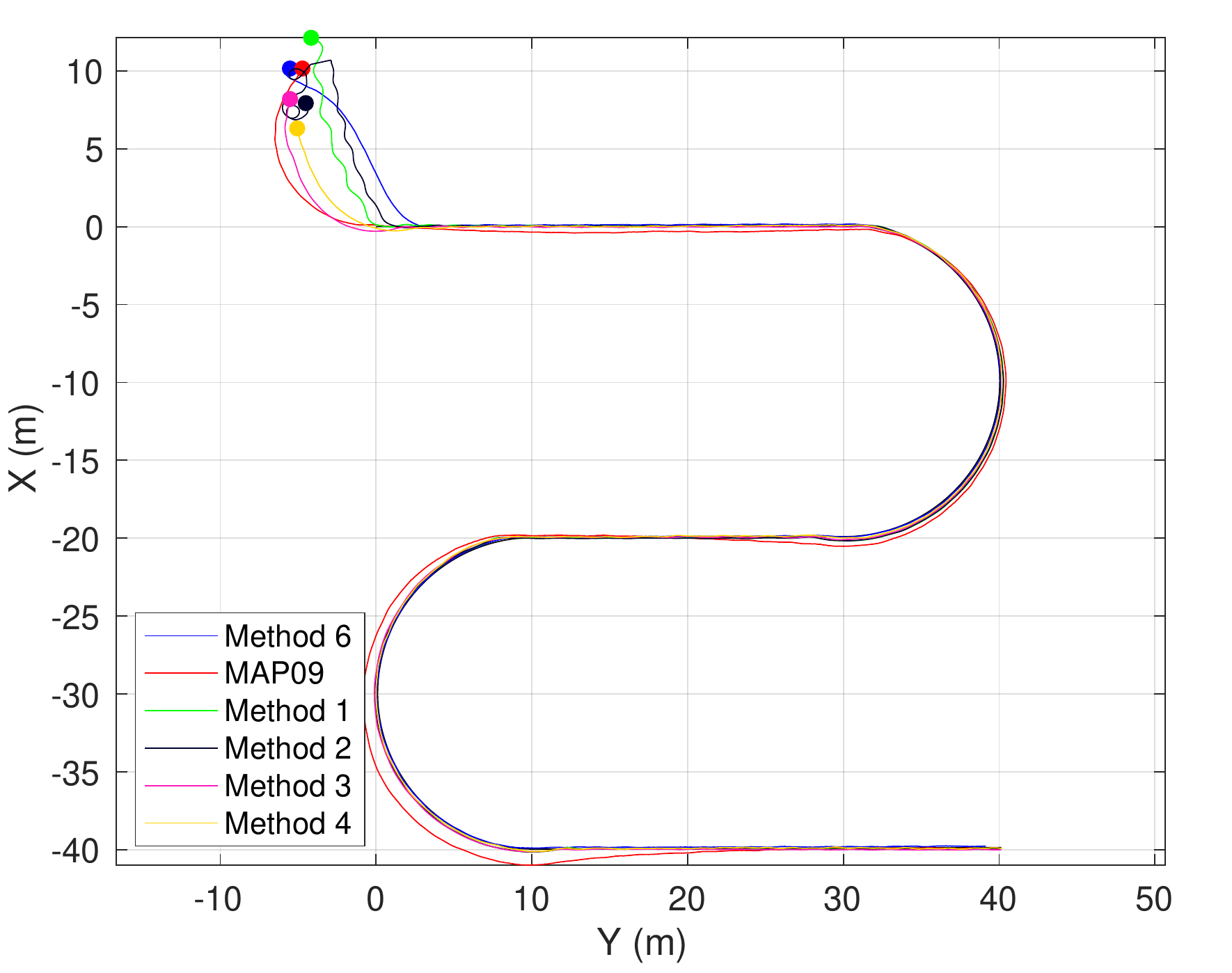} 
	\caption{Vehicle's paths using different path following methods to follow a lawnmower path in  real trials. Filled circles are the positions of the vehicle at the beginning of each run.}
	\label{fig:position_lawnmower}
\end{figure}

\begin{figure}[hbt!]
	\centering
\includegraphics[trim=1.5cm 7.5cm 1.5cm 8cm,width=0.7\linewidth,clip]{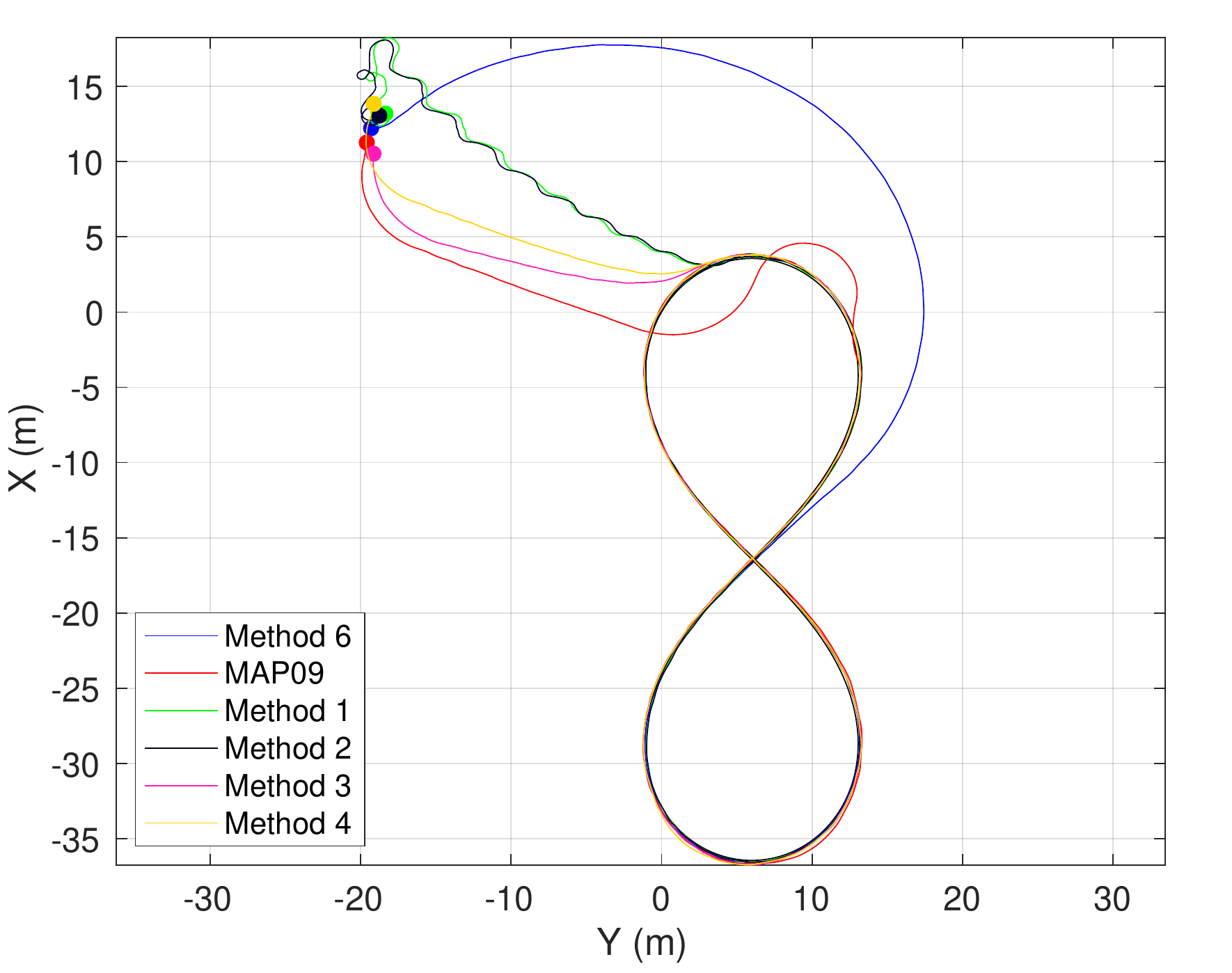} 
	\caption{Vehicle's paths using different path following methods to follow a Bernoulli lemniscate path in real trials. Filled circles are the positions of the vehicle at the beginning of each run.}
	\label{fig:position_bernouli}
\end{figure}

This section describes the results of experimental tests whereby an under-actuated vehicle was requested to follow the same path repeatedly, using different path following algorithms. As shown in Fig. \ref{fig:position_lawnmower}, during the tests the vehicle followed a lawnmowing path with a first leg of $30m$ heading east, then a half circumference turning clockwise with a radius of $10m$, a second leg of $20m$ heading west, another half circumference with a radius of $10m$ turning anticlockwise, and a final leg of $30m$ heading east. In another set of tests the vehicle was requested follow a Bernoulli lemniscate with a length of $20m$ as shown in \ref{fig:position_bernouli}. In all the tests, the assigned speed of the vehicles $U_{\rm d}$ is $0.5m/s$. The trials included the use of \emph{Method} 1-4, 6, and the method in \cite{Pramod2009}. \\
Figures \ref{fig:position_lawnmower} and \ref{fig:position_bernouli} show the paths of the vehicle during the trials, for the different methods. We observe that the vehicles follow their assigned paths and that the transient behavior and tracking performance vary according to the method used.

\begin{figure}[htb!]
\centering
\begin{minipage}{.5\textwidth}
  \captionsetup{width=.9\linewidth}
  \centering
  \includegraphics[trim=1.5cm 7.5cm 1.5cm 8cm,width=\linewidth,clip]{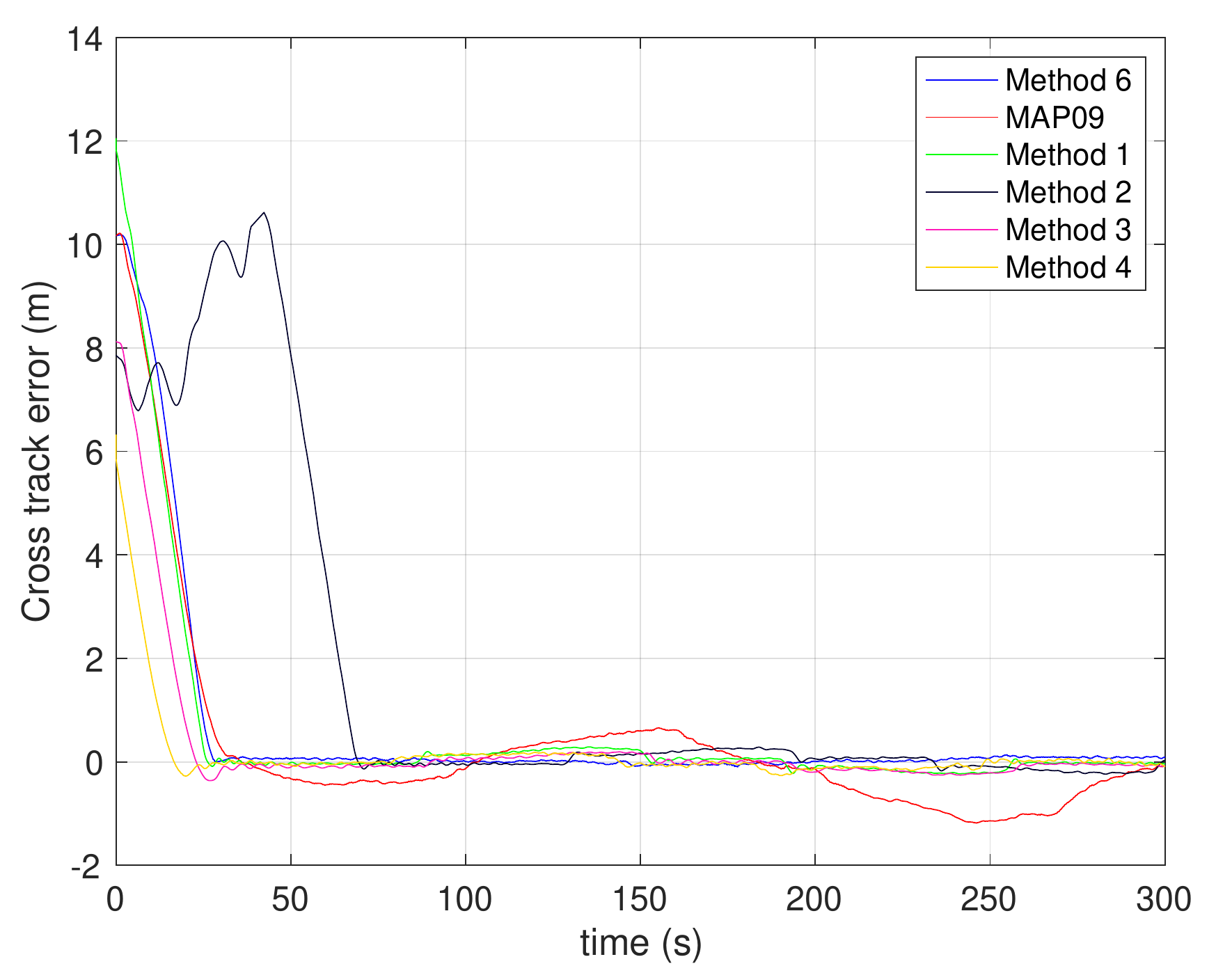} 
  \captionof{figure}{\emph{Cross-track} errors for different methods while performing lawnmower maneuvers.}
  \label{fig:cross_track_lawnmower}
\end{minipage}%
\begin{minipage}{.5\textwidth}
  \captionsetup{width=.9\linewidth}
  \includegraphics[trim=1.5cm 7.5cm 1.5cm 8cm,width=\linewidth,clip]{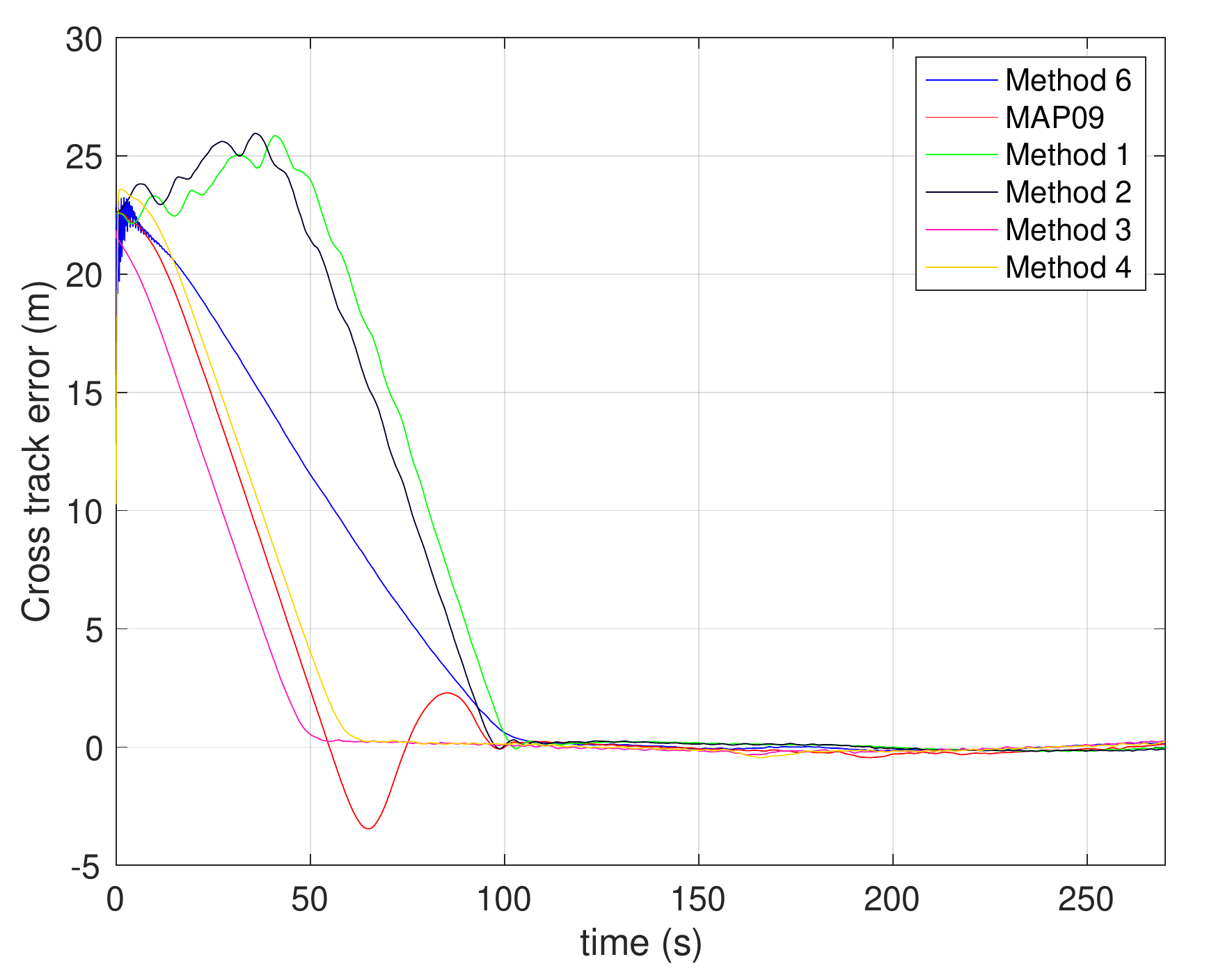} 
  \captionof{figure}{\emph{Cross-track} errors for different methods while describing a Bernoulli lemniscate.}
  \label{fig:cross_track_bernouli}
\end{minipage}
\end{figure}

\begin{figure}[htb!]
\centering
\begin{minipage}{.5\textwidth}
  \captionsetup{width=.9\linewidth}
  \centering
  \includegraphics[trim=1.5cm 7.5cm 1.5cm 8cm,width=\linewidth,clip]{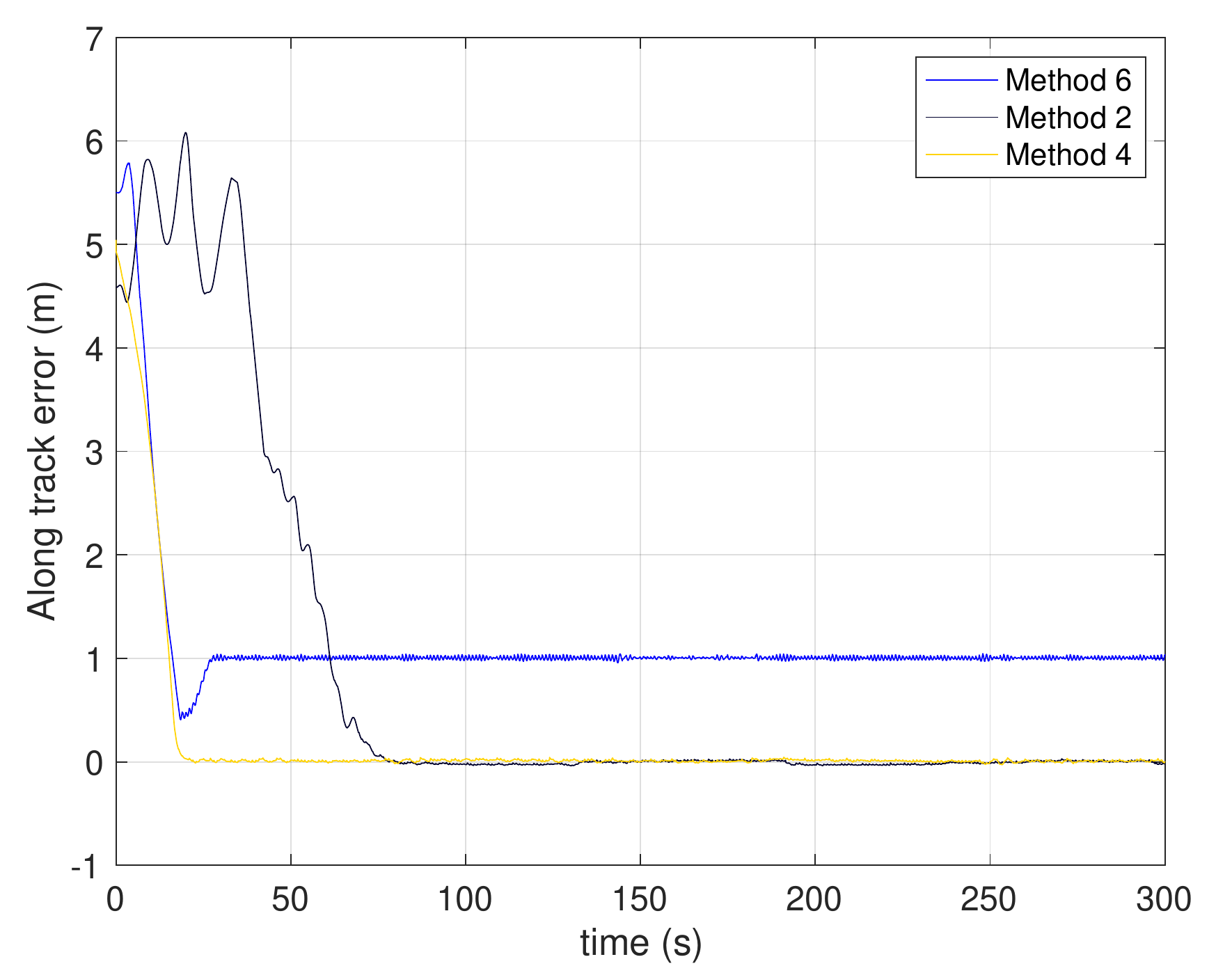} 
  \captionof{figure}{\emph{Along-track}  errors for different methods while performing lawn-mowing maneuvers.}
  \label{fig:along_track_lawnmower}
\end{minipage}%
\begin{minipage}{.5\textwidth}
  \captionsetup{width=.9\linewidth}
  \centering
  \includegraphics[trim=1.5cm 7.5cm 1.5cm 8cm,width=\linewidth,clip]{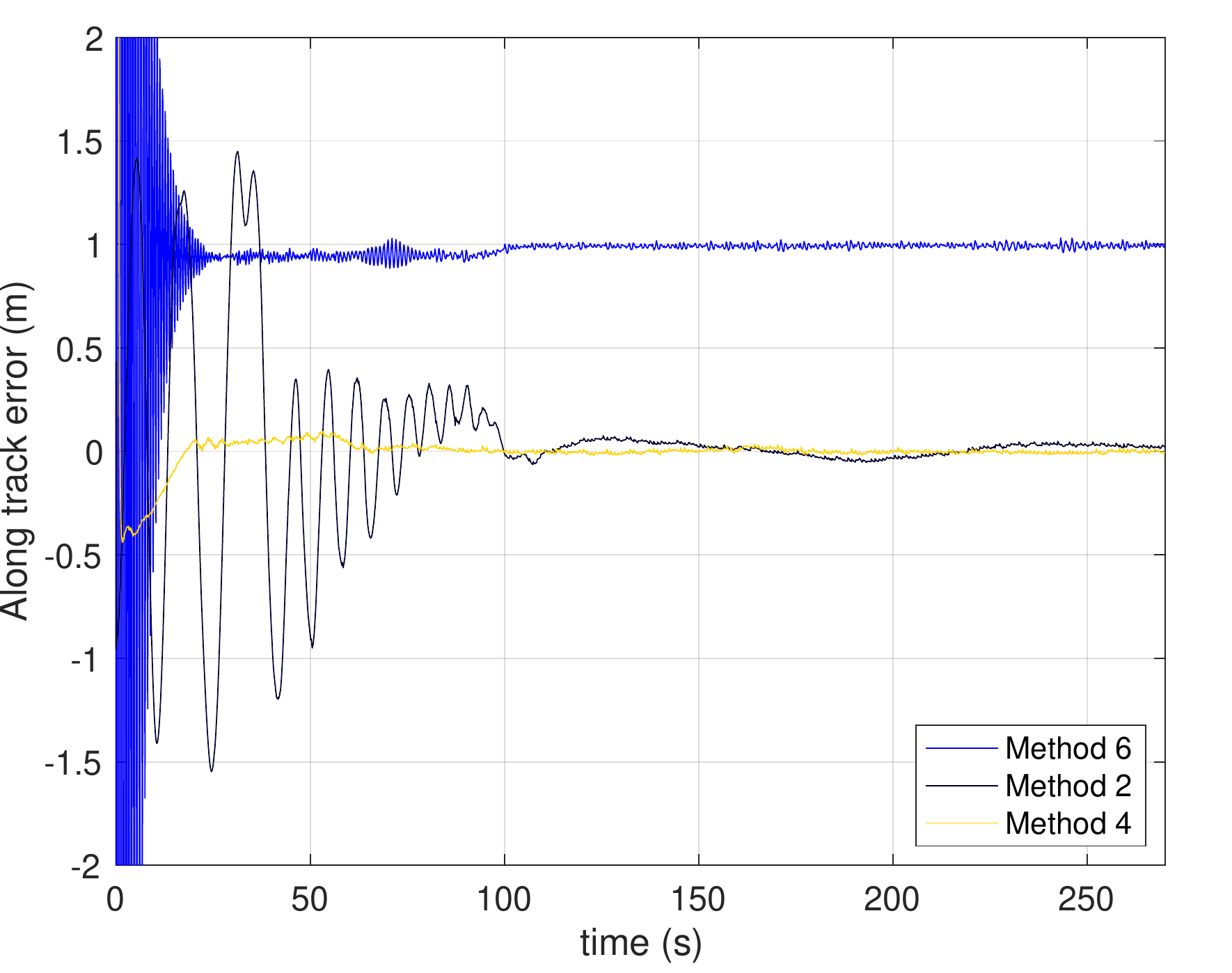} 
  \captionof{figure}{\emph{Along-track} errors for different methods while describing a Bernoulli lemniscate.}
  \label{fig:along_track_bernnouli}
\end{minipage}
\end{figure}

\begin{figure}[htb!]
\centering
\begin{minipage}{.5\textwidth}
  \captionsetup{width=.9\linewidth}
  \centering
  \includegraphics[trim=1.5cm 7.5cm 1.5cm 8cm,width=\linewidth,clip]{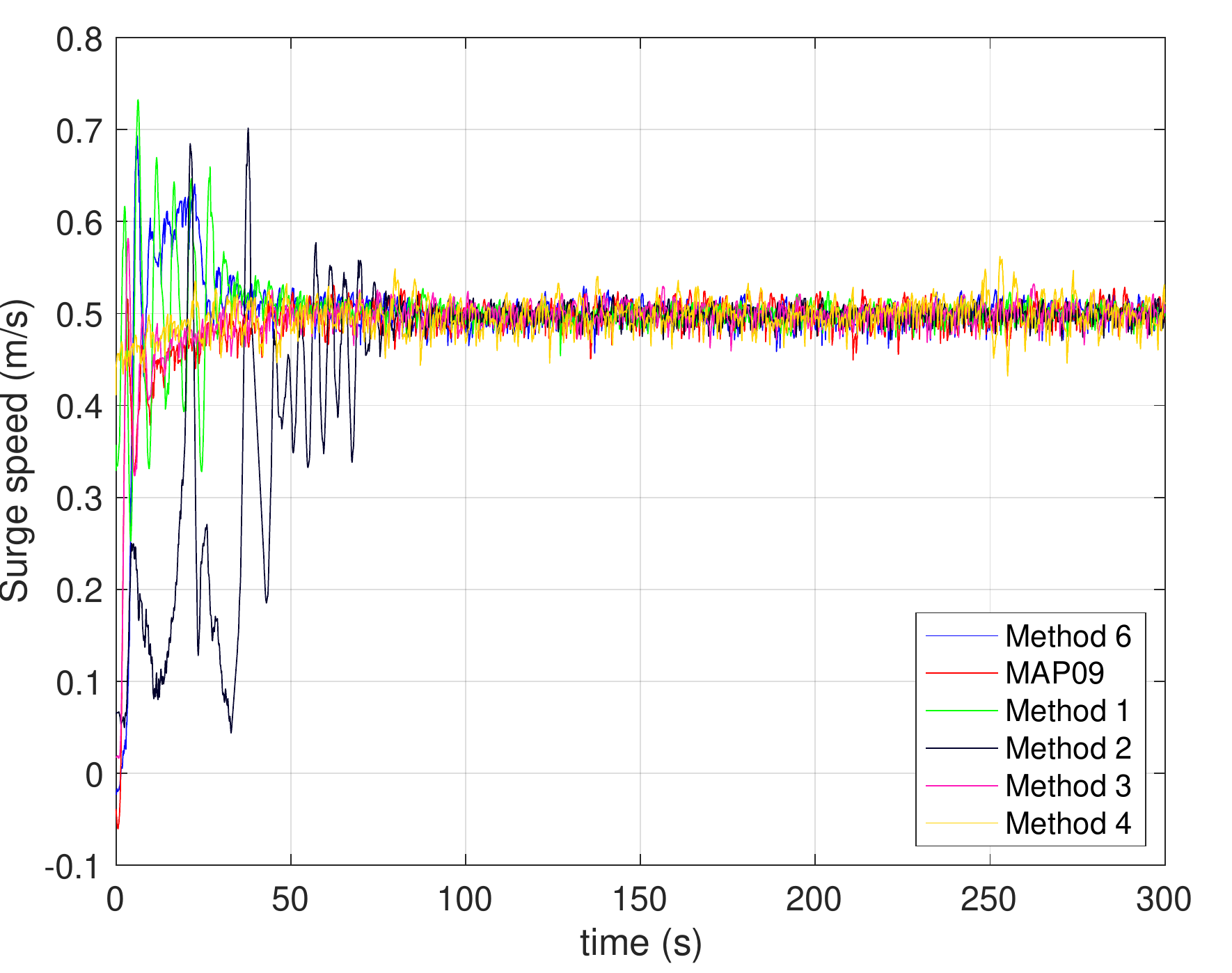} 
  \captionof{figure}{Surge speed for different methods while performing lawnmower maneuvers.}
  \label{fig:surge_speed_lawnmower}
\end{minipage}%
\begin{minipage}{.5\textwidth}
  \captionsetup{width=.9\linewidth}
  \centering
  \includegraphics[trim=1.5cm 7.5cm 1.5cm 8cm,width=\linewidth,clip]{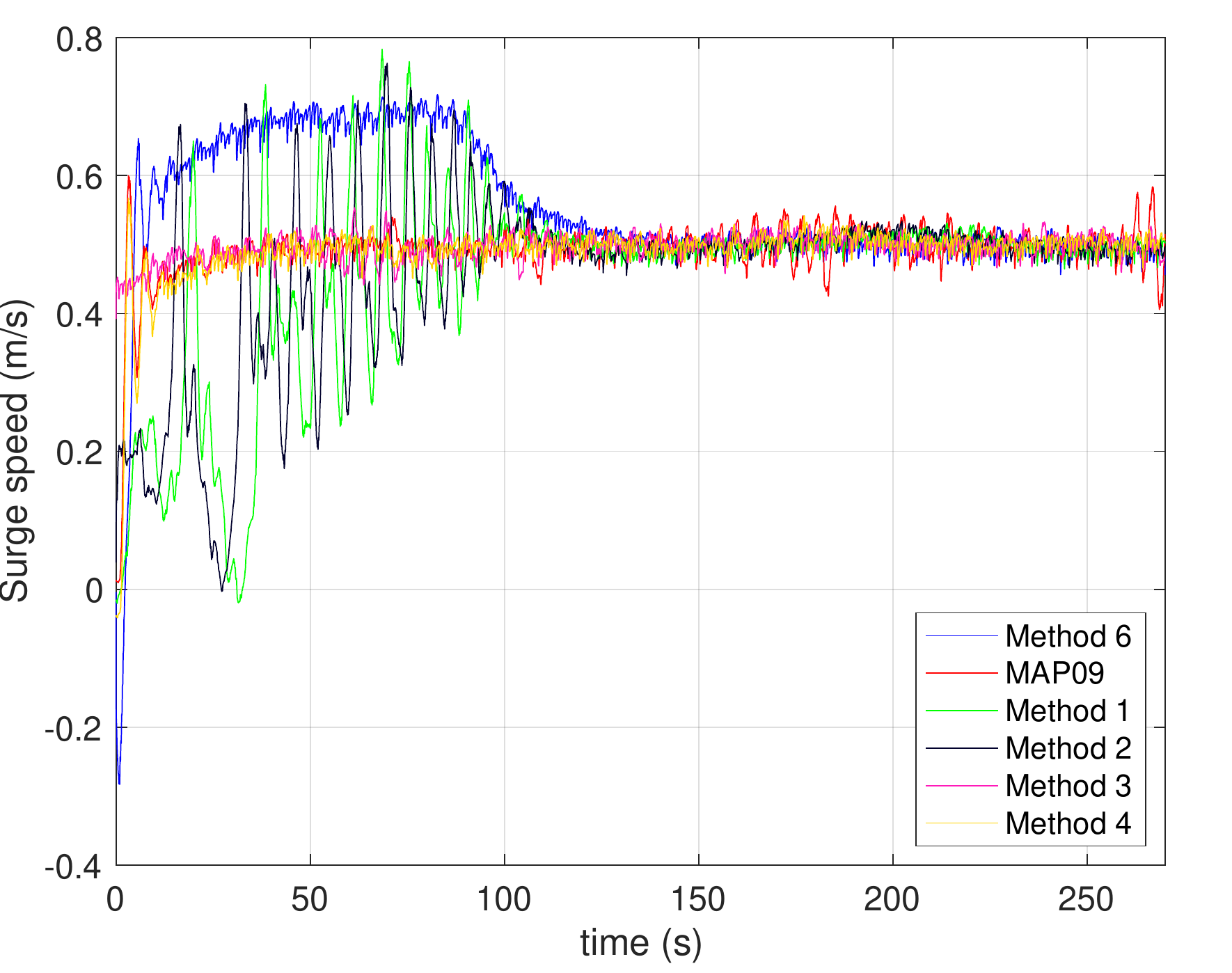} 
  \captionof{figure}{Surge speed for different methods while describing a Bernoulli lemniscate.}
  \label{fig:surge_speed_bernouli}
\end{minipage}
\end{figure}

 ~\\
The \emph{cross-track} errors $y_1$ (defined in Section \ref{section: derivation of path following error in path frame}) of the vehicles performing a lawnmowing maneuver using different path following methods is shown in Figure \ref{fig:cross_track_lawnmower}. Identical results for the case where the path is a Bernoulli lemniscate are shown in \ref{fig:cross_track_bernouli}.
From Figures \ref{fig:cross_track_lawnmower}-\ref{fig:cross_track_bernouli} one can observe that \emph{Methods 3-4} took less time to converge than the other methods. \\
The along-track errors $s_1$, defined in Section \ref{section: derivation of path following error in path frame}, of the vehicle performing a lawnmowing maneuver and following a Bernoulli lemniscate are shown in Fig. \ref{fig:along_track_lawnmower} and Fig.  \ref{fig:along_track_bernnouli}, respectively. Figures \ref{fig:along_track_lawnmower}-\ref{fig:along_track_bernnouli} show the data for \emph{Method 2}, 4 and 6 since for all the other methods the ``\emph{reference point}'' is the orthogonal projection of the vehicle's position on the path which makes the \emph{along-track} error equals to zero.
 ~\\
From Figures \ref{fig:along_track_lawnmower}-\ref{fig:along_track_bernnouli} one can observe that the along track error converges to close to zero for \emph{Method 2} and 4 and for \emph{Method 6} stabilizes at $1m$ due to the effect of $\bs \epsilon$ on the particular definition of the position error, see \eqref{eq: eB}. We can also observe, for the Bernoulli lemniscate mission with \emph{Method 6}, large oscillations at the beginning. \\
Figures \ref{fig:surge_speed_lawnmower} and \ref{fig:surge_speed_bernouli} show the surge speed $u$ of the vehicle for different methods for the lawnmower and Bernoulli lemniscate missions respectively, which clearly indicate that the vehicle asymptotically reaches the desired constant speed profile (0.5m/s). However they show that the surge speed converged faster to the desired speed for \emph{Method} 3, 4 and the Method in \cite{Pramod2009}.

\subsubsection{Results with an over-actuated vehicle}
\begin{figure}[H]
	\centering
	\includegraphics[trim=0cm 7cm 0cm 7.5cm,width=0.6\linewidth,clip]{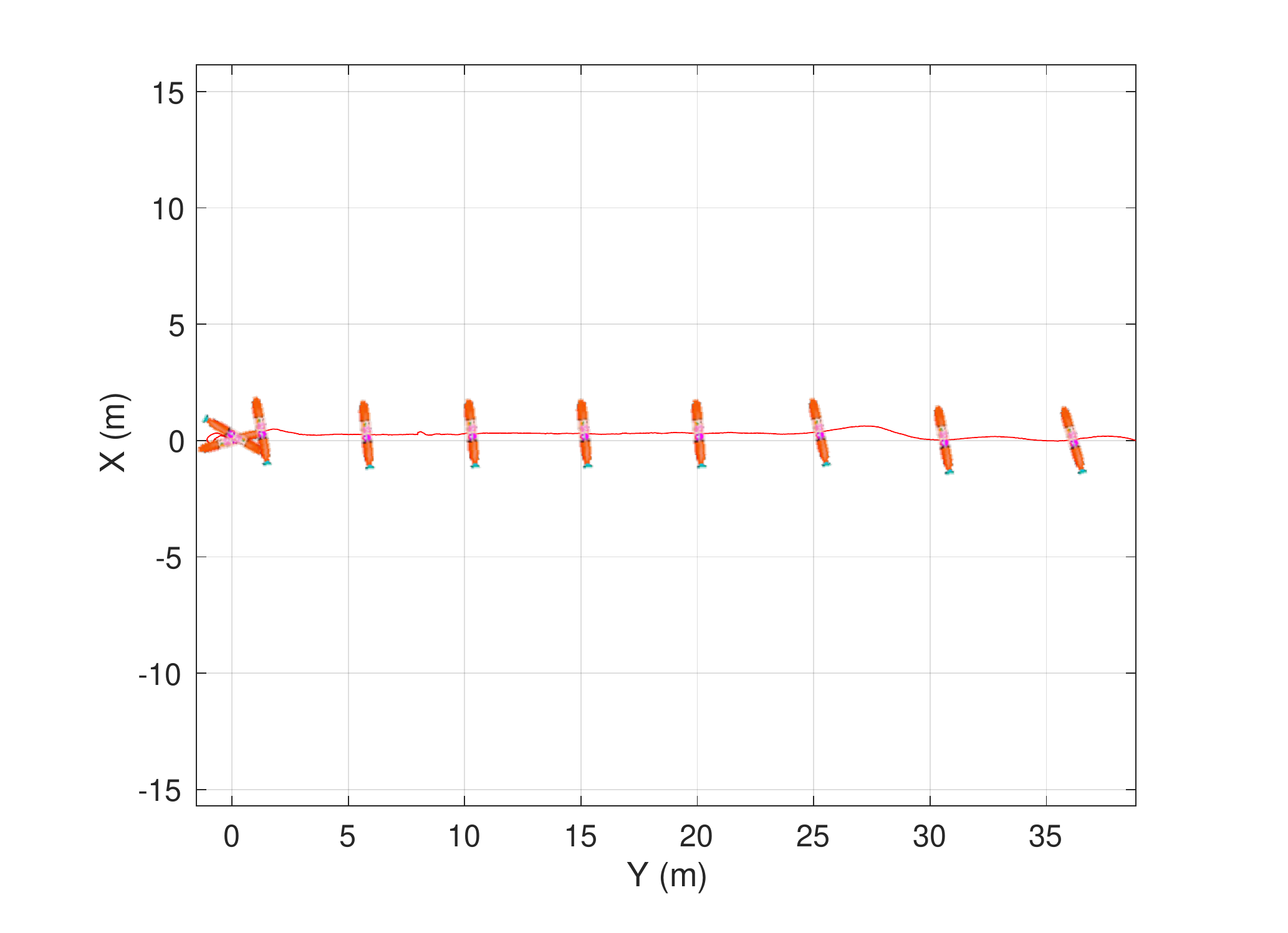} 
	\caption{Vehicle path using the method described in Section \ref{sec: fully actuated}. The initial positions of the vehicle is (0,0).}
	\label{fig:position_fully_actuated}
\end{figure}

\begin{figure}[H]
	
  \centering
  \includegraphics[trim=0cm 7cm 0cm 7.5cm,width=0.6\linewidth,clip]{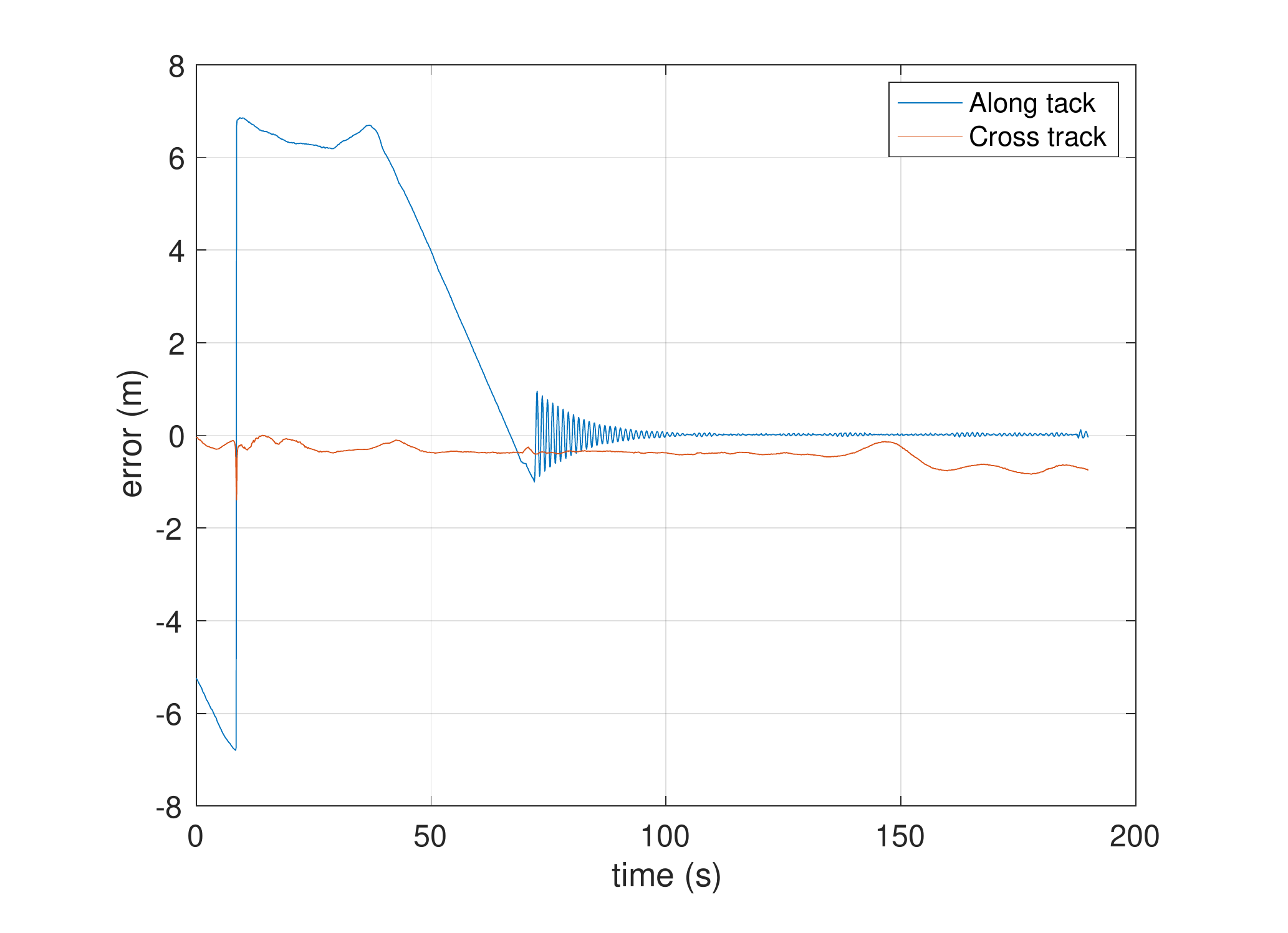} 
  \caption{Along and cross track errors during the mission of Figure \ref{fig:position_fully_actuated}.}
  \label{fig:errors_fully_actuated}
\end{figure}

\begin{figure}[H]
  \centering
  \includegraphics[trim=0cm 7cm 0cm 7.5cm,width=0.6\linewidth,clip]{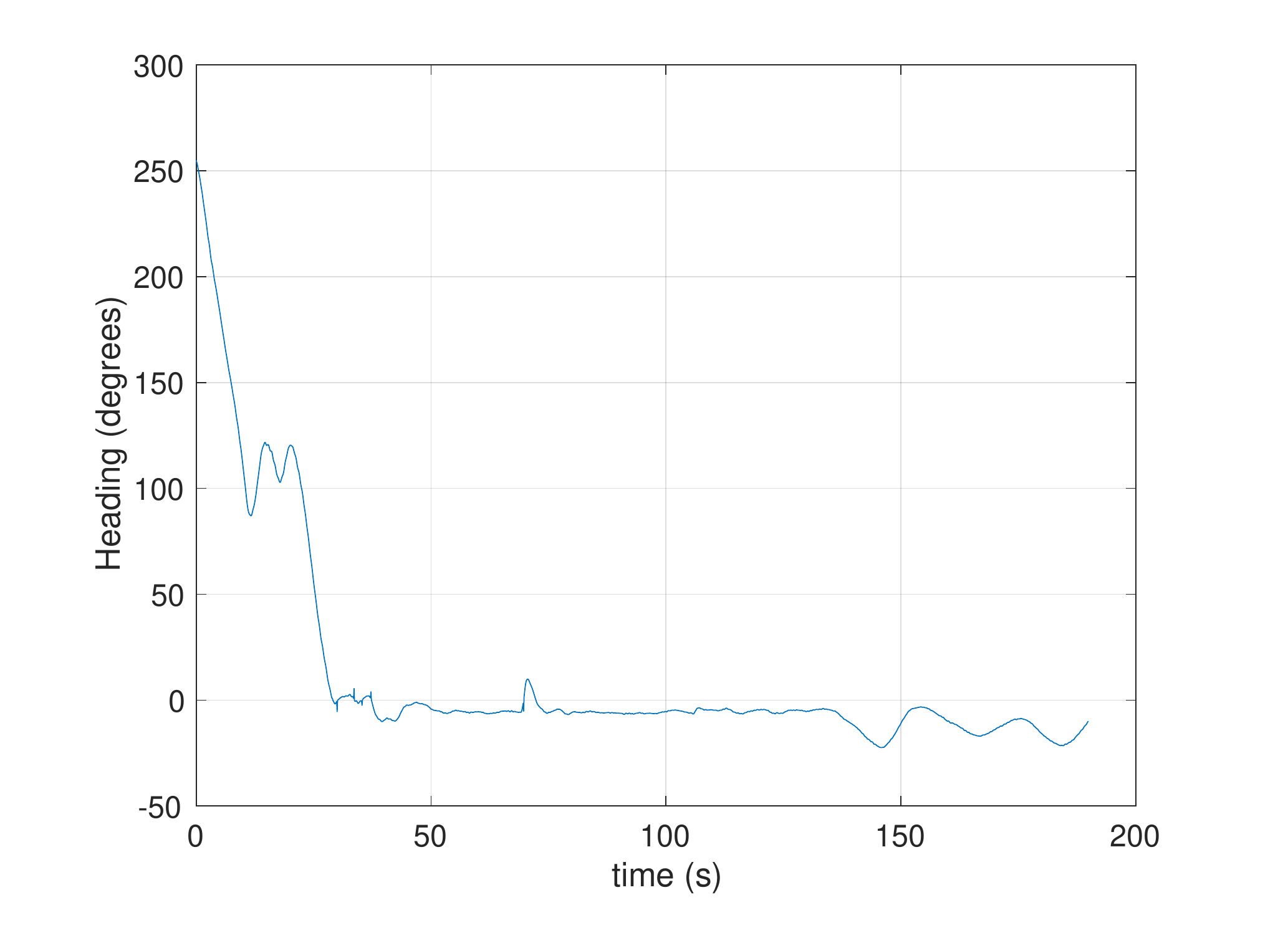} 
  \caption{Yaw during the mission of Figure \ref{fig:position_fully_actuated}.}
  \label{fig:yaw_fully_actuated}
\end{figure}

The method for over-actuated vehicles described in Section \ref{sec: fully actuated} was tested with a  over-actuated Medusa class vehicle following a line due east (along the $Y$ axis) with a heading $\psi$ of $0^{\circ}$. The obtained path, along and cross track erros, and yaw are shown in Figures \ref{fig:position_fully_actuated}, \ref{fig:errors_fully_actuated} and \ref{fig:yaw_fully_actuated}, respectively.


Given the coupling between surge, sway and yaw rate in the dynamics of the vehicle, the tunning of the PID inner loops of the real vehicle is a challenging task and we were not able to obtain adequate performance of circular maneuvers during the field trials with simple linear control laws. From Fig. \ref{fig:errors_fully_actuated} one can observe that the along-track error starts at around $5m$ because that is the distance from the initial position of the vehicle to the beginning of the path, but one can observe that the vehicle converges to the beginning of the path in around $70s$. Also one can observe that the cross-track error remains below $1m$ during the mission. Finally, from Fig. \ref{fig:yaw_fully_actuated} the heading converges to within $15^\circ$ of the reference. Future work will involve the design of new nonlinear inner loop controllers to achieve better performance.

%% file: DISCUSSION.tex
\section{Discussions} \label{sec: discussion}
\subsection{Advantages and disadvantages of the different path following methods}
\begin{figure}[hbt!]
	\centering
	\includegraphics[trim=0cm 0cm 0cm 0cm,width=0.7\linewidth,clip]{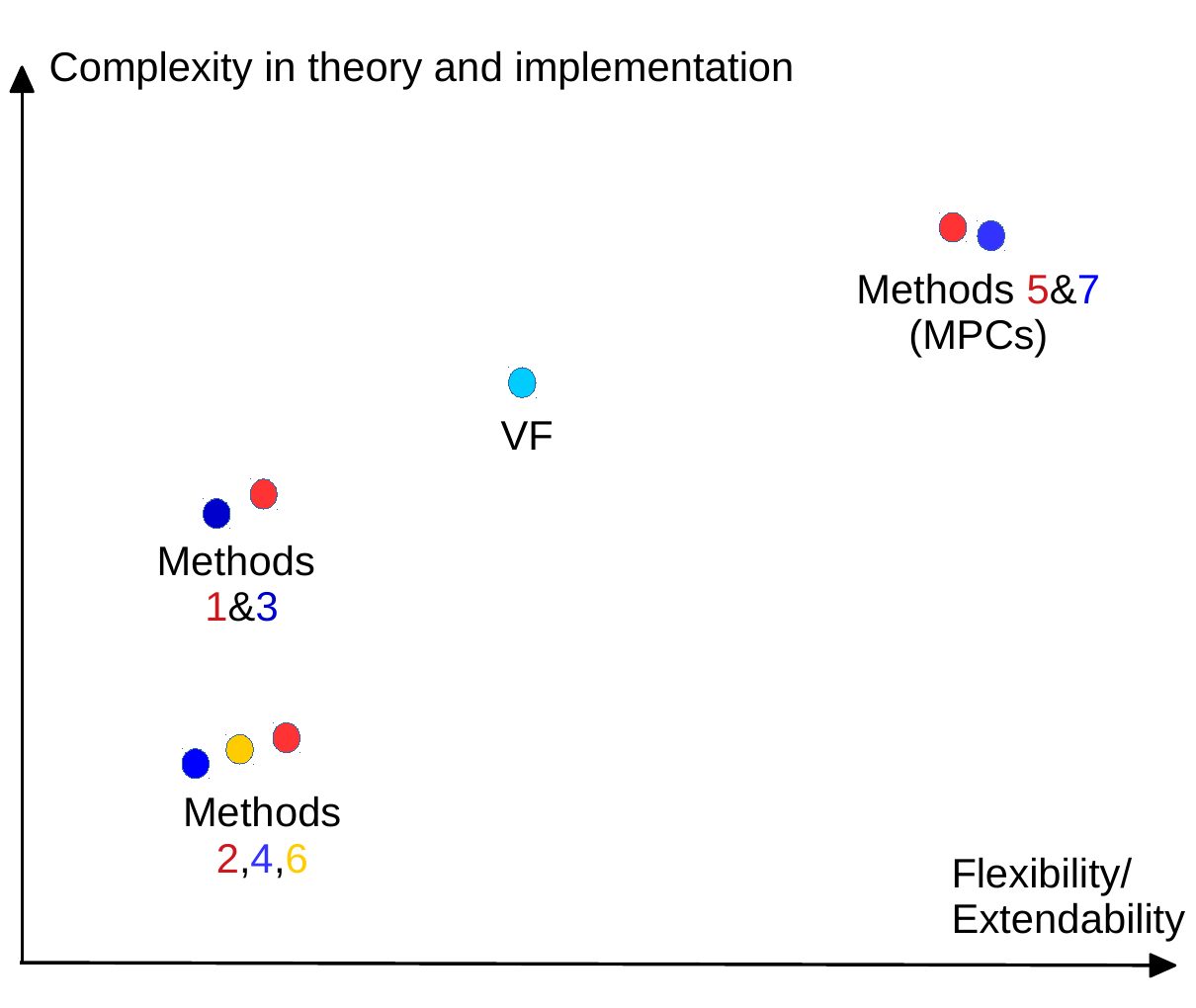} 
	\caption{A comparison among the path following methods. }
	\label{fig:compare path method}
\end{figure}

In this section we discuss some of the benefits and drawbacks of the path following methods reviewed in the previous sections and compare them with other methods in the literature such as the vector field (VF) method described in \cite{Nelson2007}. A relative comparison in terms of complexity and flexibility is shown in Fig.\ref{fig:compare path method}.\\
 From an implementation standpoint, Methods 2,4, and 6 are the simplest and can therefore be considered as good candidates to solve path following problems. These methods are elegant in the sense that the ``reference point" on the path that the vehicle must track in order to achieve path following is not necessary the closest to the vehicle. Instead, it can be initialized anywhere on the path, after which its evolution is controlled via $\dot{\gamma}$ or $\ddot{\gamma}$ to attract the vehicle to the path and make it follow that path with a desired speed. Among these methods, Method 4 is the simplest while Method 2 is the more complicated since it requires to know the knowledge of path's curvature. However, it can be expected that for time varying curvature paths, Method 2 will outperform Method 4.   \\
  Methods 1 and 3, on the other hand, are more complex in general, as they consider \emph{``reference point"} the one closest to the vehicle which, in the case of general paths requires solving an optimization on-line to find this point. Comparing the two methods, it can be said that Method 1 is more complicated as it requires to know the function of curvature along the path; however, one might expect that it outperforms Method 3 for paths with varying curvature. \\
  VF methods are more flexible than those discussed above in the sense that they can be extended to deal with the problem of obstacle avoidance \cite{Wilhelm2019}. However, they are more complex in terms of derivation and proof of stability.\\
  Methods 5 and 7 (NMPC-based approaches) are the most complicated, in terms of design and implementation, since they involve solving an optimization problem online that is non-convex, due to the kinematic constraints of the vehicles. However, these methods deal with the vehicle's input constraints (eg. velocity and angular rate) explicitly, an important feature that none of the methods mentioned before has. For this reason, NMPC based-approaches are expected to outperform the others in path following missions that require the vehicle to maneuver more aggressively, pushing the actuation values close to their limits. Another advantage of NMPC-based methods is that they can incorporate easily other tasks such as obstacle avoidance in the path following problem \cite{SHIBATA2018}.
   
\subsection{Other issues }
As explained before, the main focus of this paper is on path following methods that can be designed by taking into consideration the vehicle kinematics only. 
It is tacitly assumed that inner control loops can be implemented to make the vehicle´s relevant variables track with good accuracy the commands issued by the outer path following control loop (e.g. speed, heading, heading rate, etc). This design approach is widely adopted by control practitioners because it simplifies considerably the design, analysis, and implementation steps. Experimental results reported in the literature, see for example
\cite{Thrun2007,FranciscoLBC2019,ROB:ROB20303} and the results presented in Section \ref{sec: experiement}, suggest
that this should be the first candidate approach to solve path following control problems. From a theoretical standpoint, however it is generally not trivial to guarantee that the path following methods developed at the kinematic level also preserve adequate performance with non-ideal tracking inner-loop controllers. With Method 6, the conclusion is in the affirmative as reported in \cite{FranciscoLBC2019,Vannithesis,Pramod2009}, where it is shown that  the complete path following error system is ISS with respect to the tracking errors associated with speed and yaw-rate tracking controllers. This implies that as long as the tracking errors are bounded then the path following error is bounded. Relate results can be found in \cite{Antoniobook}. A similar property may very well hold with the other path following methods summarized in this paper;  however, proofs for these methods are not available in the literature. \\
Another issue with the inner-outer loop approach is that it requires the inner-loop controllers to be fine tuned to guarantee that the complete path following system is stable and and achieves a desired  degree of performance; the tuning process can be time-consuming and in many cases it is also expensive. In order to deal with this problem, the work in \cite{Isaac2010}  suggests augmenting the inner-loop controllers with “\L1 adaptive controllers", that aim at eliminating the effect of inner-loop uncertainty on the total path following system. See \cite{Divine2015,Haitong2021} for applications of this strategy to an underwater robot and a surface ship, respectively. \\ 
Another type of path following methods involves including explicitly the  vehicle dynamics in the design process. A typical example of this approach can be found in\cite{lapierre2006nonsingular,aguiar2007trajectory} where the authors employ backstepping technique sto derive control laws for the vehicle force and toque to achieve path following. The main challenge of this approach is that it is harder to design as the vehicle model adopted for controller design is far more complex than the kinematic model. In addition, the dynamics of the vehicle normally exhibit large uncertainty in the parameters (e.g. mass, momentums of inertia, hydrodynamic or aerodynamic drag terms, etc.) thus making the design process quite challenging. \\
Finally, there is also a considerable interest in using learning-based methods for the path following problem in recent years. These type of method aims at dealing with vehicles model and environment uncertainty. Along this line, learning-based MPC methods suggested in \cite{Ostafew2016,Rokonuzzaman2020} seem to be a promising approach while reinforcement learning-based methods described in \cite{MARTINSEN2018,Kamran2019,Wang2021} are also worth considering. Although these methods are powerful, they are far complicated than the methods derived in Section \ref{sec: pf method in 2D}. Furthermore these methods lack of stability guarantee and for RL-based methods, they lack of intuition underlying the controller and thus make it difficult to tune.  

%% file: CONCLUSION.tex
\section{Conclusions} \label{sec: conclusion}
Path following of autonomous robotic vehicles is a fairly well understood and established technique for kinematic vehicle models, and a variety of solutions to the path following problem have been published in the literature. Motivated by the need to bring many of the methods under a common umbrella, this paper presented an in-depth review of the topic of path following for autonomous vehicle moving in 2D, showing clearly how many of the existing techniques can be derived under a unified mathematical framework that makes use of nonlinear control theory. The paper provided a rigorous analysis of the main classes of methodologies identified and discussed the advantages and disadvantages of each method, comparing them from a design and implementation standpoint. In addition, the paper described the steps involved in going from theory to practice through the introduction of Matlab and Gazebo/ROS simulation toolboxes that are helpful in testing path following methods prior to their integration in the combined guidance, navigation, and control systems of autonomous vehicles. Finally, the paper discussed the results of experimental field tests with underactuated and overactuated autonomous marine vehicles performing path following maneuvers. \\
The simulations and experimental results showed that as long as the inner-loop /dynamic) tracking controllers for speed, heading, or heading rate issued by the (outer loop/kinematic ) path following controller are well tuned and there is adequate time scale separation of the inner and outer loops systems, the methods exposed in the paper for kinematic models hold very good potential for real life applications.\\
For missions involving vehicles for which the inner and outer loop dynamics do not exhibit a clear two scale separation and yet require good path following performance, methods that take explicitly into account the vehicle dynamics in the design of path following strategies may be required. Considerable research has been done in this area, but to the space limitations this survey did not address them. See \cite{lapierre2006nonsingular} for a representative early example that resorts to backstepping and vehicle parameter adaptation strategies. The example shows clearly how the explicit intertwining of kinematics and dynamics in path following control system design complicates the design process, yields controllers with increased complexity that may be difficult to implement, and requires knowledge of the vehicle´s parameters (such as hydrodynamic or aerodynamic terms) and its actuators, which normally exhibit high degrees of uncertainty. A trending approach to deal with uncertainty in vehicle and actuator models is to use learning based techniques such as learning-MPC or reinforcement learning for path following. However, existing methods lack formal stability guarantees and fail to capture physically-based intuition that is often crucial in the design and tuning of advanced control systems. Clearly, further research  is required to ascertain if such methods can be further developed to obtain proven performance guarantees, a significant step required to make a control systems applicable in practice.

%% file: APPENDIX.tex
\section{Appendix}
\subsection{Basic results}
The following lemma was used in the paper.
\begin{lemma}[Rotation Matrix Differential Equation, Theorem 2.2 \cite{fossen2011handbook}]
	Let $R^{B}_{A} \in SO(n)$\footnote{A special orthogonal group with dimension $n$, defined as $SO(n)=\{ R\in \R^{n\times n}: RR^{\rm T}=R^{\rm T} R=I_n, \det{R}=1  \}$ }$(n=2,3)$ be the rotation matrix from frame $\{A\}$ to frame $\{B\}$. Then,
	\begin{equation} \label{eq: Rotation DE}
	\dot{R}^{ B}_{A}={R}^{ B}_{A}S(\bs{\omega}^{A}_{A/B}),
	\end{equation}	
	where $S(\bs{\omega}^{A}_{A/B})$ is a Skew-symmetric matrix\footnote{A square matrix $S$ is called Skew-symmetric matrix iff $S^{\rm T}=-S$.} and $\bs{\omega}^{A}_{A/B}\in \R^{n} $ is the angular velocity vector of $\{A\}$ respect to $\{B\}$, expressed in $\{A\}$.	
	\label{lemma: rotation matrix differential equation}
\end{lemma}
\subsection{Proofs }  
\subsubsection{Proof of Theorem \ref{theorem: samson method}} \label{Proof of theorem: samson method}
Consider the Lyapunov function candidate $V_1$ defined by 
\begin{equation}
V_1=\frac{1}{2}y^2_1+\frac{1}{2k_2}\tilde{\psi}^2
\end{equation}
Taking its time derivative yields
\begin{equation} \label{eq: V1 dot}
\dot{V}_{1}=y_{1}\dot{y}_{1}+\frac{1}{k_2}\tilde{\psi}\dot{\tilde{\psi}}\stackrel{\eqref{eq: y1dot},\eqref{eq: dot_tilde_psi},\eqref{eq: r samson}}{=}y_1u\sin\delta-\frac{k_1}{k_2}\tilde{\psi}^2.
\end{equation}
Because of Condition \ref{con: for delta}, $\dot{V}_1\le 0$ for all $t$. Furthermore, the above equation shows that $\dot{V}_1$ is uniform continuous and therefore, using Barbalat's Lemma \cite{khalil2002}, we conclude that $\dot{V}_1(t)$ converges to zero as $t\to \infty$. In view of \eqref{eq: V1 dot} and Condition \ref{con: for delta}, this implies that $y_1(t), \tilde{\psi}(t) \to 0$ as $t \to \infty$. \hfill $\blacksquare$
 \subsubsection{Proof of Theorem \ref{theorem: Lapier method}} \label{Proof of theorem: Lapier method}
 
 The proof can be done by employing the Lyapunov function candidate given by
 \begin{equation}
 V_2=\frac{1}{2}\eP^2+\frac{1}{2k_2}\tilde{\psi}^2.
 \end{equation}
 Taking its time derivative yields
 \begin{equation}
 \dot{V}_2= \eP^{\top}\dot{\e}_{\rmP}-\frac{1}{k_2}\tilde{\psi}\dot{\tilde{\psi}} \stackrel{\eqref{eq: dot eP}, \eqref{eq: dot_tilde_psi}}{=}\eP^{\top}\begin{bmatrix}
 u\cos(\psi_{\rm e})-u_{\rmP} \\
 u\sin(\psi_{\rm e})
 \end{bmatrix} +\frac{1}{k_2}\tilde{\psi}(r-\kappa(\gamma)u_{\rmP}-\dot{\delta}).
 \end{equation}
In the above, we have used the fact that for any Skew-symmetric matrix $S$, ${\bf x}^{\top}S{\bf x}=0$ for all ${\bf x}$, thus $\eP^{\top} S(\bs{\omega}_{\rmP})\eP =0 $. 
 Substituting $r$ given by \eqref{eq: r samson} and $u_{\rmP}$ given by \eqref{eq: v lapier} in $\dot{V}_2$, we obtain
 \begin{equation} \label{eq: dot V2}
 \dot{V}_2=-k_3s^2_1+y_1u\sin\delta-\frac{k_1}{k_2}\tilde{\psi}^2.
 \end{equation}
 Due to Condition \ref{con: for delta}, we conclude that $\dot{V}_2\le 0$ for all $t$. Furthermore, the above equation shows that $\dot{V}_2$ is uniform continuous; thus, invoking Barbalat's lemma \cite{khalil2002}, $\dot{V}_2(t)$ converges to 0 as $t\to\infty$. In view of \eqref{eq: dot V2} this implies that $s_1(t),y_1(t),\tilde{\psi}(t)\to 0$ as $t\to\infty$. Notice that because of Condition \ref{con: for delta} once $y_1,\tilde{\psi}\to 0$, $\psi_e\to 0$ as well.  \hfill $\blacksquare$
\subsubsection{Proof of Theorem \ref{theorem: method 3}} \label{Proof of theorem: method 3}
Consider the Lyapunov function candidate $V_3$, given by $V_3=y^2_1/2$. Taking its time derivative yields
$$\dot{V}_3=y_1\dot{y}_1\stackrel{\eqref{eq: y1dot},\eqref{eq: psie in PF method 3}}{=} -u\frac{y^2_1}{\sqrt{y^{2}_1+\Delta^2_{\rm h}}} \stackrel{\eqref{eq: u samson}}{=}-U_{\rm d}\frac{y^2_1}{\sqrt{y^{2}_1+\Delta^2_{\rm h}}} $$
Since $U_{\rm d}$ (the desired speed profile for the vehicle to track) is always positive, $\dot{V}_3\le 0$ for all $y_1$. It can be seen that $\dot{V}_3$ is uniform continuous,  and therefore invoking Barbalat's lemma \cite{khalil2002} $\dot{V}_3(t)$ converges to 0 as $t\to\infty$. This implies that $y_1(t)$ converges to zero as $t\to \infty$. \hfill $\blacksquare$
\subsubsection{Proof of Theorem \ref{theorem: method 4}} \label{Proof of theorem: method 4}
First, substituting $u_{\rmP}$ given by \eqref{eq: v lapier}  and $\psi_{\rm e}$ given by \eqref{eq: psie in PF method 3} in \eqref{eq: dot eP} yields the dynamics of the position error in the resulting closed-loop system as
\begin{equation} \label{eq: doteP closed loop method 3}
\dot{\e}_{\rmP}=-S(\bs{\omega}_{\rmP})\eP-\begin{bmatrix}
-k_3s_1 \\
-uy_1/\sqrt{y^2_1+\Delta^2_{\rm h}} \end{bmatrix}.
\end{equation}
Consider the Lyapunov function candidate, given by $V_4(\eP)=\frac{1}{2}\e^{\top}_{\subscript{\P}}\eP$.
Taking its time derivative yields
\begin{equation}
\dot{V}_4=\e^{\top}_{\rmP}\dot{\e}_{\rmP}\stackrel{\eqref{eq: doteP closed loop method 3}}{=} -k_3s^{2}_1-u\frac{y^2_1}{\sqrt{y^{2}_1+\Delta^2_{\rm h}}} \stackrel{\eqref{eq: u samson}}{=}-k_3s^{2}_1-U_{\rm d}\frac{y^2_1}{\sqrt{y^{2}_1+\Delta^2_{\rm h}}}.
\end{equation}
Note that we used the fact that $\e^{\top}_{\rmP}S(\bs{\omega}_{\rmP})\e_{\rmP}=0$ for all $\e_{\rmP}$ because $S$ is a Skew-symmetric matrix. Furthermore, because $u=U_{\rm d}$, which is always positive, we conclude that $\dot{V}_4 <0$ for $\e_{\rmP}$. Since $u$ is in general a function of time then the closed-loop position error system \eqref{eq: doteP closed loop method 3} is non-autonomous, therefore we conclude that the origin of $\eP$ is UGAS. \hfill $\blacksquare$
\subsubsection{Proof of Theorem \ref{theorem: Aguiar method}} \label{Proof of theorem: Aguiar method}
Consider a Lyapunov function candidate for the path following system \eqref{eq: complete the path following error system Aguiar method}, given by 
\begin{equation} \label{eq: V6}
V_6({\bf x})=\frac{1}{2}\Enorm{\bf x}^2= \frac{1}{2}\e^{\top}_{\subscript{\B}}\eB + \frac{1}{2}e^2_{\gamma}.
\end{equation}
Taking its time derivative yields
\begin{equation}
\begin{split}
\dot{V}_6&=\e^{\top}_{\subscript{\B}}\dot{\e}_{\subscript{\B}}+e_{\gamma}\dot{e}_{\gamma}\\
         &\stackrel{\eqref{eq: following error aguiar method},\eqref{eq: gamma dot}}{=}\e^{\top}_{\rmB} \left(−S(\bs \omega){\e}_{\subscript{\B}} +\Delta {\bf u} −  \RIB(\psi)\p'_{\rm d}(\gamma)\dot{\gamma}\right)+e_{\gamma}\left(\ddot{\gamma}-\dot{v}_{\rm d}\right).  \\
         &\stackrel{\eqref{eq: Aguiar controller}, \eqref{eq: e gamma} }{=}-\e_{\rmB}^{\top}K_{\rm p}\e_{\rmB}-k_{\gamma} e^2_{\gamma}.
\end{split}
\end{equation}
Let $K={\rm diag}(K_{\rm p},k_{\gamma})$, then $\dot{V}_6= -{\bf x}^{\top}K{\bf x} \le -\lambda_{\min}(K)\Enorm{\bf x}^2$ for all ${\bf x} \neq \bs{0}$. Thus, we conclude that the origin of $\bf x$ is GES.  \hfill $\blacksquare$
\subsubsection{Proof of Lemma \ref{lemma: aguiar method with current} }
\label{proof of lemma: aguiar method with current}
Consider again the Lyapunov function candidate $V_6$ given by \eqref{eq: V6}. Substituting $\u$ in \eqref{eq: Aguiar control law for u current} and $\ddot{\gamma}$ in \eqref{eq: Aguiar control law for gammaddot} in \eqref{eq: following error aguiar method with current} and \eqref{eq: gamma dot} yields
\begin{equation}
\begin{split}
\dot{V}_6&=\e^{\top}_{\subscript{\B}}\dot{\e}_{\subscript{\B}}+e_{\gamma}\dot{e}_{\gamma}\\
 & = -\e_{\rmB}^{\top}K_{\rm p}\e_{\rmB}-k_{\gamma} e^2_{\gamma} - \e_{\rmB}^{\top}\RIB(\psi){\bf e}_c \\
 & \le -\lambda_{\rm min}(K)\Enorm{\x}^2 + \Enorm{\x}\Enorm{\e_c}
\end{split}
\end{equation}
Thus, invoking Theorem 4.19 in \cite{khalil2002} we conclude that the path following error system is ISS respect to the input $\e_c$. \hfill $\blacksquare$	  
 
\subsubsection{Proof of Theorem \ref{theorem: Fully actuated method}} \label{Proof of theorem: Fully Actuated}
Consider a Lyapunov function candidate for the path following system \eqref{eq: complete path following error fully actuated}, given by 
\begin{equation} \label{eq: V6}
V_F({\bf x})=\frac{1}{2}\Enorm{\bf x}^2= \frac{1}{2}\e^{\rm T}_{\subscript{\B}}\eB + \frac{1}{2}e^2_{\gamma}.
\end{equation}
Taking its time derivative yields
\begin{equation}
\begin{split}
\dot{V}_F&=\e^{\rm T}_{\subscript{\B}}\dot{\e}_{\subscript{\B}}+e_{\gamma}\dot{e}_{\gamma}\\
&\stackrel{\eqref{eq: following error aguiar method},\eqref{eq: gamma dot}}{=}\e^{\rm T}_{\rmB} \left(−S(\bs \omega){\e}_{\subscript{\B}} +{\bf v} −  \RIB(\psi)\p'_{\rm d}(\gamma)\dot{\gamma}\right)+e_{\gamma}\left(\ddot{\gamma}-\dot{v}_{\rm d}\right).  \\
&\stackrel{\eqref{eq: fully actuated controller}, \eqref{eq: e gamma} }{=}-\e_{\rmB}^{\rm T}K_{\rm p}\e_{\rmB}-k_{\gamma} e^2_{\gamma}.
\end{split}
\end{equation}
Let $K={\rm diag}(K_{\rm p},k_{\gamma})$, then $\dot{V}_F= -{\bf x}^{\rm T}K{\bf x} \le -\lambda_{\min}(K)\Enorm{\bf x}^2$ for all ${\bf x} \neq \bs{0}$. Thus, we conclude that the origin of $\bf x$ is GES.  \hfill $\blacksquare$